\documentclass[times,referee,twocolumn,final,authoryear]{elsarticle}

\usepackage{ycviu}
\usepackage{framed,multirow}

\usepackage{amssymb}
\usepackage{latexsym}

\usepackage{url}
\usepackage{xcolor}
\usepackage{color}
\definecolor{newcolor}{rgb}{.8,.349,.1}

\usepackage{colortbl}
\definecolor{tabgray}{gray}{.90}
\definecolor{tabpink}{rgb}{.99,.85,.95}
\definecolor{tabcyan}{cmyk}{.3,0,0,0}

\usepackage{epsfig}
\usepackage{graphicx}
\usepackage{algorithm}
\usepackage{algorithmic}
\usepackage{mathrsfs} 
\usepackage{amsmath}
\usepackage{mdwlist}
\usepackage{bbding}
\usepackage{enumerate}
\usepackage{booktabs}
\usepackage{array}
\usepackage{caption}
\usepackage{subfigure}
\usepackage{xspace}
\usepackage{paralist}
\usepackage{supertabular}

\def\ie{{\em i.e.}}
\def\eg{{\em e.g.}}

\def\ds{UA-DETRAC\xspace} 

\journal{Computer Vision and Image Understanding}

\begin{document}

\thispagestyle{empty}

\ifpreprint
  \setcounter{page}{1}
\else
  \setcounter{page}{1}
\fi

\begin{frontmatter}

\title{UA-DETRAC: A New Benchmark and Protocol for Multi-Object Detection and Tracking}
\author[1]{Longyin \snm{Wen}}
\author[2]{Dawei \snm{Du}}
\author[3]{Zhaowei \snm{Cai}}
\author[4]{Zhen \snm{Lei}}
\author[2]{Ming-Ching \snm{Chang}}
\author[5]{Honggang \snm{Qi}}
\author[6]{Jongwoo \snm{Lim}}
\author[7]{Ming-Hsuan \snm{Yang}}
\author[2]{Siwei \snm{Lyu}\corref{cor1}}
\cortext[cor1]{Corresponding author: Tel.: +1-518-437-4938.}
\ead{slyu@albany.edu}

\address[1]{JD Finance America Corporation, Mountain View, CA, USA}
\address[2]{Computer Science Department, University at Albany, State University of New York, Albany, NY, USA}
\address[3]{Department of Electrical and Computer Engineering, University of California, San Diego, CA, USA}
\address[4]{National Laboratory of Pattern Recognition, Institute of Automation, Chinese Academy of Sciences, Beijing, China}
\address[5]{School of Computer and Control Engineering, University of Chinese Academy of Sciences, Beijing, China}
\address[6]{Division of Computer Science and Engineering, Hanyang University, Seoul, Korea}
\address[7]{School of Engineering, University of California at Merced, CA, USA}

\begin{abstract}
Effective multi-object tracking (MOT) methods have been developed in recent years for a wide range of applications including visual surveillance and behavior understanding. Existing performance evaluations of MOT methods usually separate the tracking step from the detection step by using one single predefined setting of object detection for comparisons. In this work, we propose a new University at Albany DEtection and TRACking (\ds) dataset for comprehensive performance evaluation of MOT systems especially on detectors. The \ds benchmark dataset consists of $100$ challenging videos captured from real-world traffic scenes (over $140,000$ frames with rich annotations, including illumination, vehicle type, occlusion, truncation ratio, and vehicle bounding boxes) for multi-object detection and tracking. We evaluate complete MOT systems constructed from combinations of state-of-the-art object detection and tracking methods. Our analysis shows the complex effects of detection accuracy on MOT system performance. Based on these observations, we propose effective and informative evaluation metrics for MOT systems that consider the effect of object detection for comprehensive performance analysis.
\end{abstract}


\end{frontmatter}


\begin{figure*}[t]
\centering
\includegraphics[width=.9\textwidth]{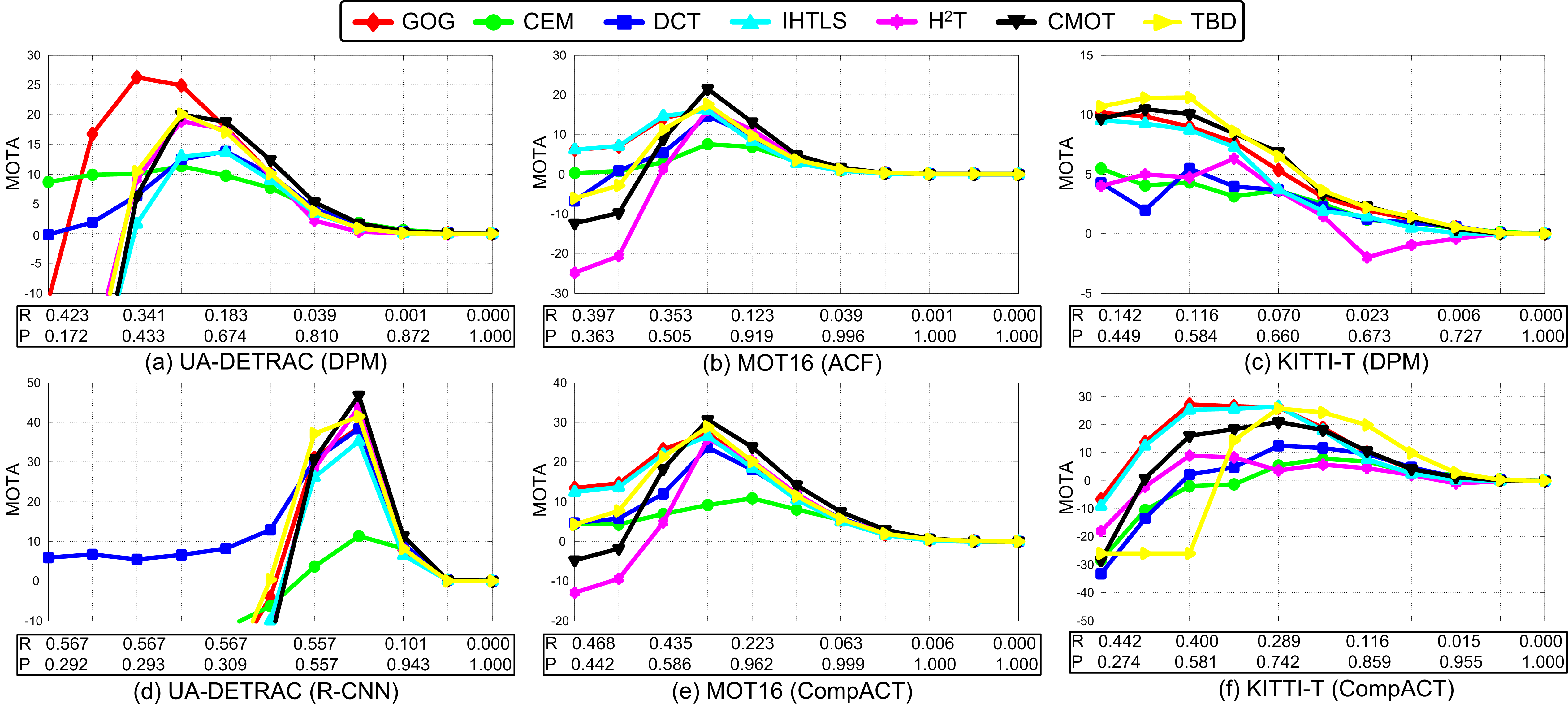}
\caption{Precision-recall curves corresponding to the MOT systems on the \ds, MOT16 \citep{DBLP:journals/corr/Anton16}, and KITTI-T \citep{DBLP:conf/cvpr/GeigerLU12} datasets, constructed by four object detection algorithms: DPM \citep{DBLP:journals/pami/FelzenszwalbGMR10}, ACF \citep{DBLP:journals/pami/DollarABP14}, R-CNN \citep{DBLP:conf/cvpr/GirshickDDM14}, and CompACT \citep{DBLP:conf/iccv/CaiMN15} with seven object tracking algorithms: GOG \citep{DBLP:conf/cvpr/PirsiavashRF11}, CEM \citep{DBLP:conf/cvpr/AndriyenkoS11}, DCT \citep{DBLP:conf/cvpr/AndriyenkoSR12}, IHTLS \citep{DBLP:conf/iccv/DicleCS13}, H\textsuperscript{2}T \citep{DBLP:conf/cvpr/Longyin14}, CMOT \citep{DBLP:conf/cvpr/BaeY14}, and TBD \citep{DBLP:journals/pami/GeigerLWSU14}. The $x$-axis corresponds to different precision/recall scores of detectors obtained by varying the detection score threshold. The $y$-axis is the MOTA score in traditional CLEAR MOT metrics \citep{DBLP:conf/clear/StiefelhagenBBGMS06} of the MOT system constructed by detection and tracking methods. Note that with different detection score thresholds, the performance differences of different MOT systems vary significantly according to the MOTA scores.
}
\label{fig:pr-mota-curves}
\end{figure*}

\section{Introduction}
\label{sec:intro}
Multiple object tracking (MOT), which aims to extract trajectories of numerous moving objects in an image sequence, is a crucial task in video understanding. A robust and reliable MOT system is the basis for a wide range of applications including video surveillance, autonomous driving, and sports video analysis. To construct an automatic tracking system, most effective MOT approaches, \eg, \citep{DBLP:journals/pami/KhanBD05,DBLP:conf/cvpr/ZhangLN08,DBLP:conf/cvpr/BenfoldR11,DBLP:journals/pami/BreitensteinRLKG11,DBLP:conf/eccv/IzadiniaSLS12,DBLP:conf/eccv/YangN12,DBLP:journals/pami/HuangLN13,DBLP:conf/cvpr/YangLWYL14,DBLP:conf/cvpr/Longyin14,dehghan2015cvpr-gmmcp}, require a pre-trained detector, \eg, \citep{DBLP:journals/pami/FelzenszwalbGMR10,DBLP:journals/pami/DollarABP14,DBLP:conf/cvpr/GirshickDDM14,DBLP:conf/cvpr/YanLWL14,DBLP:conf/iccv/CaiMN15,DBLP:conf/cvpr/RedmonDGF16} to discover the target objects in the video frames (usually with bounding boxes). As such, a general MOT system entails an object detection step to find target locations in each video frame, and an object tracking step that generates target trajectories across video frames\footnote{In this work, we use the explicit definition of MOT system, \ie, {\em MOT system} $=$ {\em detection} $+$ {\em tracking}.}.

\begin{table*}[t]
  \centering
    \caption{Summary of existing object detection or tracking datasets. First six columns: the number of training/testing data ($1k=10^{3}$) indicating the number of images containing at least one object, the number of object tracks, and the number of unique object bounding boxes. Remaining columns: additional properties of each dataset, \ie, ``D'': detection task, ``T'': tracking task, ``P'': target object is pedestrian, and ``C'': target object is vehicle.}
  \label{tab:object-tracking-detection-dataset}
  \setlength{\tabcolsep}{2.8pt}
  \footnotesize{
  \begin{tabular}{c||c|c|c|c|c|c||c|c|c|c|c|c|c}
  \hline
  \multirow{2}{*}{Dataset} &\multicolumn{3}{c|}{Training Set} &\multicolumn{3}{c||}{Testing Set} &\multicolumn{7}{c}{Properties}  \\
  \cline{2-14}
  &Frame   &Tracks
  &Boxes
   &Frame   &Tracks
  &Boxes
  &Color
  &Video   &Task
  &Object
  &Illumination
  &Occlusion   &Year \\
  \hline
  \hline

  INRIA \citep{DBLP:conf/cvpr/DalalT05}                  &1.2$k$ &- &1.2$k$ &741 &- &566 &\Checkmark & &D &P & & &2005 \\
  ETH \citep{DBLP:conf/iccv/EssLG07}                     &490 &- &1.6$k$ &1.8$k$ &- &9.4$k$ &\Checkmark &\Checkmark &D &P & & &2007 \\
  NICTA \citep{DBLP:conference/ivs/Overett08}            &- &- &18.7$k$ &- &- &6.9k &\Checkmark & &D &P & & &2008 \\
  TUD-B \citep{DBLP:conf/cvpr/WojekWS09}              &1.09$k$ &- &1.8$k$ &508 &- &1.5$k$ &\Checkmark & &D &P & & &2009 \\
  Caltech \citep{DBLP:journals/pami/DollarWSP12}         &67$k$ &- &192$k$ &65$k$ &- &155$k$ &\Checkmark &\Checkmark &D &P & &\Checkmark &2012 \\
  CUHK \citep{DBLP:conf/cvpr/OuyangW12}                  &- &- &- &1.06$k$ &- &- &\Checkmark & &D &P & & &2012 \\
  KITTI-D \citep{DBLP:conf/cvpr/GeigerLU12}                   &7.48$k$ &- &40.6$k$ &7.52$k$ &- &39.7$k$ &\Checkmark &\Checkmark &D &P,C & &\Checkmark &2014 \\
  KAIST \citep{Hwang-2015-CVPR}           &50.2$k$ &- &41.5$k$ &45.1$k$ &- &44.7$k$ &\Checkmark &\Checkmark &D &P &\Checkmark &\Checkmark &2015 \\

  \hline
  \hline
  BU-TIV \citep{DBLP:conf/cvpr/WuFTB14}                  &- &- &- &6556 &- &- &\Checkmark &\Checkmark &T &P,C & & &2014 \\
  MOT15 \citep{DBLP:journals/corr/Leal-TaixeMRRS15}        &5.5$k$ &500 &39.9$k$ &5.8$k$ &721 &61$k$ &\Checkmark &\Checkmark &T &P &\Checkmark & &2015 \\
  MOT16 \citep{DBLP:journals/corr/Anton16}                  &5.3$k$ &467 &110$k$ &5.9$k$ &742 &182$k$ &\Checkmark &\Checkmark &T &P,C &\Checkmark &\Checkmark &2016 \\
  \hline
  \hline
  TUD \citep{DBLP:conf/cvpr/AndrilukaRS08}               &610 &- &610 &451 &31 &2.6$k$ &\Checkmark &\Checkmark &D,T &P & & &2008 \\
  PETS2009 \citep{DBLP:conf/avss/FerrymanE09}            &- &- &- &1.5$k$ &106 &18.5$k$ &\Checkmark &\Checkmark &D,T &P &\Checkmark & &2009 \\
  KITTI-T \citep{DBLP:conf/cvpr/GeigerLU12}                   &8$k$ &- &- &11$k$ &- &-  &\Checkmark &\Checkmark &T &C & &\Checkmark &2014 \\
  \hline
  \hline
  \ds                                      &84$k$ &5.9$k$ &578$k$ &56$k$ &2.3$k$ &632$k$ &\Checkmark &\Checkmark &D,T &C &\Checkmark &\Checkmark &2015 \\
  \hline

  \end{tabular}}

\end{table*}

Despite significant advances in recent years, relatively less effort has been made to large scale and comprehensive evaluations of MOT methods, especially for the effect of object detection to MOT performance. Existing MOT evaluation methods usually separate the object detection (\eg, \citep{DBLP:journals/ijcv/EveringhamEGWWZ15,DBLP:journals/pami/DollarWSP12,DBLP:conf/cvpr/GeigerLU12,ILSVRC15}) and object tracking steps (\eg, \citep{DBLP:conf/avss/FerrymanE09,DBLP:conf/pets/Bashir06,DBLP:conf/cvpr/GeigerLU12,DBLP:conf/cvpr/MilanSR13a,DBLP:journals/corr/Leal-TaixeMRRS15}) in comparisons. While this evaluation strategy is widely adopted in the literature, it is insufficient for analyzing complete MOT systems (see Figure \ref{fig:pr-mota-curves}). In particular, it is important to understand the effect of detection accuracy on the complete MOT system performance, which can only be revealed in a comprehensive quantitative study on object detection and tracking steps {\em jointly}.

In this work, we propose a new large-scale University at Albany DEtection and TRACking (\ds) dataset. The \ds dataset includes $100$ challenging videos with more than $140,000$ frames of real-world traffic scenes. These videos are manually annotated with a total of $1.21$ million labeled bounding boxes of vehicles and useful attributes, \eg, illumination of scenes, vehicle type, and
occlusion. Different from other self-driving car datasets (\eg, KITTI \citep{DBLP:conf/cvpr/GeigerLU12}, Berkeley DeepDrive BDD100k \citep{DBLP:journals/corr/abs-1805-04687}, Baidu Apolloscapes \citep{DBLP:conf/cvpr/HuangCGCZWLY18} and Oxford Robotic Car \citep{DBLP:journals/ijrr/MaddernPLN17} datasets), the proposed dataset focuses on detecting and tracking vehicles, which is a thoroughly annotated MOT evaluation dataset containing traffic scenes. Moreover, it poses new challenges for object detection and tracking algorithms. Please see Table \ref{tab:object-tracking-detection-dataset} for a detailed comparison to other benchmark datasets.

We evaluate the complete MOT systems constructed from combinations of ten object tracking schemes (\citep{DBLP:conf/cvpr/AndriyenkoS11,DBLP:conf/cvpr/PirsiavashRF11,DBLP:conf/cvpr/AndriyenkoSR12,DBLP:conf/iccv/DicleCS13,DBLP:conf/cvpr/Longyin14,DBLP:conf/cvpr/BaeY14,DBLP:journals/pami/GeigerLWSU14,DBLP:conf/iccv/KimLCR15,DBLP:conf/avss/BochinskiES17,DBLP:conf/avss/LyuCDLWCCSMDCBG18}) and six object detection methods (\citep{DBLP:journals/pami/FelzenszwalbGMR10,DBLP:journals/pami/DollarABP14,DBLP:conf/cvpr/GirshickDDM14,DBLP:conf/iccv/CaiMN15,DBLP:journals/pami/RenHG017,DBLP:conf/icmcs/WangLWZYX17}), on the \ds, MOT16 \citep{DBLP:journals/corr/Anton16}, and KITTI-T \citep{DBLP:conf/cvpr/GeigerLU12} datasets\footnote{Since the testing sets from the MOT16 \citep{DBLP:journals/corr/Anton16} and KITTI-T \citep{DBLP:conf/cvpr/GeigerLU12} datasets are not publicly available, the experiments are carried out on the training sets. For the MOT16 \citep{DBLP:journals/corr/Anton16} dataset, we train the pedestrian detectors on the INRIA dataset \citep{DBLP:conf/cvpr/DalalT05} and the first $4$ sequences of the training set of MOT16 \citep{DBLP:journals/corr/Anton16}, and train the tracking models on the first $4$ sequences of the training set in MOT16. The remaining $3$ sequences are used for evaluation. For the KITTI-T \citep{DBLP:conf/cvpr/GeigerLU12} dataset, the first $13$ sequences are used to train the vehicle detection and tracking models, and the remaining $8$ sequences are used for evaluation.}. While existing performance evaluation protocols use a single predefined setting of object detection to compare different object tracking methods, our experimental results (see Figure \ref{fig:pr-mota-curves}) show that the performance (\eg, relative rankings of different methods) of MOT systems vary significantly using different settings for object detection. For example, as shown in Figure \ref{fig:pr-mota-curves}(a), the CEM tracker obtains higher MOTA score than the DCT tracker at the precision-recall values $(0.433, 0.341)$, but lower MOTA score at the precision-recall values $(0.674, 0.183)$. Similar results are observed for other trackers in the MOT16 \citep{DBLP:journals/corr/Anton16} and KITTI-T \citep{DBLP:conf/cvpr/GeigerLU12} datasets. As such, using a single predefined setting of object detection is not sufficient to reveal the full behavior of the whole MOT systems and can lead to uninformative evaluations and conclusions.

\begin{figure*}[t]
\centering
\includegraphics[width=.95\textwidth]{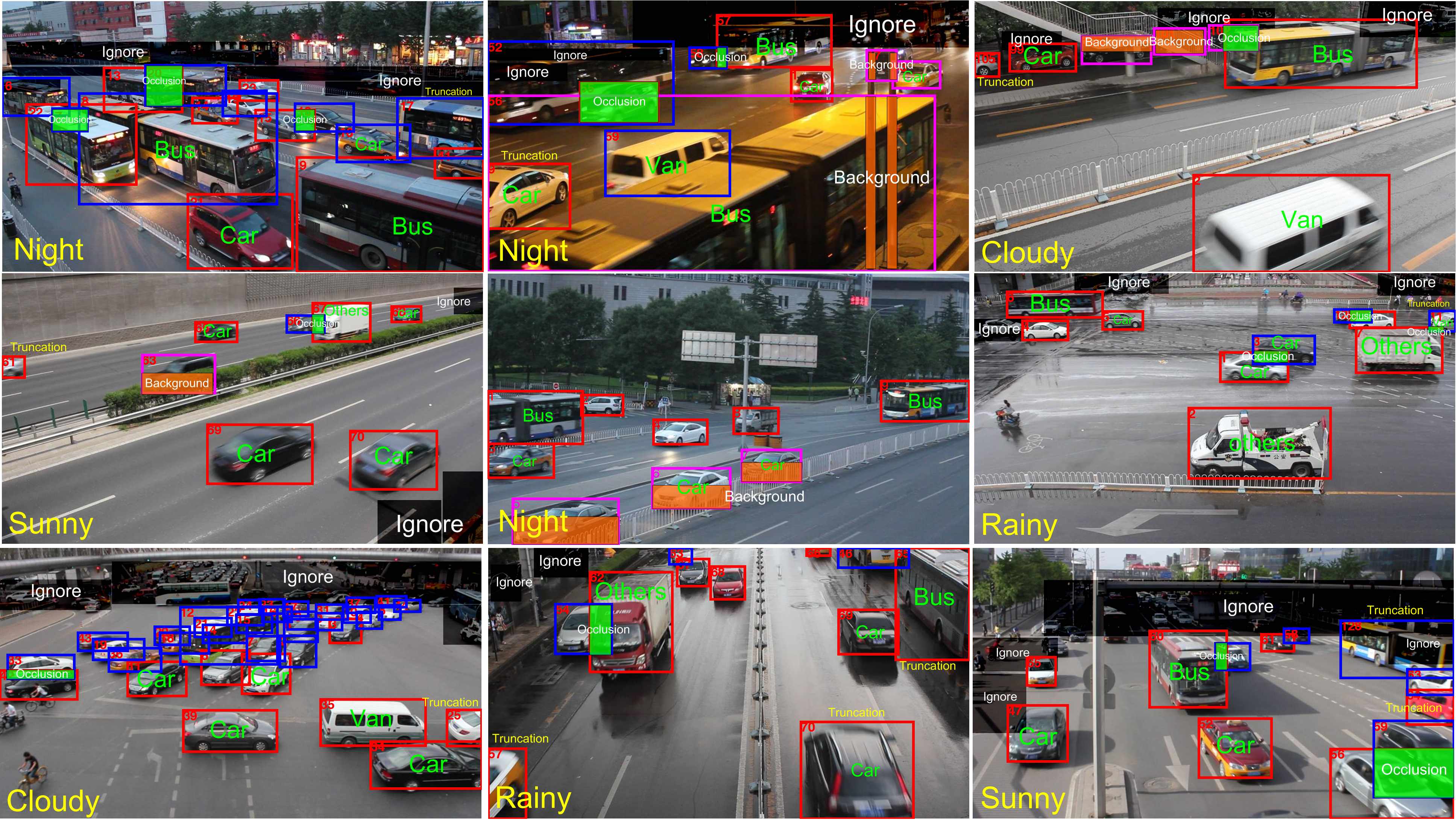}
\caption{Sample annotated frames in the \ds dataset. Bounding box colors indicate the occlusion level, as fully visible (red), partially occluded by other vehicles (blue), or partially occluded by background (pink). Black opaque areas are background regions that are not used in the benchmark dataset; green opaque areas are regions occluded by other vehicles; and orange opaque regions are areas occluded by background clutters. The illumination conditions are indicated by the texts in the bottom left corner of each frame.}
\label{fig:annotation}
\end{figure*}

Based on these observations, we propose a new evaluation protocol and metrics for MOT. The proposed \ds protocol considers the effect of object detection from the perspective of system evaluation. One recent work \citep{solera2015_413} also addresses the issue of MOT performance evaluation with a single predefined setting of detection results and suggests to use multiple perturbed ground truth annotations as detection inputs for analysis. However, evaluation with perturbed ground truth annotations does not reflect the performance of an object detector in practice. In contrast, our analysis is based on the actual outputs of the state-of-the-art object detectors with full range of precision-recall rates. From this perspective, our analysis and evaluation protocol reflect how a complete MOT system performs in practice. The main contributions of this work are summarized as follows. 1) We present a large scale \ds dataset for both vehicle detection and MOT evaluation, which differs from existing databases significantly in terms of data volume, annotation quality, and difficulty (see Table \ref{tab:object-tracking-detection-dataset}). 2) We propose a new protocol and evaluation metrics for MOT by taking the effect of object detection module into account. 3) Based on the \ds dataset and evaluation protocol, we thoroughly evaluate complete MOT systems by combining the state-of-the-art detection and tracking algorithms, and analyze the conditions under which the existing object detection and tracking methods may fail.

\section{\ds Benchmark Dataset}
\label{sec:ua-detrac-benchmark}
The \ds dataset consists of $100$ videos, selected from over $10$ hours of image sequences acquired by a Canon EOS 550D camera at $24$ different locations, which represent various traffic patterns and conditions including urban highway, traffic crossings and T-junctions. Notably, to ensure the diversity, we capture the data at different locations with various illumination conditions and shooting angles. The videos are recorded at $25$ frames per seconds (fps) with the JPEG image resolution of $960\times540$ pixels. A website\footnote{\url{http://detrac-db.rit.albany.edu}.} is constructed for performance evaluation of both detection and tracking methods on the \ds dataset using a submission protocol similar to that of the MOT15 benchmark dataset \citep{DBLP:journals/corr/Leal-TaixeMRRS15}.

\subsection{Data Collection and Annotation}
\label{sec:data-collection-annotation}
{\flushleft \textbf{Video Annotation.}}
More than $140,000$ frames in the \ds dataset are annotated with $8,250$ vehicles, and a total of $1.21$ million bounding boxes of vehicles are labeled. We ask over $10$ domain experts to annotate the collected data for more than two months. We carry out several rounds of cross-check to ensure high quality annotations. Similar to PASCAL VOC \citep{DBLP:journals/ijcv/EveringhamEGWWZ15}, there are some regions discarded in each frame, which cover vehicles that cannot be annotated due to low resolution. Figure \ref{fig:annotation} shows sample frames with annotated attributes in the \ds dataset.

The \ds dataset is divided into training (\ds-train) and testing (\ds-test) sets, with $60$ and $40$ sequences, respectively. We select training videos that are taken at different locations from the testing videos, but ensure the training and testing videos have similar traffic conditions and attributes. This setting reduces the chances of detection or tracking methods to overfit to particular scenarios. All the evaluated detection and tracking algorithms are trained on the \ds-train set and evaluated on the \ds-test set.

\begin{figure*}[t]
\centering
\includegraphics[width=.75\linewidth]{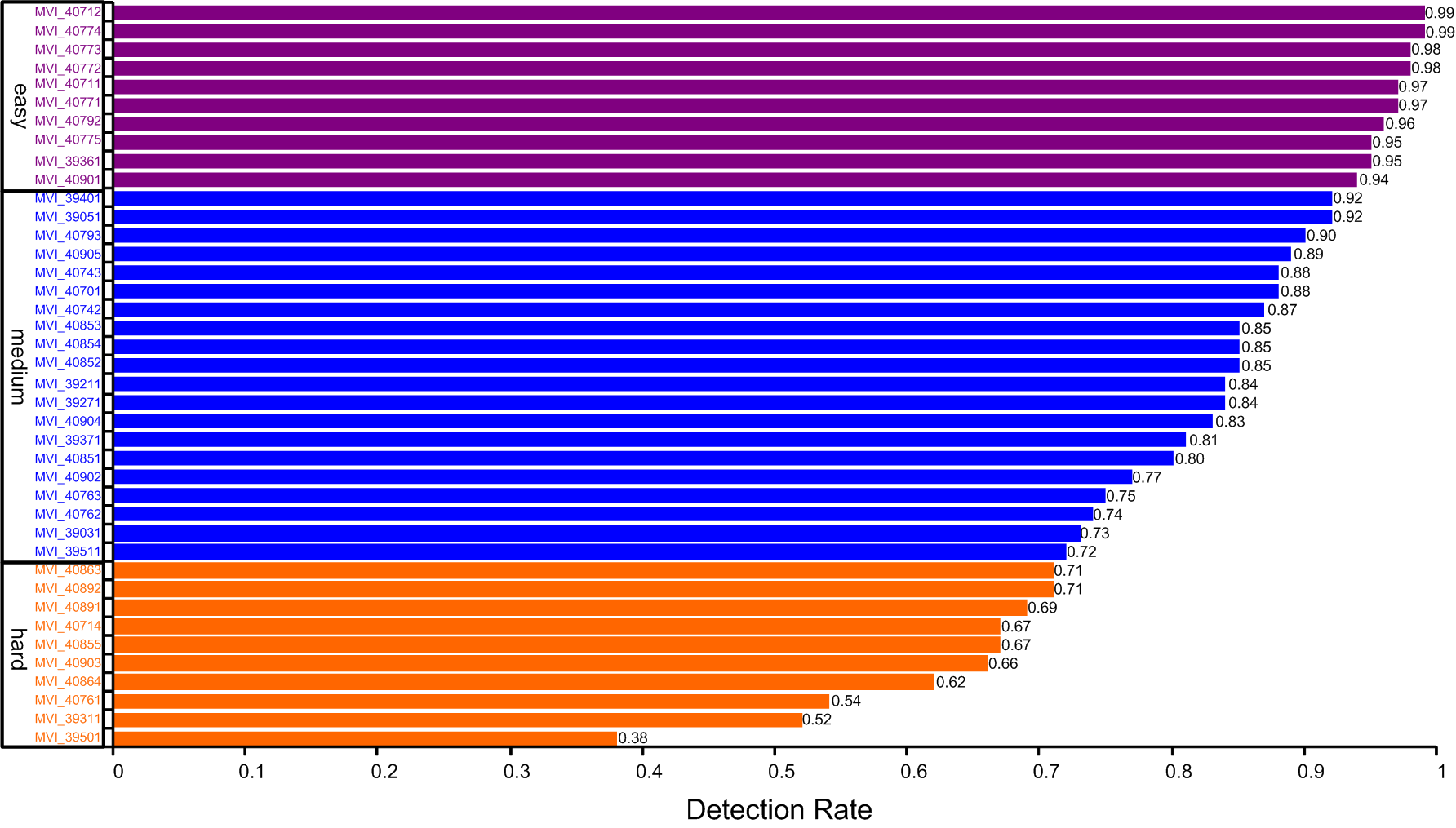}
\caption{Recall rates for different sequences in the \ds-test set. Each sequence is ranked in the descending order based on the recall rate of the EdgeBox method \citep{DBLP:conf/eccv/ZitnickD14} with the number of proposal fixed at $5,000$. The sequences of three levels of difficulties, \ie, {\bf {\em easy}}, {\bf {\em medium}}, and {\bf {\em hard}} are denoted in purple, blue, and orange, respectively.}
\label{fig:det-difficult}
\end{figure*}

The \ds dataset contains videos with large variations in scale, pose, illumination, occlusion and background clutters. For evaluation on object detection, similar to the KITTI detection \citep{DBLP:conf/cvpr/GeigerLU12} and WIDER FACE \citep{DBLP:conf/cvpr/YangLLT16} datasets, we define three levels of difficulties in the \ds-test set, \ie, {\em easy} ($10$ sequences), {\em medium} ($20$ sequences), and {\em hard} ($10$ sequences) based on the recall rate of the EdgeBox method \citep{DBLP:conf/eccv/ZitnickD14}. Figure \ref{fig:det-difficult} shows the distribution of the \ds-test set in terms of detection difficulty. The average recall rates of these three levels are $97.0\%$, $85.0\%$, and $64.0\%$, respectively, with $5,000$ proposals per frame.

\begin{figure*}[t]
\centering
\includegraphics[width=.9\linewidth]{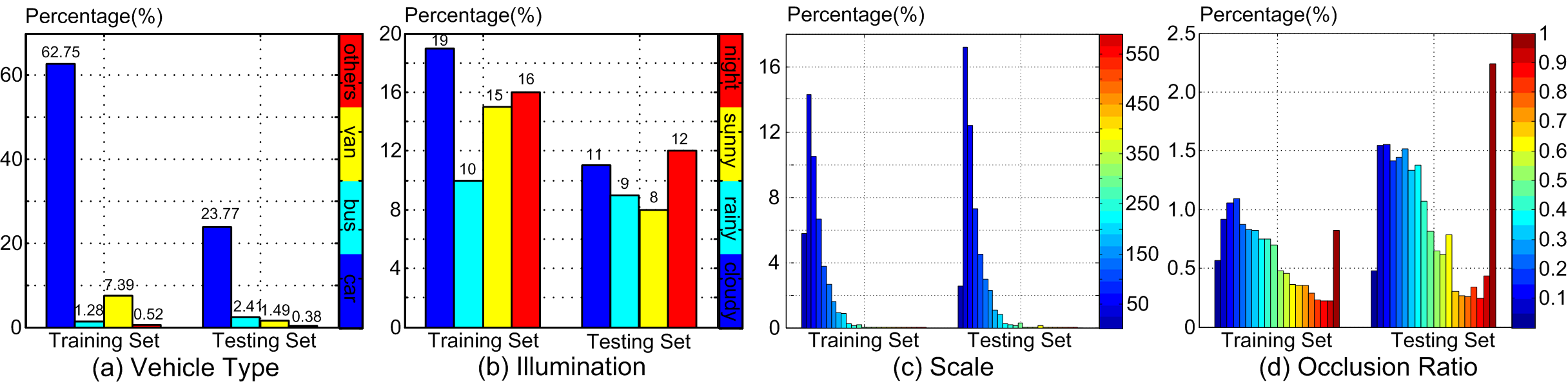}
\caption{Attribute statistics of the \ds benchmark dataset.}
\label{fig:attributes}
\end{figure*}

\begin{figure*}[t]
\centering
\includegraphics[width=.75\linewidth]{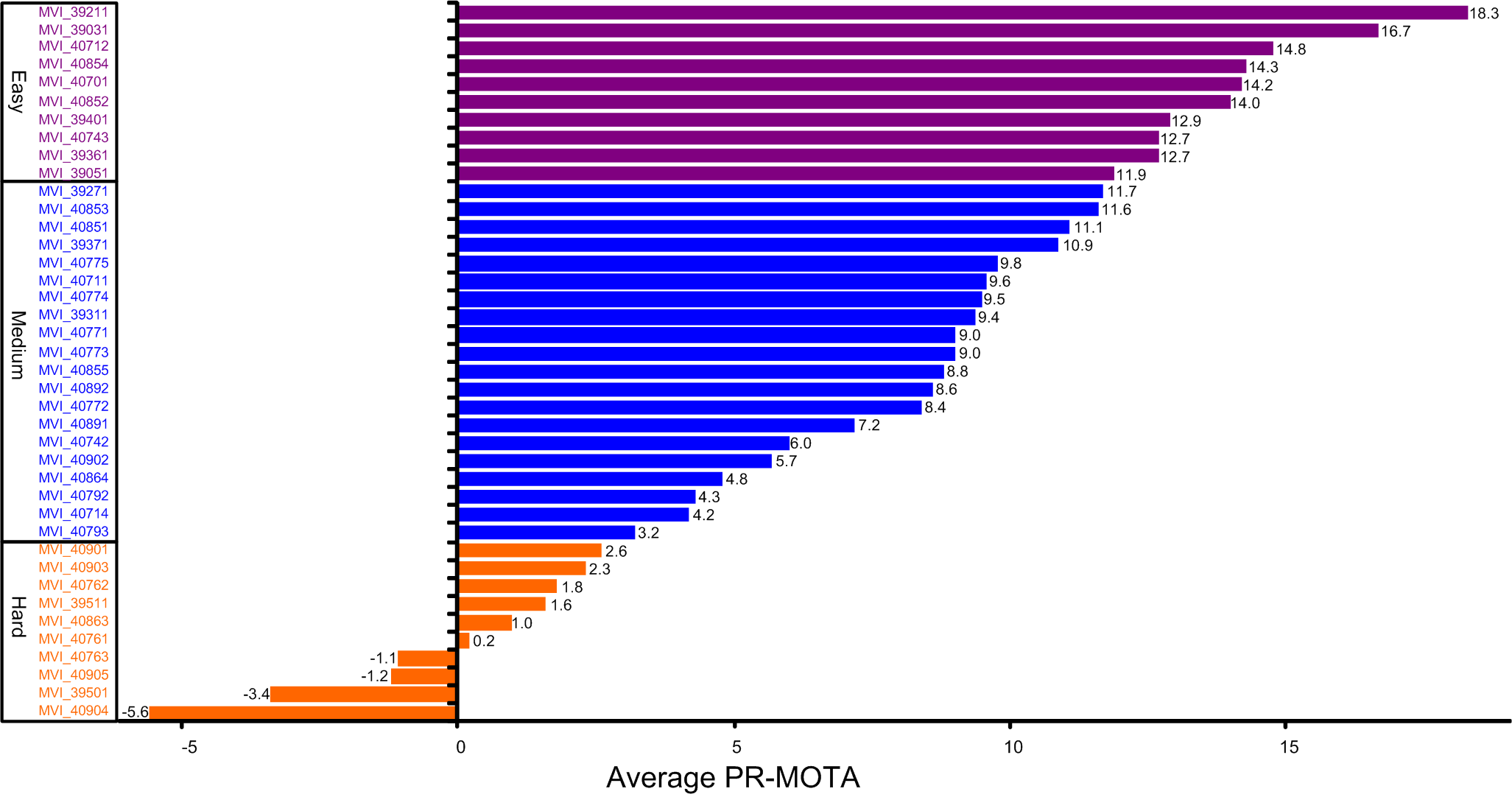}
\caption{Average PR-MOTA scores for different sequences in the \ds-test set. Each sequence is ranked in descending order based on average PR-MOTA score of the MOT systems constructed from combinations of six representative object tracking methods, \ie, GOG \citep{DBLP:conf/cvpr/PirsiavashRF11}, CEM \citep{DBLP:conf/cvpr/AndriyenkoS11}, DCT \citep{DBLP:conf/cvpr/AndriyenkoSR12}, IHTLS \citep{DBLP:conf/iccv/DicleCS13}, H\textsuperscript{2}T \citep{DBLP:conf/cvpr/Longyin14}, and CMOT \citep{DBLP:conf/cvpr/BaeY14}, and four representative object detection methods, \ie, DPM \citep{DBLP:journals/pami/FelzenszwalbGMR10}, ACF \citep{DBLP:journals/pami/DollarABP14}, R-CNN \citep{DBLP:conf/cvpr/GirshickDDM14}, and CompACT \citep{DBLP:conf/iccv/CaiMN15}. The sequences of three levels of difficulties, \ie, {\bf {\em easy}}, {\bf {\em medium}}, and {\bf {\em hard}} are denoted in purple, blue, and orange, respectively.
}
\label{fig:mot-difficult}
\end{figure*}

For MOT evaluation, we also define three levels of difficulties among the $40$ testing sequences, \ie, {\em easy} ($10$ sequences), {\em medium} ($20$ sequences), and {\em hard} ($10$ sequences) based on the average PR-MOTA score (defined in Section \ref{sec:protocol-MOT}) of the MOT systems constructed from combinations of six representative object tracking methods (\ie, GOG \citep{DBLP:conf/cvpr/PirsiavashRF11}, CEM \citep{DBLP:conf/cvpr/AndriyenkoS11}, DCT \citep{DBLP:conf/cvpr/AndriyenkoSR12}, IHTLS \citep{DBLP:conf/iccv/DicleCS13}, H\textsuperscript{2}T \citep{DBLP:conf/cvpr/Longyin14}, and CMOT \citep{DBLP:conf/cvpr/BaeY14}) and four representative object detection methods (\ie, DPM \citep{DBLP:journals/pami/FelzenszwalbGMR10}, ACF \citep{DBLP:journals/pami/DollarABP14}, R-CNN \citep{DBLP:conf/cvpr/GirshickDDM14}, and CompACT \citep{DBLP:conf/iccv/CaiMN15}). Figure \ref{fig:mot-difficult} shows the distribution of the \ds-test set in terms of tracking difficulty.

To analyze the performance of object detection and tracking algorithms thoroughly, we annotate sequences with several attributes:
\begin{itemize} 
  \item {\noindent \textbf{Vehicle type.}}
We annotate four types of vehicles as {\em car}, {\em bus}, {\em van}, and {\em others} (including other vehicle types such as trucks and tankers)\footnote{Some vehicles in our dataset are rarely occurring special vehicles, and the number of them are limited. To facilitate model training, we combine them together to form the ``other'' category.}. The distribution of vehicle type is shown in Figure \ref{fig:attributes}(a).
\item {\noindent \textbf{Illumination.}}
We consider four categories of illumination conditions, \ie, {\em cloudy}, {\em night}, {\em sunny}, and {\em rainy}. The distribution of video sequences based on illumination attribute is presented in Figure \ref{fig:attributes}(b).
  \item {\noindent \textbf{Scale.}}
We define the scale of the annotated vehicle bounding boxes as the square root of the area in pixels. The distribution of vehicle scale in the dataset is presented in Figure \ref{fig:attributes}(c). We label vehicles with three scales: {\em small scale} ($0$-$50$ pixels), {\em medium scale} ($50$-$150$ pixels), and {\em large scale} (more than $150$ pixels).
  \item {\noindent \textbf{Occlusion ratio.}}
We use the fraction of vehicle bounding box being occluded to define the occlusion ratio. We annotate the occlusion relations between vehicle bounding boxes and compute the occlusion ratio. Vehicles are annotated with three categories: {\em no occlusion}, {\em partial occlusion}, and {\em heavy occlusion}. Specifically, a vehicle is considered partially occluded if the occlusion ratio of a vehicle is in the range of $1\%$ to $50\%$, and heavily occluded if the occlusion ratio is larger than $50\%$. The distribution of occluded vehicles in videos is shown in Figure \ref{fig:attributes}(d).
  \item {\noindent \textbf{Truncation ratio.}}
The truncation ratio indicates the degree of vehicle parts appears outside a frame\footnote{Note that it is difficult to annotate the outside region of objects accurately. We thus approximately estimate the outside regions of objects by referring the last complete bounding boxes of objects before exiting the scenes. Meanwhile, the truncation ratio is only used to determine the validation of the annotated objects. That is, for both detection and tracking, we only consider the objects with the ratio smaller than $50\%$ in training. Thus, the annotation errors in truncation ratio have little impact on the benchmark.}. If a vehicle is not fully captured within a frame, we annotate the bounding box across the frame boundary and compute the truncation ratio based on the region outside the image. The truncation ratio is used in the training process of the evaluated detection and tracking algorithms where we discard the annotated bounding box when its ratio is larger than $50\%$.
\end{itemize}

\subsection{Relevance to Existing Benchmark Datasets}

\subsubsection{Object Detection Datasets}
Numerous benchmark datasets, \eg, PASCAL VOC \citep{DBLP:journals/ijcv/EveringhamEGWWZ15}, ImageNet \citep{ILSVRC15}, Caltech \citep{DBLP:journals/pami/DollarWSP12}, KITTI-D \citep{DBLP:conf/cvpr/GeigerLU12}, and KAIST \citep{Hwang-2015-CVPR}, have been developed for object detection. These datasets are mainly developed for object detection in single images that can be used to train detectors for MOT systems. Recently, to facilitate the research in autonomous driving field, the Berkeley DeepDrive BDD100k \citep{DBLP:journals/corr/abs-1805-04687} contains over $100,000$ videos with annotations of image level tagging, object bounding boxes, drivable areas, lane markings, and full-frame instance segmentation. The Baidu Apolloscapes \citep{DBLP:conf/cvpr/HuangCGCZWLY18} provides high density 3D point cloud map, per-pixel, per-frame semantic image label, lane mark label, and semantic instance segmentation annotations. The Oxford Robotic Car \citep{DBLP:journals/ijrr/MaddernPLN17} is a autonomous driving dataset with approximate $20$ million images with LIDAR, GPS and INS ground-truth in all weather conditions.

\subsubsection{Object Tracking Datasets}
Several multi-object tracking benchmarks have also been collected for evaluating object tracking methods. Some of the most widely used multi-object tracking evaluation datasets include the PETS09 \citep{DBLP:conf/avss/FerrymanE09}, PETS16 \citep{website:pets2016/pets2016}, KITTI \citep{DBLP:conf/cvpr/GeigerLU12}, MOT15 \citep{DBLP:journals/corr/Leal-TaixeMRRS15} and MOT16 \citep{DBLP:journals/corr/Anton16}, and UAVDT \citep{DBLP:conf/eccv/DuQYYDLZHT18} datasets. The PETS09 and PETS16 datasets focus on multi-pedestrian detection, tracking as well as counting. The KITTI dataset is designed for object tracking and detection, which are acquired from a moving vehicle with viewpoint of the driver. The MOT15 dataset aims to provide a unified platform and evaluation protocol for object tracking. It includes a dataset of $22$ videos mostly from surveillance cameras where the targets of interest are pedestrians. In addition, it also provides a platform where new datasets and multi-object tracking methods can be incorporated in a plug-and-play manner. The MOT16 benchmark dataset is an extension of MOT15 with more challenging sequences and thorough annotations. Recently, the UAVDT dataset is proposed to advance object tracking algorithms applied in drone based scenes.

Compared to existing multi-object tracking datasets, the \ds benchmark dataset is designed for vehicle surveillance scenarios with significantly more video frames, annotated bounding boxes and attributes. The vehicles in the videos are acquired at different view angles and frequently occluded. Meanwhile, the \ds benchmark is designed for performance evaluation of both object detection and multi-object tracking. Table \ref{tab:object-tracking-detection-dataset} summarizes the differences between existing and proposed \ds benchmarks in various aspects.

\subsection{Object Detection Algorithms}
We review the state-of-the-art object detection methods in the context of MOT (\eg, humans, faces and vehicles), and describe $12$ evaluated algorithms in the \ds dataset.

\subsubsection{Review of Object Detection Methods}
\cite{DBLP:journals/ijcv/ViolaJ04} develop an adaptive boosting algorithm based on a cascade of classifiers and Haar-like feature to detect faces effectively and efficiently. To achieve robust performance, \cite{DBLP:conf/icb/ZhangCXLL07} propose to use the discriminative multi-block local binary pattern (MB-LBP) features, which capture more image structural information than Haar-like features, to represent face images. Gradient features are important cues for detection. \cite{DBLP:conf/cvpr/DalalT05} use the histogram of oriented gradients (HOG) to describe local dominant edge cues for pedestrian detection in single images. In addition, both optical flows and HOG features have been used to detect pedestrians in videos \citep{DBLP:conf/eccv/DalalTS06}. In \citep{DBLP:journals/pami/DollarABP14}, it has been demonstrated that significant performance gain for pedestrian can be achieved when features from multiple channels are used.

In contrast to hand-crafted features, \eg, Haar \citep{DBLP:journals/ijcv/ViolaJ04}, HOG \citep{DBLP:conf/cvpr/DalalT05}, and MB-LBP \citep{DBLP:conf/icb/ZhangCXLL07}, data-driven hierarchical features (\eg, CNN features \citep{DBLP:conf/nips/KrizhevskySH12}) have recently been shown to be effective in numerous vision tasks. The region-based CNN (R-CNN) method \citep{DBLP:conf/cvpr/GirshickDDM14} combines region proposals with convolutional neural networks (CNNs), and achieves better performance in the PASCAL VOC challenge \citep{DBLP:journals/ijcv/EveringhamEGWWZ15} than systems based on hand-crafted features. Recently, R-CNN are extended \citep{DBLP:journals/corr/Girshick15,DBLP:journals/pami/HeZR015} to attend to RoIs on feature maps using the RoIPool operation, achieving fast speed and better accuracy. Faster R-CNN \citep{DBLP:journals/pami/RenHG017} attempt to learn the attention mechanism with a region proposal network (RPN) to further improve the methods \citep{DBLP:journals/corr/Girshick15,DBLP:journals/pami/HeZR015}. \cite{DBLP:conf/iccv/CaiMN15} develop a boosting approach to solve the cascade learning problem of combining features of different complexities, which exploits features of high complexity in the later stages, where only a few difficult candidate patches remain to be classified. As such, the method performs well in terms of accuracy and speed. \cite{DBLP:conf/cvpr/RedmonDGF16} formulate object detection as a regression problem to efficiently predict bounding boxes and associate class probabilities from an image by a single neural network in one evaluation. To account for the scale issue in object detection, \cite{DBLP:conf/eccv/CaiFFV16} present a unified multi-scale CNN algorithm, where object detection is performed at multiple output layers with each focusing on objects within certain scale ranges.

Parts-based representations have been widely used in object detection. \cite{DBLP:journals/pami/FelzenszwalbGMR10} present the deformable part-based model (DPM), which describes the part positions as latent variables in the structural SVM framework, for object detection. They extend this strategy and develop a multi-resolution method operating as a deformable parts model and a rigid template to handle large and small objects, respectively. However, the DPM-based methods are computationally expensive for practical applications. To improve the efficiency, \cite{DBLP:conf/cvpr/YanLWL14} propose an algorithm by constraining the root filter to be low rank, designing a neighborhood-aware cascade to capture the dependence among regions for aggressive pruning, and constructing look-up tables to replace HOG feature extraction with simpler operations.

Considerable efforts have also been made to improve the quality of object proposals for object detection. Instead of scanning through regions sequentially for object detection, \cite{DBLP:conf/cvpr/LampertBH08} propose a branch-and-bound scheme to efficiently maximize the classifier function over all possible sliding windows based on bag-of-words image representation \citep{DBLP:conf/iccv/SivicZ03}. \cite{DBLP:conf/iccv/SandeUGS11} develop a selective search approach using over-segmented regions to generate limited number of possible object locations. \cite{DBLP:conf/eccv/ZitnickD14} exploit the number of contours in a bounding box to facilitate generating object proposals. \cite{DBLP:journals/pami/RenHG017} introduce a Region Proposal Network (RPN) that shares full-image convolutional features with the detection network, which enables region proposals efficiently.

\subsubsection{Evaluated Object Detectors}
We evaluate $12$ state-of-the-art object detection algorithms in this work using the \ds dataset\footnote{We use the original codes for evaluation of object detection.} including the DPM \citep{DBLP:journals/pami/FelzenszwalbGMR10}, ACF \citep{DBLP:journals/pami/DollarABP14}, R-CNN \citep{DBLP:conf/cvpr/GirshickDDM14}, CompACT \citep{DBLP:conf/iccv/CaiMN15}, Faster R-CNN \citep{DBLP:journals/pami/RenHG017}, EB \citep{DBLP:conf/icmcs/WangLWZYX17}, YOLOv3 \citep{DBLP:journals/corr/abs-1804-02767}, GP-FRCNN \citep{DBLP:conf/avss/AminG17}, CSP \citep{DBLP:conf/cvpr/LiuLRHY19}, HAT \citep{DBLP:journals/ijcv/WuKSC19}, FG-BR\_Net \citep{DBLP:journals/tip/FuCYJZ019} and RD2 \citep{DBLP:conf/cvpr/ZhangWBLL18} methods. Specifically, GP-FRCNN \citep{DBLP:conf/avss/AminG17} and RD2 \citep{DBLP:conf/cvpr/ZhangWBLL18} are the winners in the UA-DETRAC Challenges in 2017 \citep{DBLP:conf/avss/LyuCDWQLWKHCCAB17} and 2018 \citep{DBLP:conf/avss/LyuCDLWCCSMDCBG18}. We retrain these methods on the \ds-train set and evaluate the performance on the \ds-test set.

The DPM method is trained using a mixture of $3$ star-structured models where each one has $2$ latent orientations. The ACF cascade uses $2,048$ decision trees of depth $4$. For the CompACT scheme \citep{DBLP:conf/iccv/CaiMN15}, we train a cascade of $2,048$ decision trees of depth $4$. For the ACF and CompACT methods, the template size is set to $64\times{64}$ pixels. To detect vehicles with different aspect ratios, the original images are resized to six different aspect ratios before being scanned by the detectors such that only a single model is needed. A bounding box regression model based on the ACF features is trained for the ACF and CompACT detectors for better performance. For the R-CNN algorithm, we fine-tune AlexNet \citep{DBLP:conf/nips/KrizhevskySH12} on the \ds-train set. Instead of using the selective search method \citep{DBLP:journals/ijcv/UijlingsSGS13} to generate proposals\footnote{Since the selective search method is less effective in generating accurate region proposals of vehicles, we use the outputs of ACF as proposals \citep{DBLP:conf/cvpr/HosangOBS15} for the R-CNN method. We need to ensure the proposals generated by ACF with high recall. If a proposal is not generated in the vicinity of an object, it will not be detected by R-CNN.}, the output bounding boxes of the ACF method are warped to $227\times{227}$ pixels and then fed into the R-CNN model for classification. For the Faster R-CNN algorithm, we fine-tune the VGG-16 backbone \citep{DBLP:journals/corr/SimonyanZ14a} on the \ds-train set. We use the default $3$ scales and $3$ aspect ratios to set the anchors, and top-$2000$ ranked proposals generated by the region proposal network (RPN) are used to train the second stage Fast R-CNN. The positive samples are all annotated vehicles from the \ds-train set with less than $50\%$ occlusion and truncation ratios, and the KITTI-D dataset \citep{DBLP:conf/cvpr/GeigerLU12} is used for mining hard negatives. The minimum size of the detected object is set to $25\times{25}$ pixels for all detectors. 

The EB method \citep{DBLP:conf/icmcs/WangLWZYX17} is constructed by the pre-trained VGG-16 network \citep{DBLP:journals/corr/SimonyanZ14a} on the ImageNet classification dataset. 
The YOLOv3 \citep{DBLP:journals/corr/abs-1804-02767} scheme uses the Darknet-53 backbone to perform feature extraction. 
The GP-FRCNN \citep{DBLP:conf/avss/AminG17} model is an extension of the Faster R-CNN detector \citep{DBLP:journals/pami/RenHG017} by re-ranking the generic object proposals with an approximate geometric estimation of the scenes. 
The CSP \citep{DBLP:conf/cvpr/LiuLRHY19} approach uses the ResNet-50 network as the backbone, which is also pre-trained on the ImageNet dataset \citep{DBLP:conf/cvpr/DengDSLL009}. 
The HAT \citep{DBLP:journals/ijcv/WuKSC19} algorithm is constructed based on the VGG-16 model, and uses the LSTM method for category-specific attention with $128$ hidden cells. 
The FG-BR\_Net \citep{DBLP:journals/tip/FuCYJZ019} method uses the OMoGMF \citep{DBLP:journals/pami/YongMZZ18} model as the base block of the proposed background subtraction-recurrent neural network. 
The RD2 \citep{DBLP:conf/cvpr/ZhangWBLL18} scheme is a variant of the RefineDet \citep{DBLP:conf/cvpr/ZhangWBLL18} method with the Squeeze-and-Excitation Network (SENet) \citep{DBLP:conf/cvpr/HuSS18}. 
It averages the results of two detectors with different backbones, \ie, SEResNeXt-50 and ResNet-50.

\subsection{Object Tracking Algorithms}
We briefly review the multi-object tracking algorithms, and then describe ten state-of-the-art object tracking approaches evaluated in this work.

\subsubsection{Review of MOT Methods}
Early multi-object tracking methods formulate the task as the state estimation problem using Kalman \citep{DBLP:journals/tac/LevenL09,DBLP:conf/iccv/PellegriniESG09} and particle filters \citep{DBLP:journals/ijcv/IsardB98,DBLP:journals/pami/KhanBD05,DBLP:conf/cvpr/MikamiOY09,DBLP:conf/cvpr/YangLWYL14}. These methods typically predict the target states in short duration effectively but do not perform well in complex scenarios.

Recently, more effective multi-target tracking algorithms are developed based on the tracking-by-detection framework. Typically, detection results from consecutive frames are linked based on similarities in appearance and motion to form long tracks \eg, joint probabilistic data association filter (JPDAF) \citep{Fortmann83analgorithm} and multiple hypotheses tracking (MHT) \citep{Reid79analgorithm} methods. The JPDA method \citep{Fortmann83analgorithm} considers all possible matches between the tracked targets and detections in each time frame to compute the joint probabilistic score to complete the tracking task. However, with an increasing number of targets in the sequence, the computational complexity of the method becomes intractable. \cite{DBLP:conf/iccv/RezatofighiMZSD15} present a computationally tractable approximation to the original JPDA algorithm based on a recent method to find the $m$-best solutions of an integer linear program. The MHT method \citep{Reid79analgorithm} builds a tree of potential track hypotheses for each candidate target, and evaluates the likelihoods of the hypothesized matches over several time steps. To further improve MHT in exploiting higher-order information, \cite{DBLP:conf/iccv/KimLCR15} train
an online appearance model for each track hypothesis. By design, the MHT method is more effective than the JPDAF scheme for long-term association problem at the expense of computational loads.

Several algorithms consider associations of detection/tracklet pairs as an optimization task based on the K-shortest path \citep{DBLP:journals/pami/BerclazFTF11}, maximum weight independent sets \citep{DBLP:conf/cvpr/BrendelAT11}, maximum multi-clique optimization \citep{dehghan2015cvpr-gmmcp}, tensor power iterations \citep{DBLP:conf/cvpr/ShiLHYX14}, network flows \citep{DBLP:conf/cvpr/ZhangLN08,DBLP:conf/cvpr/PirsiavashRF11,DBLP:conf/cvpr/Leal-TaixeFKRS14}, linear programs \citep{DBLP:conf/cvpr/JiangFL07}, Hungarian algorithm \citep{DBLP:conf/cvpr/BaeY14}, generalized linear assignment optimization \citep{DBLP:conf/iccv/DicleCS13}, and subgraph decomposition \citep{DBLP:conf/cvpr/TangAAS15}. To exploit motion information of targets, \cite{DBLP:conf/cvpr/Longyin14} formulate the multi-object tracking task as exploring dense structures on a hypergraph, whose nodes are detections and hyper-edges describe the corresponding high-order relations. The run-time bottleneck of \citep{DBLP:conf/cvpr/Longyin14} is addressed in \citep{DBLP:journals/pami/WenLLLL16} using a RANSAC approach to extract the dense structures on hypergraph efficiently. \cite{DBLP:conf/cvpr/AndriyenkoS11} formulate multi-object tracking as an energy minimization problem by using physical constraints such as target dynamics, mutual exclusion, and track persistence. \cite{DBLP:conf/cvpr/YamaguchiBOB11} develop an agent-based behavioral model of pedestrians to improve tracking performance, which predicts human behavior based on an energy minimization problem. \cite{DBLP:conf/cvpr/AndriyenkoSR12} tackle multi-object tracking as a discrete-continuous optimization problem that integrates data association and trajectory estimation in an energy function in a way similar to \citep{DBLP:journals/ijcv/DelongOIB12}.

\subsubsection{Evaluated Object Trackers}
Using the \ds dataset, we evaluate performance of MOT systems constructed by different combinations of $6$ state-of-the-art object detection algorithms including DPM \citep{DBLP:journals/pami/FelzenszwalbGMR10}, ACF \citep{DBLP:journals/pami/DollarABP14}, R-CNN \citep{DBLP:conf/cvpr/GirshickDDM14}, CompACT \citep{DBLP:conf/iccv/CaiMN15}, Faster R-CNN \citep{DBLP:journals/pami/RenHG017}, and EB \citep{DBLP:conf/icmcs/WangLWZYX17}, and $10$ object tracking approaches including GOG \citep{DBLP:conf/cvpr/PirsiavashRF11}, CEM \citep{DBLP:conf/cvpr/AndriyenkoS11}, DCT \citep{DBLP:conf/cvpr/AndriyenkoSR12}, IHTLS \citep{DBLP:conf/iccv/DicleCS13}, H\textsuperscript{2}T \citep{DBLP:conf/cvpr/Longyin14}, CMOT \citep{DBLP:conf/cvpr/BaeY14}, TBD \citep{DBLP:journals/pami/GeigerLWSU14}, MHT \citep{DBLP:conf/iccv/KimLCR15}, IOU \citep{DBLP:conf/avss/BochinskiES17} and KIOU \citep{DBLP:conf/avss/LyuCDLWCCSMDCBG18}. All codes of the object detection and tracking algorithms are publicly available or provided by the authors of the corresponding publications. All these methods take object detection results in each frame as inputs and generate target trajectories to complete tracking task. We use the \ds-train set to determine the parameters of these methods empirically\footnote{We use the grid search of one parameter over a range of values while keep other parameters fixed. For each setting of parameters of the tracker, we generate a PR-MOTA curve and compute the corresponding PR-MOTA score. Then, we determine the parameters of the tracker corresponding to the maximum PR-MOTA score.}, and the \ds-test set for performance evaluation.

\section{\ds Evaluation Protocol}
\label{sec:evaluation-setup}
As discussed in Section \ref{sec:intro}, existing multi-object tracking evaluation protocols that use a single predefined object detection setting as input may not reflect the complete MOT performance well. In this section, we introduce the evaluation protocol for object detection and MOT that better reveal complete performance.

\subsection{Evaluation Protocol for Object Detection}
{\flushleft \textbf{Evaluation metric.}}
We generate the full {\em precision vs. recall} (PR) curve for each object detection algorithm. The PR curve is generated by varying the score threshold of an object detector to generate different precision and recall values. Per-frame detector evaluation is performed as in the KITTI-D benchmark \citep{DBLP:conf/cvpr/GeigerLU12}, where the hit/miss threshold of the overlap between a pair of detected and ground truth bounding boxes is set to $0.7$.

{\flushleft \textbf{Ranking detection methods.}}
The {\em average precision} (AP) score of the PR curve is used to rank the performance of a detector. We follow the strategy in the PASCAL VOC challenge \citep{DBLP:journals/ijcv/EveringhamEGWWZ15} to compute the AP score, \ie, calculate the average precisions at the fixed $11$ recall values from $0$ to $1$: $\{0, 0.1, 0.2, \cdots, 0.9, 1.0\}$.

\subsection{Evaluation Protocol for Object Tracking}
\label{sec:protocol-MOT}
{\flushleft \textbf{Existing evaluation metric.}}
We first introduce a set of performance evaluation metrics for object tracking widely used in the literature including {\em mostly tracked} (MT), {\em mostly lost} (ML), {\em identity switches} (IDS), {\em fragmentations of target trajectories} (FM), {\em false positives} (FP), {\em false negatives} (FN), and two CLEAR MOT metrics \citep{DBLP:conf/clear/StiefelhagenBBGMS06}, {\em multi-object tracking accuracy} (MOTA) as well as {\em multi-object tracking precision} (MOTP). The FP metric describes the number of false alarms by a tracker, and FN is the number of targets missed by any tracked trajectories in each frame. The IDS metric describes the number of times that the matched identity of a tracked trajectory changes, while FM is the number of times that trajectories are disconnected. Both IDS and FM metrics reflect the accuracy of tracked trajectories. The ML and MT metrics measure the percentage of tracked trajectories less than $20\%$ and more than $80\%$ of the time span based on the ground truth respectively. The MOTA metric for all sequences in the benchmark is defined by \citep{solera2015_413},
\begin{equation}
\label{equ_mota}
\textrm{MOTA} = 100\cdot(1-\frac{\sum_{v}\sum_t{(\textrm{FN}_{v,t}+\textrm{FP}_{v,t}+\textrm{IDS}_{v,t})}}{\sum_{v}\sum_t{\textrm{GT}_{v,t}}})[\%],
\end{equation}
where $\textrm{FN}_{v,t}$ is the number of false negatives, and $\textrm{FP}_{v,t}$ is the number of false positives at time index $t$ of sequence $v$, with the hit/miss threshold of the bounding box overlap between an output trajectory and the ground truth set to be $0.7$. In addition, $\textrm{IDS}_{v,t}$ is the number identity switches of a trajectory, and $\textrm{GT}_{v,t}$ is the number of ground truth objects. The MOTP metric is the average dissimilarity between all true positives and the corresponding ground truth targets, as the average overlap between all correctly matched hypotheses and respective objects. We note that the MOTA score is computed by the FN, FP and IDS scores of the tracking results.

\begin{figure}[t]
\centering
\includegraphics[width=.95\linewidth]{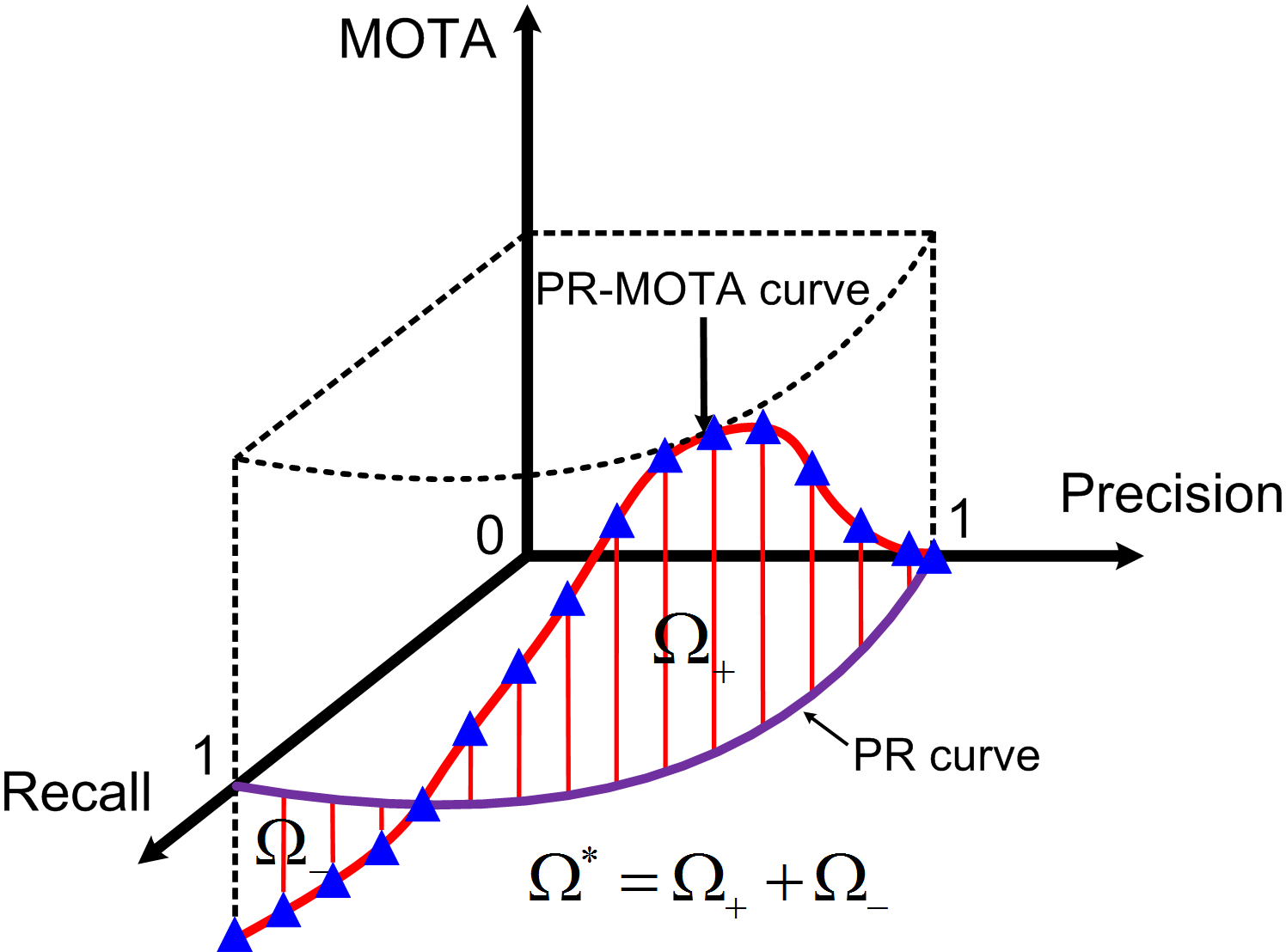}
\caption{Proposed \ds metric $\Omega^\ast$ of the PR-MOTA curve: the purple curve is the precision-recall curve describing the performance of object detection and the red one is the PR-MOTA curve. The blue triangles represent the sampling points used to generate the PR-MOTA curve.}
\label{fig:combine-evaluation}
\end{figure}

{\flushleft \textbf{Proposed evaluation metric.}}
In this work, we show it is necessary to consider the effect of detection performance on MOT evaluation and introduce the \ds metrics\footnote{Notably, it is not sufficient to compare two MOT systems based on MOTA scores that generated by two settings of input detection with different FN and FP values. Similar to the case in object detection, it is not meaningful to compare the performance of two detectors based on the different points on the PR curves. Thus, the maximum value on the PR-MOTA curve is not a good choice to compare the performance of the trackers. We expect a tracker achieving better performance as it can generate better performance with any different settings of detections. In this work, we use the average MOTA score over the PR curve of a detector, \ie, PR-MOTA, for comparison.}, \ie, the PR-MOTA, PR-MOTP, PR-MT, PR-ML, PR-IDS, PR-FM, PR-FP, and PR-FN scores, to take the effect of object detection into account, based on the basic evaluation metrics. First, we take the basic evaluation metric MOTA as an example to describe the PR-MOTA score. The PR-MOTA curve (see Figure \ref{fig:combine-evaluation}) is a three-dimensional curve characterizing the relationship between object detection (precision and recall) and tracking (MOTA). In the following, we describe the steps to generate a PR-MOTA curve and calculate the score:
\begin{enumerate}
 \item We first vary the detection threshold\footnote{Specifically, we vary the threshold $10$ times with equal interval from the minimal to the maximum detection scores to generate the PR-MOTA curve.} gradually to generate different object detection results (bounding boxes) corresponding to different values of precision $p$ and recall $r$. The two-dimensional curve corresponding to $(p,r)$ is the {\em precision-recall} (PR) curve ${\cal C}$ that delineates the region of possible PR values of a detector.
\item For a set of detection results determined by $(p,r)$, we apply an object tracking algorithm and compute the resulting MOTA score $\Psi(p, r)$.
The MOTA scores for $(p,r)$ values on the PR curve form a three-dimensional curve, \ie, the PR-MOTA curve, as shown in Figure \ref{fig:combine-evaluation}.
 \item From the PR-MOTA curve, we compute the integral score $\Omega^\ast$ to measure MOT performance, \ie, the PR-MOTA score $\Omega^\ast = \frac{1}{2}\int_{{\cal C}} \Psi(p, r) \mathrm{d}{\bf s}$\footnote{Note that we have $\int_{{\cal C}} \Psi(p, r) \mathrm{d}{\bf s} \in (-\infty, 200\%]$.
The proof can be found in the appendix.
To convert it to percentage, we multiply it by $\frac{1}{2}$ to ensure the PR-MOTA score is within the range of $(-\infty, 100\%]$. As it is difficult to compute the integration directly, we approximate it with additions over sampled locations over the PR curve ${\cal C}$.} ($\Omega^\ast$ is the line integral along the PR curve ${\cal C}$).
In other words, the PR-MOTA score $\Omega^\ast$ corresponds to the (signed) area of the curved surface formed by the PR-MOTA curve along the PR curve, as shown by the shaded area in Figure \ref{fig:combine-evaluation}.
\end{enumerate}
Using the PR-MOTA score, we can compare different multi-object tracking algorithms by considering the effect of detection modules. The scores of other seven metrics, \eg, PR-MOTP and PR-IDS, are similarly computed.

\begin{figure}[t]
\centering
\includegraphics[width=\linewidth]{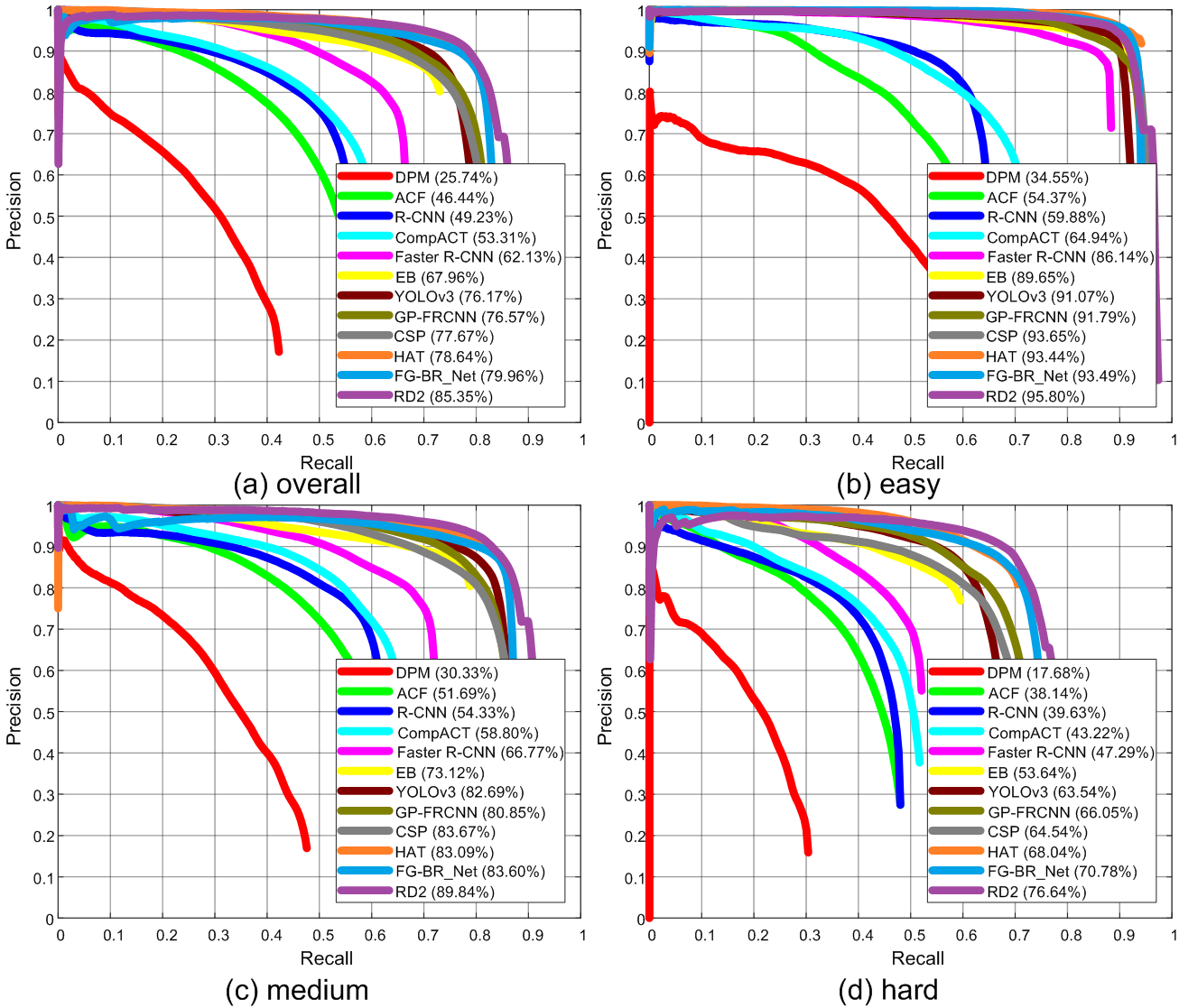}
\caption{Precision {\em vs.} recall curves of the detection algorithms in {\bf {\em overall/easy/medium/hard} subsets} of \ds benchmark dataset. The scores in the legend are the AP scores for evaluating the performance of object detection algorithms.}
\label{fig:det-p-1}
\end{figure}

\begin{figure}[t]
\centering
\includegraphics[width=\linewidth]{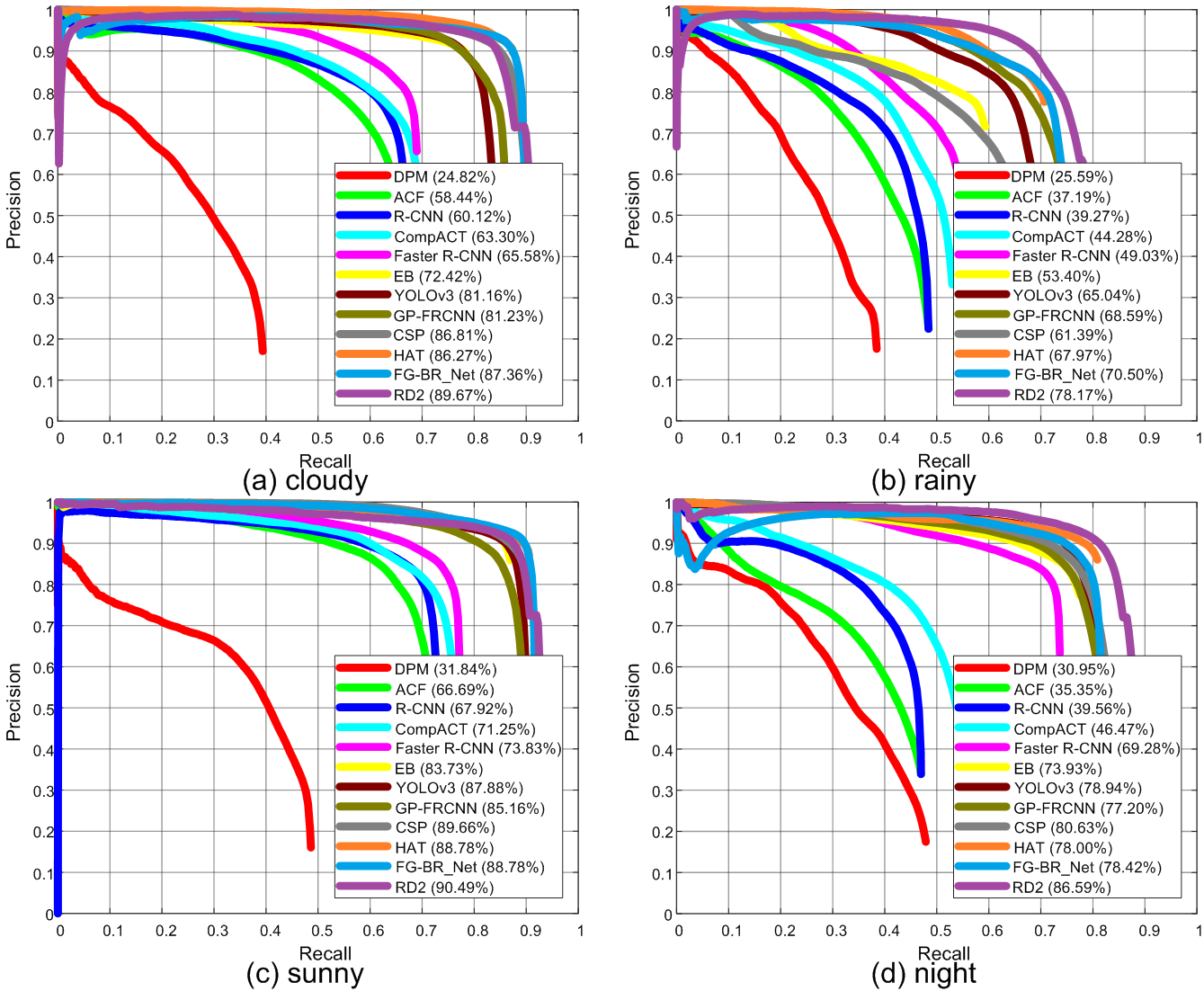}
\caption{Precision {\em vs.} recall curves of the detection algorithms in {\bf {\em cloudy/rainy/sunny/night} subsets} of \ds benchmark dataset. The scores in the legend are the AP scores for evaluating the performance of object detection algorithms.}
\label{fig:det-p-2}
\end{figure}

\begin{figure}[t]
\centering
\includegraphics[width=\linewidth]{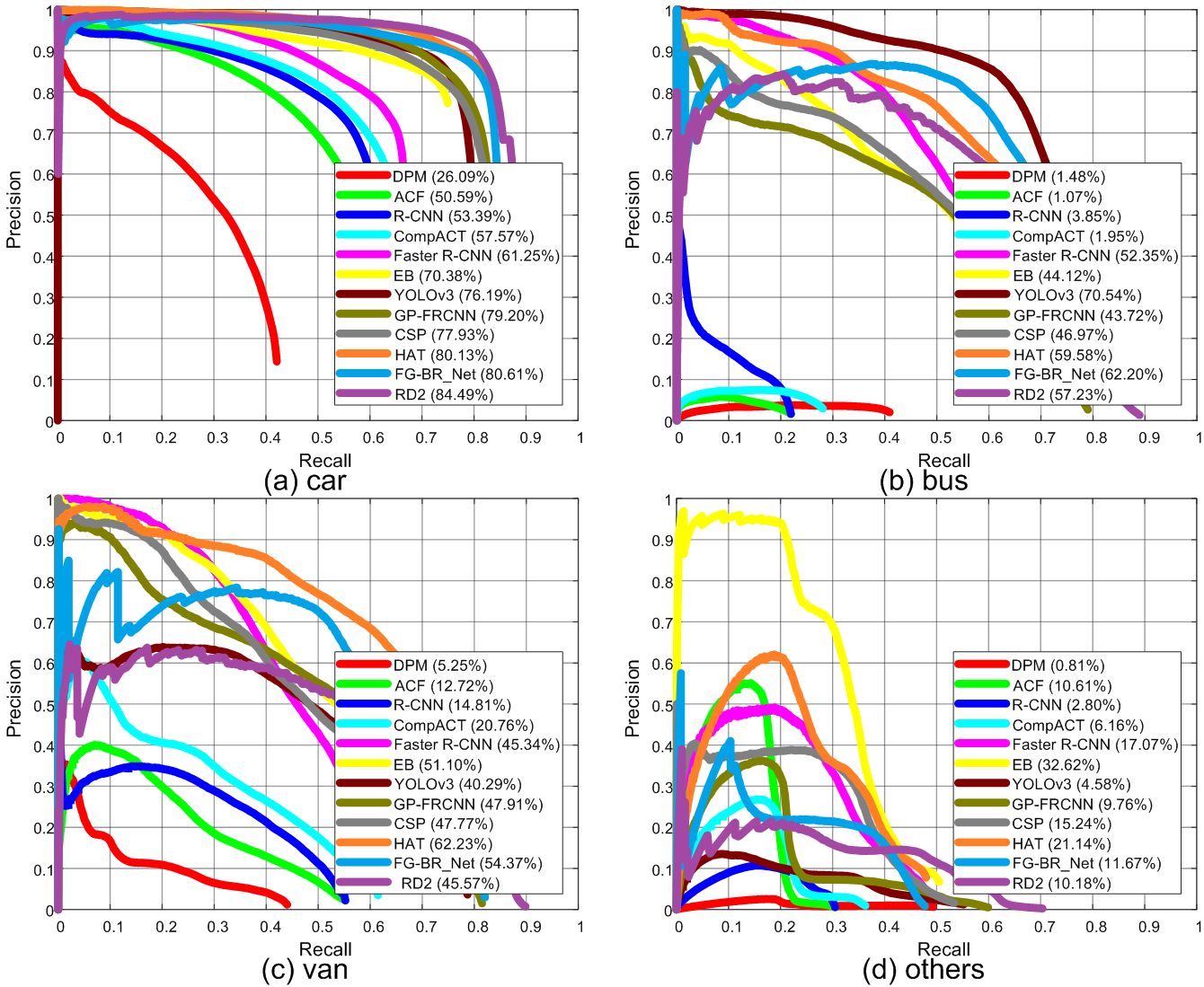}
\caption{Precision {\em vs.} recall curves of the detection algorithms in {\bf car/bus/van/others subsets} of \ds benchmark dataset. The scores in the legend are the AP scores for evaluating the performance of object detection algorithms.}
\label{fig:det-p-3}
\end{figure}

{\flushleft \textbf{Ranking MOT methods.}}
We rank the performance of MOT methods based on the PR-MOTA scores (larger PR-MOTA score indicates higher rank). If the PR-MOTA scores of two MOT methods are the same, we rank them based on the PR-MOTP scores (larger PR-MOTP score indicates higher rank).

\subsection{Comparisons with Existing Evaluation Protocols}
It has been shown recently \citep{DBLP:conf/cvpr/MilanSR13a} that the widely used MOT evaluation metrics including the MOTA, MOTP or IDS scores \citep{DBLP:conf/clear/StiefelhagenBBGMS06,DBLP:conf/cvpr/LiHN09}, do not fully reveal how a MOT system performs. Furthermore, there are several issues associated with the MOT evaluation protocols. Early studies \citep{DBLP:conf/eccv/ZamirDS12,DBLP:conf/cvpr/AndriyenkoSR12} use different object detection methods for evaluating MOT methods. It is well known that detection results affect the performance of MOT methods significantly. Most recent evaluation methods (\eg, \citep{DBLP:journals/pami/HuangLN13,DBLP:conf/cvpr/Longyin14,DBLP:journals/corr/Leal-TaixeMRRS15,DBLP:journals/corr/Anton16}) adopt a protocol that uses the same predefined setting of detection results to evaluate MOT methods, in order to make the evaluation independent of variations of detection results. It has been shown in \citep{solera2015_413} that the performance of MOT systems cannot be clearly reflected with a predefined setting of detection inputs, and multiple synthetic detections generated by controlled noise are used for comparisons. However, these synthetically generated detection results do not fully correspond to how detectors perform in real world scenarios. In addition, in \citep{solera2015_413}, the detections are randomly perturbed independently for each frame, which is different from how real detectors operate. In contrast, the \ds protocol considers the complete performance of a detector for MOT evaluation. Using the three-dimensional curve of detection (PR) and tracking scores (\eg, MOTA and MOTP), the \ds protocol can better reflect the behavior of the whole MOT systems.

\section{Analysis and Discussion}
\subsection{Object Detection}
{\flushleft \textbf{Overall performance.}}
The results of five state-of-the-art object detectors on the \ds dataset, shown in Figure \ref{fig:det-p-1}(a) with the PR curves, indicate that there remains much room for improvement for object detection algorithms. Specifically, the DPM and ACF methods do not perform well on vehicle detection with only $25.74\%$ and $46.44\%$ AP scores respectively. The R-CNN algorithm performs slightly better than the ACF method with AP score of $49.23\%$. The CompACT algorithm achieves more accurate results with $53.31\%$ AP score than the aforementioned methods by learning complexity-aware cascades. 
The recent proposed detectors achieve more than $60\%$ AP scores, \ie, Faster R-CNN with $62.13\%$ AP score, YOLOv3 with $76.17\%$ AP score, and RD2 with $85.35\%$ AP score. As shown in Figure \ref{fig:det-p-1}(b)-(d), from the {\em easy} to {\em hard} subsets, the AP scores of detectors drop $15$ to $20\%$. 
For example, the best detector RD2 only achieves $76.64\%$ AP score on the {\em hard} subset, which demonstrates that more effective detectors are needed for the challenging scenarios in the \ds dataset.

{\flushleft \textbf{Illumination.}} Most detectors are developed based on the assumption that objects can be spotted in the scenes with poor lighting conditions.
Figure \ref{fig:det-p-2} shows that all methods achieve AP scores below $80\%$ in the {\em rainy} and {\em night} scenes, except for the RD2 algorithm, which achieves $86.59\%$ AP in the {\em night} scene. 
In contrast, object detectors perform relatively well in other scenes with better lighting conditions. For example, the RD2 detector achieves $89.67\%$ and $90.49\%$ AP scores in the {\em cloudy} and {\em sunny} days, respectively.

\begin{figure*}[t]
\centering
\includegraphics[width=.85\linewidth]{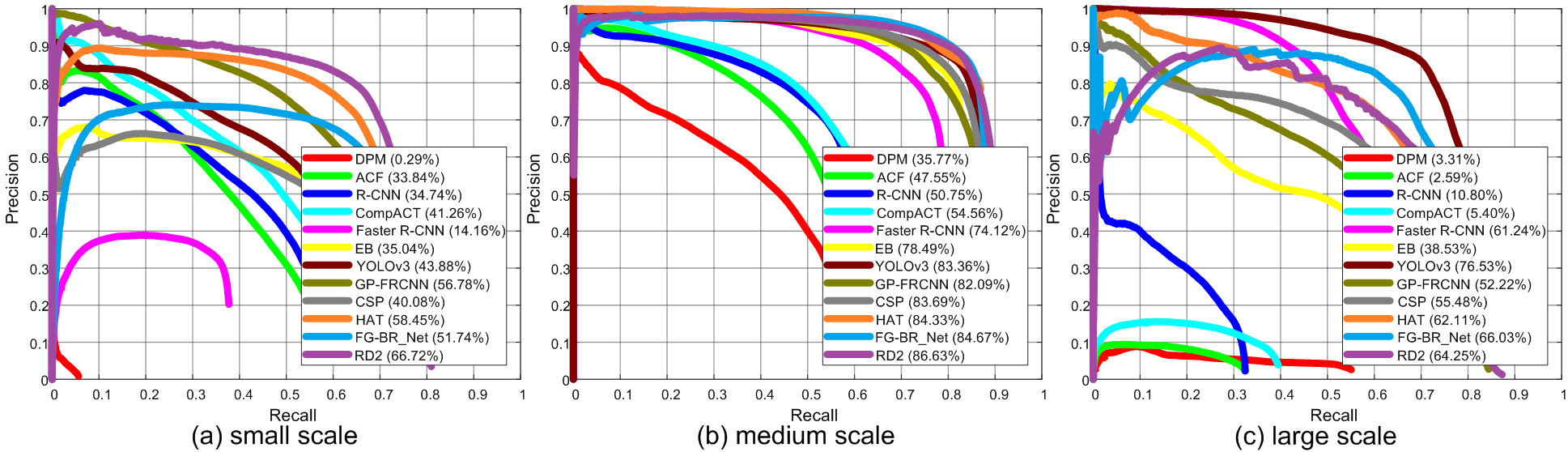}
\caption{Precision {\em vs.} recall curves of the detection algorithms in {\bf {\em small/medium/large} scale subsets} of \ds benchmark dataset. The scores in the legend are the AP scores for evaluating the performance of object detection algorithms.}
\label{fig:det-p-4}
\end{figure*}

\begin{figure*}[t]
\centering
\includegraphics[width=.85\linewidth]{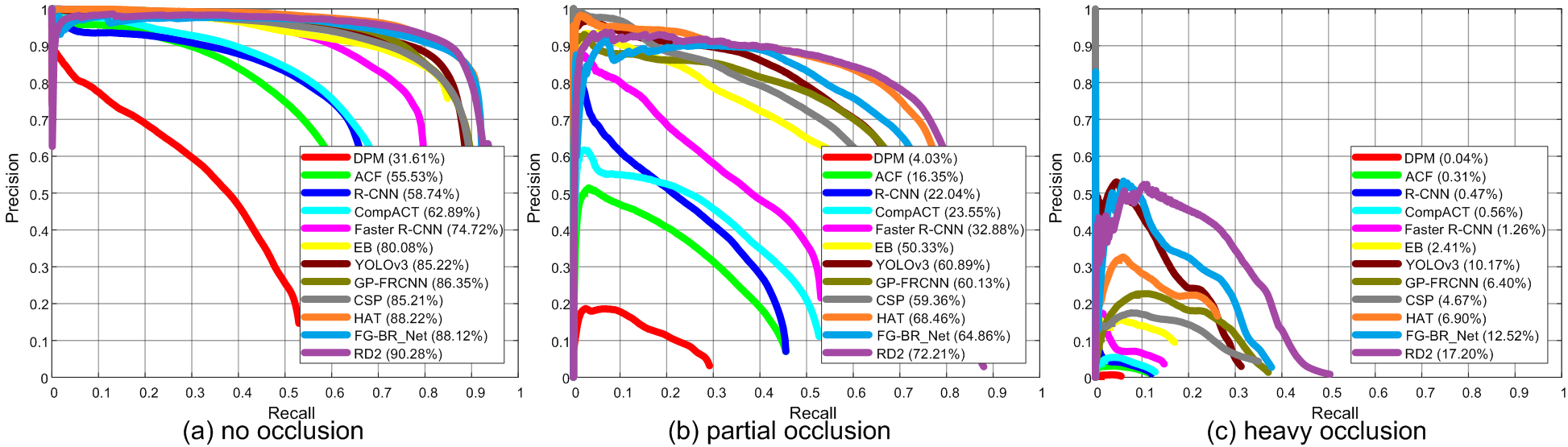}
\caption{Precision {\em vs.} recall curves of the detection algorithms in {\bf no/partial/heavy occlusion subsets} of \ds benchmark dataset. The scores in the legend are the AP scores for evaluating the performance of object detection algorithms.
}
\label{fig:det-p-5}
\end{figure*}

{\flushleft \textbf{Vehicle type.}} As shown in Figure \ref{fig:det-p-3}(a)-(d), the detectors perform relatively well only on cars among all kinds of vehicles, \eg, the RD2 method achieves $84.49\%$ AP score. 
It is worth mentioning that the AP score of the Faster R-CNN method is only $17.07\%$ in terms of the \textit{other} category, better than the other detectors including the RD2 method.
The reason can be that the region proposal network (RPN) is effective in reducing 
the number of candidate object locations and filtering out most background proposals to address the class imbalance issue. 
Moreover, the poor results can be attributed to two factors. First, it is difficult to handle large variations of scale and aspect ratio for different vehicles. Second, the limited amount of training samples affects the performance of object detectors, \ie, only $0.52\%$ vehicles are in the \textit{others} category in the training set, see Figure \ref{fig:attributes}(a).

{\flushleft \textbf{Scale.}} Figure \ref{fig:det-p-4} shows the detection results for each scale of vehicles in the \ds-test set. For {\em small} scale vehicles, most detectors achieve over AP scores of $30\%$ except the DPM method, where the RD2 method obtains the best AP score of $66.72\%$. 
The CSP approach employs the anchor-free framework to represent objects by points, resulting in inferior AP score of $40.08\%$. 
For {\em medium} scale vehicles, the RD2 method obtains the best AP score of $86.63\%$, which benefits from the proposed anchor refinement module to filter out negative anchors and refine positive anchors to provide better initialization for the location regression. The YOLOv3 method achieves the best AP score of $76.53\%$ for large scale vehicles. These results show that more effective detectors need to be developed to deal with small scale vehicles.

{\flushleft \textbf{Occlusion ratio.}} Figure \ref{fig:det-p-5} shows the effect of occlusion on detection performance
in three categories, \ie, no occlusion, partial occlusion, and heavy occlusion, as described in Section \ref{sec:data-collection-annotation}. When partial occlusion occurs (occlusion ratio is between $1\%-50\%$), the performance of detectors drops significantly (more than $15\%$ AP score). Furthermore, when heavy occlusion occurs (occlusion ratio is over $50\%$), the AP scores of all detectors are less than $20\%$. Significant efforts need to be made for vehicle detection under heavy occlusions.

\begin{table}[t]
  \centering
  \footnotesize
  \caption{Comparison results based on the MOTA/MOTP and PR-MOTA/MOTP metrics.}
  \renewcommand{\arraystretch}{0.5}
  \setlength{\tabcolsep}{5pt}
    \begin{tabular}{cccccc}
      \toprule
    Detection & Tracking & PR-MOTA & MOTA  & PR-MOTP & MOTP \\
    \midrule
    EB    & KIOU  & 21.1  & 62.1  & 28.6  & 81.9 \\
    \midrule
    Faster R-CNN & MHT   & 14.5  & 58.2  & 32.5  & 78.4 \\
   \midrule
    CompACT & GOG   & 14.2  & 44.4  & 37.0  & 80.8 \\
    \midrule
    R-CNN & DCT   & 11.7  & 38.4  & 38.0  & 80.6 \\
    \midrule
    ACF   & GOG   & 10.8  & 35.7  & 37.6  & 80.3 \\
    \midrule
    DPM   & GOG   & 5.5   & 26.2  & 28.2  & 76.2 \\
    \bottomrule
    \end{tabular}%
  \label{tab:comparison}%
\end{table}%

\begin{table*}[t]
  \centering
  \caption{PR-MOTA, PR-MOTP, PR-MT, PR-ML, PR-IDS, PR-FM, PR-FP, and PR-FN scores of the MOT systems constructed by five object detection algorithms and seven object tracking algorithms on the {\bf overall} \ds benchmark dataset. The evaluation results of the winners in the UA-DETRAC Challenge 2017 and 2018 are also reported. Bold faces correspond to the best performance of the MOT systems on that metric. The pink, cyan, and gray rows denote the trackers ranked in the first, second, and third places based on the PR-MOTA score with the corresponding detector.}
  \setlength{\tabcolsep}{9.5pt}
  \renewcommand{\arraystretch}{0.5}
  \footnotesize{
  \begin{tabular}{cccccccccc}
  \toprule
  Detection &Tracking &PR-MOTA &PR-MOTP &PR-MT &PR-ML &PR-IDS &PR-FM &PR-FP &PR-FN  \\
  \midrule
   \multicolumn{10}{c}{UA-DETRAC 2017 Challenge Winner} \\
  \midrule
  EB
  &IOU
  &19.4
  &28.9
  &17.7
  &18.4
  &2311.3
  &2445.9
  &14796.5
  &171806.8 \\
  \midrule

  \multicolumn{10}{c}{UA-DETRAC 2018 Challenge Winner} \\
  \midrule
  EB
  &KIOU
  &\textbf{21.1}
  &28.6
  &\textbf{21.9}
  &\textbf{17.6}
  &462.2
  &712.1
  &19046.9
  &159178.3 \\
  \midrule
  \midrule
  &GOG  &13.7  &33.7  &12.2  &20.2  &2213.4 &2466.5  &11941.6  &165757.8 \\
  &CEM   &3.1   &33.4  &1.9   &30.9   &503.8  &672.8  &17152.9  &228871.7 \\
  &DCT   &10.9 &18.9  &11.9  &19.9  &630.3  &530.8  &32104.7  &164767.6  \\
  &IHTLS &13.8 &18.9 &12.5 &20.0 &456.6 &2011.3 &13048.7 &165695.8 \\
  &H\textsuperscript{2}T &13.8 &18.9 &11.2 &20.5 &686.6 &841.3 &9522.3 &168479.6 \\

  \multirow{-5}{*}{Faster R-CNN}
  &\multicolumn{1}{>{\columncolor{tabgray}}c}{CMOT}
  &\multicolumn{1}{>{\columncolor{tabgray}}c}{14.2}
  &\multicolumn{1}{>{\columncolor{tabgray}}c}{18.9}
  &\multicolumn{1}{>{\columncolor{tabgray}}c}{14.1}
  &\multicolumn{1}{>{\columncolor{tabgray}}c}{20.0}
  &\multicolumn{1}{>{\columncolor{tabgray}}c}{157.3}
  &\multicolumn{1}{>{\columncolor{tabgray}}c}{647.5}
  &\multicolumn{1}{>{\columncolor{tabgray}}c}{15854.1}
  &\multicolumn{1}{>{\columncolor{tabgray}}c}{160390.4} \\

  &\multicolumn{1}{>{\columncolor{tabcyan}}c}{TBD}
  &\multicolumn{1}{>{\columncolor{tabcyan}}c}{14.4}
  &\multicolumn{1}{>{\columncolor{tabcyan}}c}{19.5}
  &\multicolumn{1}{>{\columncolor{tabcyan}}c}{13.5}
  &\multicolumn{1}{>{\columncolor{tabcyan}}c}{19.9}
  &\multicolumn{1}{>{\columncolor{tabcyan}}c}{1332.8}
  &\multicolumn{1}{>{\columncolor{tabcyan}}c}{1483.6}
  &\multicolumn{1}{>{\columncolor{tabcyan}}c}{10613.2}
  &\multicolumn{1}{>{\columncolor{tabcyan}}c}{163363.1} \\

  &\multicolumn{1}{>{\columncolor{tabpink}}c}{MHT}
  &\multicolumn{1}{>{\columncolor{tabpink}}c}{14.5}
  &\multicolumn{1}{>{\columncolor{tabpink}}c}{32.5}
  &\multicolumn{1}{>{\columncolor{tabpink}}c}{15.9}
  &\multicolumn{1}{>{\columncolor{tabpink}}c}{19.1}
  &\multicolumn{1}{>{\columncolor{tabpink}}c}{492.3}
  &\multicolumn{1}{>{\columncolor{tabpink}}c}{576.7}
  &\multicolumn{1}{>{\columncolor{tabpink}}c}{18141.4}
  &\multicolumn{1}{>{\columncolor{tabpink}}c}{\textbf{156227.8}} \\
  \midrule

  &\multicolumn{1}{>{\columncolor{tabpink}}c}{GOG}
  &\multicolumn{1}{>{\columncolor{tabpink}}c}{14.2}
  &\multicolumn{1}{>{\columncolor{tabpink}}c}{37.0}
  &\multicolumn{1}{>{\columncolor{tabpink}}c}{13.9}
  &\multicolumn{1}{>{\columncolor{tabpink}}c}{19.9}
  &\multicolumn{1}{>{\columncolor{tabpink}}c}{3334.6}
  &\multicolumn{1}{>{\columncolor{tabpink}}c}{3172.4}
  &\multicolumn{1}{>{\columncolor{tabpink}}c}{32092.9}
  &\multicolumn{1}{>{\columncolor{tabpink}}c}{180183.8} \\
  &CEM	                 &5.1 &35.2	&3.0	&35.3	&267.9	&352.3	 &12341.2 &260390.4\\
  &DCT	                 &10.8 &37.1 &6.7  &29.3 &141.4	&132.4	 &13226.1 &223578.8\\
  &IHTLS	             &11.1 &36.8 &13.8 &19.9 &953.6	&3556.9	 &53922.3 &180422.3\\

  &H\textsuperscript{2}T &12.4 &35.7 &14.8 &19.4 &852.2 &1117.2 &51765.7 &173899.8 \\

  \multirow{-7}{*}{CompACT}
  &\multicolumn{1}{>{\columncolor{tabgray}}c}{CMOT}
  &\multicolumn{1}{>{\columncolor{tabgray}}c}{12.6}
  &\multicolumn{1}{>{\columncolor{tabgray}}c}{36.1}
  &\multicolumn{1}{>{\columncolor{tabgray}}c}{16.1}
  &\multicolumn{1}{>{\columncolor{tabgray}}c}{18.6}
  &\multicolumn{1}{>{\columncolor{tabgray}}c}{285.3}
  &\multicolumn{1}{>{\columncolor{tabgray}}c}{1516.8}
  &\multicolumn{1}{>{\columncolor{tabgray}}c}{57885.9}
  &\multicolumn{1}{>{\columncolor{tabgray}}c}{167110.8} \\

  &\multicolumn{1}{>{\columncolor{tabcyan}}c}{TBD}
  &\multicolumn{1}{>{\columncolor{tabcyan}}c}{13.6}
  &\multicolumn{1}{>{\columncolor{tabcyan}}c}{37.3}
  &\multicolumn{1}{>{\columncolor{tabcyan}}c}{15.3}
  &\multicolumn{1}{>{\columncolor{tabcyan}}c}{19.3}
  &\multicolumn{1}{>{\columncolor{tabcyan}}c}{2026.9}
  &\multicolumn{1}{>{\columncolor{tabcyan}}c}{2467.3}
  &\multicolumn{1}{>{\columncolor{tabcyan}}c}{43247.8}
  &\multicolumn{1}{>{\columncolor{tabcyan}}c}{173837.3} \\
  \midrule

  &GOG	                 &10.0 &38.3 &13.5	&20.1 &7834.5 &7401.0 &58378.5 &192302.7\\
  &CEM	                 &2.7 &35.5	&2.3	&34.1	&778.9	&1080.4	 &34768.9 &269043.8\\

  &\multicolumn{1}{>{\columncolor{tabpink}}c}{DCT}
  &\multicolumn{1}{>{\columncolor{tabpink}}c}{11.7}
  &\multicolumn{1}{>{\columncolor{tabpink}}c}{38.0}
  &\multicolumn{1}{>{\columncolor{tabpink}}c}{10.1}
  &\multicolumn{1}{>{\columncolor{tabpink}}c}{22.8}
  &\multicolumn{1}{>{\columncolor{tabpink}}c}{758.7}	
  &\multicolumn{1}{>{\columncolor{tabpink}}c}{742.9}	
  &\multicolumn{1}{>{\columncolor{tabpink}}c}{36561.2}
  &\multicolumn{1}{>{\columncolor{tabpink}}c}{210855.6} \\

  &IHTLS	             &8.3 &38.3 &12.0	&21.4 &1536.4 &5954.9 &68662.6 &199268.8\\

  &\multicolumn{1}{>{\columncolor{tabgray}}c}{H\textsuperscript{2}T}
  &\multicolumn{1}{>{\columncolor{tabgray}}c}{11.1}
  &\multicolumn{1}{>{\columncolor{tabgray}}c}{37.3}
  &\multicolumn{1}{>{\columncolor{tabgray}}c}{14.6}
  &\multicolumn{1}{>{\columncolor{tabgray}}c}{19.8}
  &\multicolumn{1}{>{\columncolor{tabgray}}c}{1481.9}
  &\multicolumn{1}{>{\columncolor{tabgray}}c}{1820.8}
  &\multicolumn{1}{>{\columncolor{tabgray}}c}{66137.2}
  &\multicolumn{1}{>{\columncolor{tabgray}}c}{184358.2} \\

  \multirow{-7}{*}{R-CNN}
  &CMOT &11.0 &37.0 &15.7 &19.0 &506.2 &2551.1 &74253.6 &177532.6 \\

  &\multicolumn{1}{>{\columncolor{tabcyan}}c}{TBD}
  &\multicolumn{1}{>{\columncolor{tabcyan}}c}{11.6}
  &\multicolumn{1}{>{\columncolor{tabcyan}}c}{\textbf{38.7}}
  &\multicolumn{1}{>{\columncolor{tabcyan}}c}{14.6}
  &\multicolumn{1}{>{\columncolor{tabcyan}}c}{20.3}
  &\multicolumn{1}{>{\columncolor{tabcyan}}c}{4110.2}
  &\multicolumn{1}{>{\columncolor{tabcyan}}c}{4427.7}
  &\multicolumn{1}{>{\columncolor{tabcyan}}c}{56027.6}
  &\multicolumn{1}{>{\columncolor{tabcyan}}c}{188676.9} \\
  \midrule

  &\multicolumn{1}{>{\columncolor{tabpink}}c}{GOG}
  &\multicolumn{1}{>{\columncolor{tabpink}}c}{10.8}
  &\multicolumn{1}{>{\columncolor{tabpink}}c}{37.6}
  &\multicolumn{1}{>{\columncolor{tabpink}}c}{12.2}
  &\multicolumn{1}{>{\columncolor{tabpink}}c}{22.3}
  &\multicolumn{1}{>{\columncolor{tabpink}}c}{3950.8}
  &\multicolumn{1}{>{\columncolor{tabpink}}c}{3987.3}
  &\multicolumn{1}{>{\columncolor{tabpink}}c}{45201.5}
  &\multicolumn{1}{>{\columncolor{tabpink}}c}{197094.2} \\

  &CEM	 &4.5 &35.9	&2.9	&37.1	&265.4	&366.0	 &15180.3 &270643.2\\
  &DCT &7.9 &37.9 &4.8 &34.4 &108.1 &101.4 &13059.7 &251166.4 \\
  &IHTLS	 &6.6 &37.4	&11.5	&22.4	&1243.1	&4723.0	 &72757.5 &198673.5\\

  &\multicolumn{1}{>{\columncolor{tabgray}}c}{H\textsuperscript{2}T}
  &\multicolumn{1}{>{\columncolor{tabgray}}c}{8.2}
  &\multicolumn{1}{>{\columncolor{tabgray}}c}{36.5}	
  &\multicolumn{1}{>{\columncolor{tabgray}}c}{13.1}
  &\multicolumn{1}{>{\columncolor{tabgray}}c}{21.3}
  &\multicolumn{1}{>{\columncolor{tabgray}}c}{1122.8}
  &\multicolumn{1}{>{\columncolor{tabgray}}c}{1445.8}
  &\multicolumn{1}{>{\columncolor{tabgray}}c}{71567.4}
  &\multicolumn{1}{>{\columncolor{tabgray}}c}{189649.1} \\

  \multirow{-7}{*}{ACF}
  &CMOT	                 &7.8 &36.8	&14.3	&20.7	&418.3	&2161.7	 &81401.4 &183400.2\\

  &\multicolumn{1}{>{\columncolor{tabcyan}}c}{TBD}
  &\multicolumn{1}{>{\columncolor{tabcyan}}c}{9.1}
  &\multicolumn{1}{>{\columncolor{tabcyan}}c}{38.0}
  &\multicolumn{1}{>{\columncolor{tabcyan}}c}{14.1}
  &\multicolumn{1}{>{\columncolor{tabcyan}}c}{21.6}
  &\multicolumn{1}{>{\columncolor{tabcyan}}c}{2689.0}
  &\multicolumn{1}{>{\columncolor{tabcyan}}c}{3101.0}
  &\multicolumn{1}{>{\columncolor{tabcyan}}c}{64555.7}
  &\multicolumn{1}{>{\columncolor{tabcyan}}c}{189346.3} \\
  \midrule

  & \multicolumn{1}{>{\columncolor{tabpink}}c}{GOG}	
  & \multicolumn{1}{>{\columncolor{tabpink}}c}{5.5}
  & \multicolumn{1}{>{\columncolor{tabpink}}c}{28.2}
  & \multicolumn{1}{>{\columncolor{tabpink}}c}{4.1}
  & \multicolumn{1}{>{\columncolor{tabpink}}c}{27.7}
  & \multicolumn{1}{>{\columncolor{tabpink}}c}{1873.9}
  & \multicolumn{1}{>{\columncolor{tabpink}}c}{1988.5}
  & \multicolumn{1}{>{\columncolor{tabpink}}c}{38957.6}
  & \multicolumn{1}{>{\columncolor{tabpink}}c}{230126.6} \\

  &\multicolumn{1}{>{\columncolor{tabcyan}}c}{CEM}	
  &\multicolumn{1}{>{\columncolor{tabcyan}}c}{3.3}
  &\multicolumn{1}{>{\columncolor{tabcyan}}c}{27.9}
  &\multicolumn{1}{>{\columncolor{tabcyan}}c}{1.3}
  &\multicolumn{1}{>{\columncolor{tabcyan}}c}{37.8}
  &\multicolumn{1}{>{\columncolor{tabcyan}}c}{265.0}
  &\multicolumn{1}{>{\columncolor{tabcyan}}c}{317.1}
  &\multicolumn{1}{>{\columncolor{tabcyan}}c}{13888.7}
  &\multicolumn{1}{>{\columncolor{tabcyan}}c}{270718.5} \\

  &\multicolumn{1}{>{\columncolor{tabgray}}c}{DCT}
  &\multicolumn{1}{>{\columncolor{tabgray}}c}{2.7}
  &\multicolumn{1}{>{\columncolor{tabgray}}c}{29.3}
  &\multicolumn{1}{>{\columncolor{tabgray}}c}{0.5}
  &\multicolumn{1}{>{\columncolor{tabgray}}c}{42.7}
  &\multicolumn{1}{>{\columncolor{tabgray}}c}{\textbf{72.2}}
  &\multicolumn{1}{>{\columncolor{tabgray}}c}{\textbf{68.8}}
  &\multicolumn{1}{>{\columncolor{tabgray}}c}{\textbf{7785.8}}
  &\multicolumn{1}{>{\columncolor{tabgray}}c}{280762.2} \\

  &IHTLS	             &-3.0 &27.9 &1.1	&29.8	&1583.6	&4153.5	 &79197.5	&244232.8\\
  &H\textsuperscript{2}T &-0.7 &28.8 &2.1	&28.4	&1738.8	&1525.6	 &71631.0	&236520.9\\
  \multirow{-7}{*}{DPM}
  &CMOT	                 &-3.4 &28.4 &5.1	&26.6	&447.5	&1040.5	 &104768.3	&221991.7\\
  &TBD                       &-1.5 &30.3 &5.5 &27.0 &1914.3 &1707.2 &88863.0 &224179.7 \\
  \bottomrule
  \end{tabular}}
  \label{tab:mot-results}
\end{table*}

\begin{figure}[t]
\centering
\includegraphics[width=\linewidth]{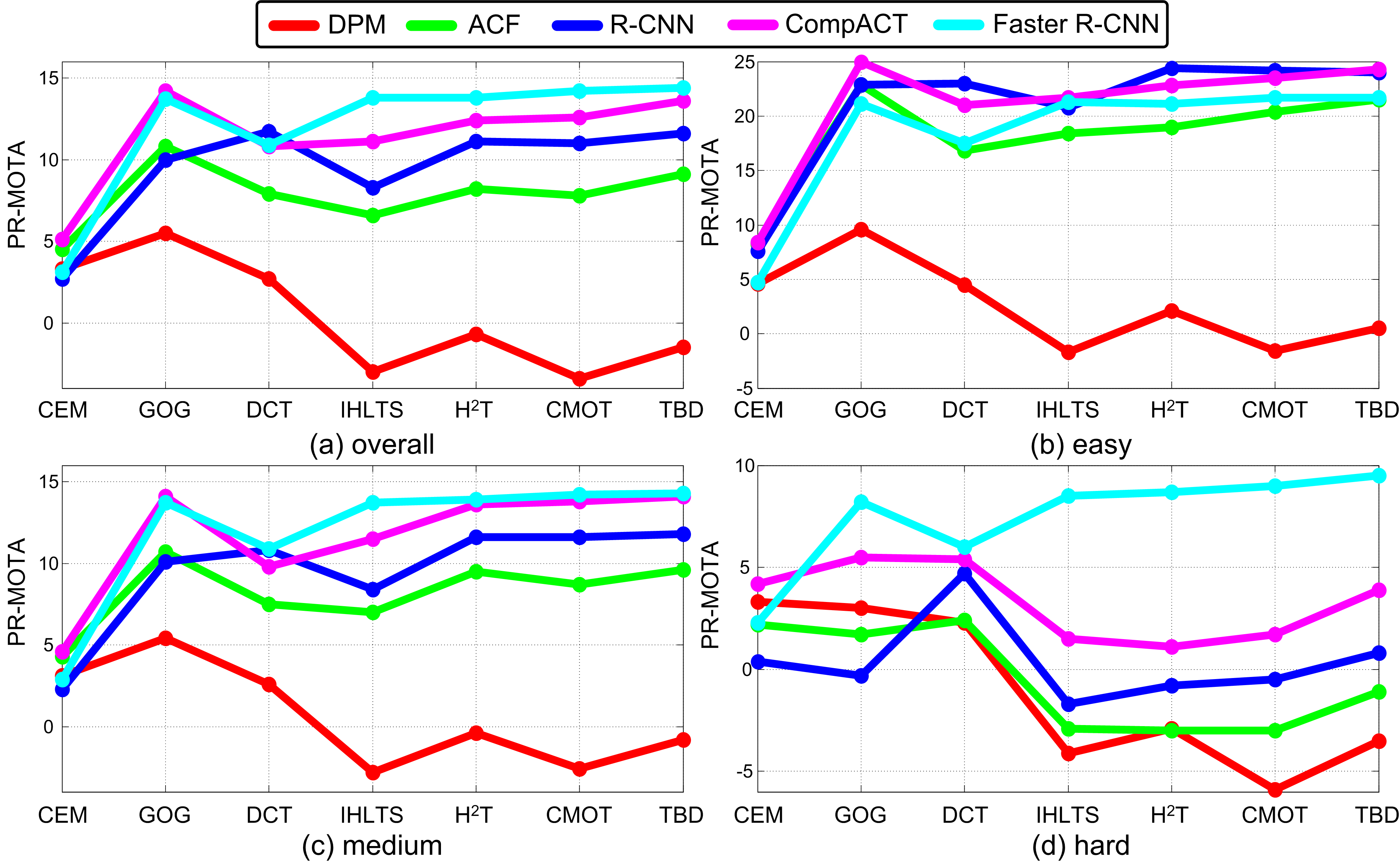}
\caption{Performance trends of the MOT systems constructed by different detection and tracking algorithms. The $x$-axis corresponds to different tracking algorithms, and the $y$-axis is the PR-MOTA scores of different MOT systems. Different colors of the curves indicate different object detection algorithms.}
\label{fig:performance-trends}
\end{figure}

\begin{table*}[t]
  \centering
  \caption{PR-MOTA, PR-MOTP, PR-MT, PR-ML, PR-IDS, PR-FM, PR-FP, and PR-FN scores of the MOT systems constructed by five object detection algorithms and seven object tracking algorithms on the {\bf {\em easy} subset} of the \ds benchmark dataset. The evaluation results of the winners in the UA-DETRAC Challenge 2017 and 2018 are also reported. Bold faces correspond to the best performance of the MOT systems on that metric. The pink, cyan, and gray rows denote the trackers ranked in the first, second, and third places based on the PR-MOTA score with the corresponding detector.}

  \label{tab:tracking-easy-results}
  \setlength{\tabcolsep}{9.5pt}
  \renewcommand{\arraystretch}{0.5}
  \footnotesize{
  \begin{tabular}{cccccccccc}
  \toprule
  \multicolumn{10}{c}{{\bf easy subset}} \\
  \midrule
  Detection &Tracking &PR-MOTA &PR-MOTP &PR-MT &PR-ML &PR-IDS &PR-FM &PR-FP &PR-FN  \\
  \midrule
   \multicolumn{10}{c}{UA-DETRAC 2017 Challenge Winner} \\
  \midrule
  EB
  &IOU
  &34.0
  &37.8
  &27.8
  &20.4
  &573.0
  &602.9
  &1579.6
  &33765.7 \\
  \midrule
  \multicolumn{10}{c}{UA-DETRAC 2018 Challenge Winner} \\
  \midrule
  EB
  &KIOU
  &\textbf{36.4}
  &37.6
  &\textbf{33.7}
  &20.6
  &78.1
  &121.6
  &2103.8
  &\textbf{30722.4} \\
  \midrule
  \midrule
  &GOG                         &21.1 &41.1 &16.4 &21.9 &582.2 &609.3 &1300.3 &33751.3 \\
  &CEM                          &4.7 &39.7 &3.0 &34.9  &121.1 &154.0  &3818.8 &52270.9 \\
  &DCT                           &17.5  &22.8 &15.4  &21.8 &131.2 &105.1 &5840.0  &34153.8 \\
  &IHTLS &21.3 &23.3 &16.5 &21.7 &87.3 &480.3 &1418.2 &33826.7 \\
  &H\textsuperscript{2}T          &21.1 &23.3  &14.4  &22.3  &120.0  &143.7  &\textbf{786.1}  &34666.3 \\
  \multirow{-5}{*}{Faster R-CNN}
  &\multicolumn{1}{>{\columncolor{tabcyan}}c}{CMOT}
  &\multicolumn{1}{>{\columncolor{tabcyan}}c}{21.7}
  &\multicolumn{1}{>{\columncolor{tabcyan}}c}{23.6}
  &\multicolumn{1}{>{\columncolor{tabcyan}}c}{18.5}
  &\multicolumn{1}{>{\columncolor{tabcyan}}c}{21.8}
  &\multicolumn{1}{>{\columncolor{tabcyan}}c}{25.1}
  &\multicolumn{1}{>{\columncolor{tabcyan}}c}{112.3}
  &\multicolumn{1}{>{\columncolor{tabcyan}}c}{2241.7}
  &\multicolumn{1}{>{\columncolor{tabcyan}}c}{32648.9} \\

  &\multicolumn{1}{>{\columncolor{tabcyan}}c}{TBD}
  &\multicolumn{1}{>{\columncolor{tabcyan}}c}{21.7}
  &\multicolumn{1}{>{\columncolor{tabcyan}}c}{24.1}
  &\multicolumn{1}{>{\columncolor{tabcyan}}c}{17.3}
  &\multicolumn{1}{>{\columncolor{tabcyan}}c}{21.7}
  &\multicolumn{1}{>{\columncolor{tabcyan}}c}{356.8}
  &\multicolumn{1}{>{\columncolor{tabcyan}}c}{385.4}
  &\multicolumn{1}{>{\columncolor{tabcyan}}c}{1128.3}
  &\multicolumn{1}{>{\columncolor{tabcyan}}c}{33444.6} \\

  &\multicolumn{1}{>{\columncolor{tabpink}}c}{MHT}
  &\multicolumn{1}{>{\columncolor{tabpink}}c}{23.0}
  &\multicolumn{1}{>{\columncolor{tabpink}}c}{24.1}
  &\multicolumn{1}{>{\columncolor{tabpink}}c}{21.0}
  &\multicolumn{1}{>{\columncolor{tabpink}}c}{21.5}
  &\multicolumn{1}{>{\columncolor{tabpink}}c}{38.8}
  &\multicolumn{1}{>{\columncolor{tabpink}}c}{48.8}
  &\multicolumn{1}{>{\columncolor{tabpink}}c}{1804.3}
  &\multicolumn{1}{>{\columncolor{tabpink}}c}{31442.1} \\
  \midrule
  &\multicolumn{1}{>{\columncolor{tabpink}}c}{GOG}
  &\multicolumn{1}{>{\columncolor{tabpink}}c}{25.0}
  &\multicolumn{1}{>{\columncolor{tabpink}}c}{47.0}
  &\multicolumn{1}{>{\columncolor{tabpink}}c}{21.4}
  &\multicolumn{1}{>{\columncolor{tabpink}}c}{18.8}
  &\multicolumn{1}{>{\columncolor{tabpink}}c}{852.3}
  &\multicolumn{1}{>{\columncolor{tabpink}}c}{795.8}
  &\multicolumn{1}{>{\columncolor{tabpink}}c}{6493.7}
  &\multicolumn{1}{>{\columncolor{tabpink}}c}{34383.1} \\

  &CEM     &8.4	 &43.7 &4.0  &39.6 &75.9  &91.2  &3365.3  &59106.9 \\
  &DCT     &21.0 &45.7 &11.4 &26.4 &48.1  &44.0  &4187.6  &42475.8 \\
  &IHTLS   &21.7 &46.6 &21.0 &19.4 &184.0 &761.0 &10848.4 &34806.9 \\
  &H\textsuperscript{2}T &22.8 &44.1 &22.6 &19.5 &168.7 &198.6 &10563.3 &33690.7 \\

  \multirow{-6}{*}{CompACT}
  &\multicolumn{1}{>{\columncolor{tabgray}}c}{CMOT}
  &\multicolumn{1}{>{\columncolor{tabgray}}c}{23.5}
  &\multicolumn{1}{>{\columncolor{tabgray}}c}{45.9}
  &\multicolumn{1}{>{\columncolor{tabgray}}c}{24.3}
  &\multicolumn{1}{>{\columncolor{tabgray}}c}{17.8}
  &\multicolumn{1}{>{\columncolor{tabgray}}c}{42.3}
  &\multicolumn{1}{>{\columncolor{tabgray}}c}{263.4}
  &\multicolumn{1}{>{\columncolor{tabgray}}c}{11820.6}
  &\multicolumn{1}{>{\columncolor{tabgray}}c}{31687.2} \\

  &\multicolumn{1}{>{\columncolor{tabcyan}}c}{TBD}
  &\multicolumn{1}{>{\columncolor{tabcyan}}c}{24.3}
  &\multicolumn{1}{>{\columncolor{tabcyan}}c}{47.2}
  &\multicolumn{1}{>{\columncolor{tabcyan}}c}{22.7}
  &\multicolumn{1}{>{\columncolor{tabcyan}}c}{18.7}
  &\multicolumn{1}{>{\columncolor{tabcyan}}c}{443.3}
  &\multicolumn{1}{>{\columncolor{tabcyan}}c}{555.7}
  &\multicolumn{1}{>{\columncolor{tabcyan}}c}{8615.8}
  &\multicolumn{1}{>{\columncolor{tabcyan}}c}{33516.7} \\
  \midrule
  &GOG     &22.9 &47.7 &23.3 &18.6 &1603.0 &1515.2 &9530.0  &35296.2 \\
  &CEM     &7.6  &43.6 &4.0  &35.6 &171.8  &223.2  &6825.4  &58504.9 \\
  &DCT &23.0 &46.3  &18.1  &19.2  &180.8  &173.5  &8319.1  &37772.2 \\
  &IHTLS   &20.8 &47.7 &20.5 &19.8 &297.8  &1306.5 &11487.9 &37172.3 \\

  &\multicolumn{1}{>{\columncolor{tabpink}}c}{H\textsuperscript{2}T}
  &\multicolumn{1}{>{\columncolor{tabpink}}c}{24.4}
  &\multicolumn{1}{>{\columncolor{tabpink}}c}{46.7}
  &\multicolumn{1}{>{\columncolor{tabpink}}c}{24.3}
  &\multicolumn{1}{>{\columncolor{tabpink}}c}{18.3}
  &\multicolumn{1}{>{\columncolor{tabpink}}c}{288.7}
  &\multicolumn{1}{>{\columncolor{tabpink}}c}{320.6}
  &\multicolumn{1}{>{\columncolor{tabpink}}c}{10318.7}
  &\multicolumn{1}{>{\columncolor{tabpink}}c}{33920.7} \\

  \multirow{-6}{*}{R-CNN}
  &\multicolumn{1}{>{\columncolor{tabcyan}}c}{CMOT}
  &\multicolumn{1}{>{\columncolor{tabcyan}}c}{24.2}
  &\multicolumn{1}{>{\columncolor{tabcyan}}c}{46.8}
  &\multicolumn{1}{>{\columncolor{tabcyan}}c}{26.0}
  &\multicolumn{1}{>{\columncolor{tabcyan}}c}{\textbf{17.7}}
  &\multicolumn{1}{>{\columncolor{tabcyan}}c}{71.4}
  &\multicolumn{1}{>{\columncolor{tabcyan}}c}{404.9}
  &\multicolumn{1}{>{\columncolor{tabcyan}}c}{12250.1}
  &\multicolumn{1}{>{\columncolor{tabcyan}}c}{32384.1} \\

  &\multicolumn{1}{>{\columncolor{tabgray}}c}{TBD}
  &\multicolumn{1}{>{\columncolor{tabgray}}c}{24.0}
  &\multicolumn{1}{>{\columncolor{tabgray}}c}{48.2}
  &\multicolumn{1}{>{\columncolor{tabgray}}c}{23.8}
  &\multicolumn{1}{>{\columncolor{tabgray}}c}{19.1}
  &\multicolumn{1}{>{\columncolor{tabgray}}c}{943.1}
  &\multicolumn{1}{>{\columncolor{tabgray}}c}{950.1}
  &\multicolumn{1}{>{\columncolor{tabgray}}c}{8776.4}
  &\multicolumn{1}{>{\columncolor{tabgray}}c}{35263.3} \\
  \midrule
  &\multicolumn{1}{>{\columncolor{tabpink}}c}{GOG}
  &\multicolumn{1}{>{\columncolor{tabpink}}c}{22.9}
  &\multicolumn{1}{>{\columncolor{tabpink}}c}{\textbf{48.6}}
  &\multicolumn{1}{>{\columncolor{tabpink}}c}{20.3}
  &\multicolumn{1}{>{\columncolor{tabpink}}c}{20.9}
  &\multicolumn{1}{>{\columncolor{tabpink}}c}{1070.1}
  &\multicolumn{1}{>{\columncolor{tabpink}}c}{1039.9}
  &\multicolumn{1}{>{\columncolor{tabpink}}c}{9086.1}
  &\multicolumn{1}{>{\columncolor{tabpink}}c}{37899.7} \\

  &CEM     &8.2  &45.3 &4.2  &42.3 &73.3   &92.9   &3501.1  &62814.7 \\
  &DCT     &16.8 &47.4 &8.2  &34.4 &43.0   &39.0   &4100.3  &51456.3 \\
  &IHTLS   &18.4 &48.3 &19.2 &21.5 &287.4  &1157.4 &14449.8 &38849.3 \\
  &H\textsuperscript{2}T &19.0 &45.7 &21.4 &22.2 &234.1 &282.0 &14873.0 &37750.6 \\

  \multirow{-6}{*}{ACF}
  &\multicolumn{1}{>{\columncolor{tabgray}}c}{CMOT}
  &\multicolumn{1}{>{\columncolor{tabgray}}c}{20.4}
  &\multicolumn{1}{>{\columncolor{tabgray}}c}{47.6}
  &\multicolumn{1}{>{\columncolor{tabgray}}c}{23.5}
  &\multicolumn{1}{>{\columncolor{tabgray}}c}{19.7}
  &\multicolumn{1}{>{\columncolor{tabgray}}c}{66.4}
  &\multicolumn{1}{>{\columncolor{tabgray}}c}{390.0}
  &\multicolumn{1}{>{\columncolor{tabgray}}c}{16193.1}
  &\multicolumn{1}{>{\columncolor{tabgray}}c}{34825.4} \\

  &\multicolumn{1}{>{\columncolor{tabcyan}}c}{TBD}
  &\multicolumn{1}{>{\columncolor{tabcyan}}c}{21.5}
  &\multicolumn{1}{>{\columncolor{tabcyan}}c}{48.9}
  &\multicolumn{1}{>{\columncolor{tabcyan}}c}{22.5}
  &\multicolumn{1}{>{\columncolor{tabcyan}}c}{20.8}
  &\multicolumn{1}{>{\columncolor{tabcyan}}c}{611.2}
  &\multicolumn{1}{>{\columncolor{tabcyan}}c}{702.8}
  &\multicolumn{1}{>{\columncolor{tabcyan}}c}{12499.0}
  &\multicolumn{1}{>{\columncolor{tabcyan}}c}{36667.2} \\
  \midrule
  &\multicolumn{1}{>{\columncolor{tabpink}}c}{GOG}
  &\multicolumn{1}{>{\columncolor{tabpink}}c}{9.6}
  &\multicolumn{1}{>{\columncolor{tabpink}}c}{34.9}
  &\multicolumn{1}{>{\columncolor{tabpink}}c}{4.9}
  &\multicolumn{1}{>{\columncolor{tabpink}}c}{37.8}
  &\multicolumn{1}{>{\columncolor{tabpink}}c}{550.6}
  &\multicolumn{1}{>{\columncolor{tabpink}}c}{573.8}
  &\multicolumn{1}{>{\columncolor{tabpink}}c}{8584.1}
  &\multicolumn{1}{>{\columncolor{tabpink}}c}{60684.7} \\

  &\multicolumn{1}{>{\columncolor{tabcyan}}c}{CEM}
  &\multicolumn{1}{>{\columncolor{tabcyan}}c}{4.6}
  &\multicolumn{1}{>{\columncolor{tabcyan}}c}{35.5}
  &\multicolumn{1}{>{\columncolor{tabcyan}}c}{1.3}
  &\multicolumn{1}{>{\columncolor{tabcyan}}c}{51.6}
  &\multicolumn{1}{>{\columncolor{tabcyan}}c}{64.6}
  &\multicolumn{1}{>{\columncolor{tabcyan}}c}{71.0}
  &\multicolumn{1}{>{\columncolor{tabcyan}}c}{3321.4}
  &\multicolumn{1}{>{\columncolor{tabcyan}}c}{72758.6} \\

  &\multicolumn{1}{>{\columncolor{tabgray}}c}{DCT}
  &\multicolumn{1}{>{\columncolor{tabgray}}c}{4.5}
  &\multicolumn{1}{>{\columncolor{tabgray}}c}{37.2}
  &\multicolumn{1}{>{\columncolor{tabgray}}c}{0.6}
  &\multicolumn{1}{>{\columncolor{tabgray}}c}{57.1}
  &\multicolumn{1}{>{\columncolor{tabgray}}c}{\textbf{22.0}}
  &\multicolumn{1}{>{\columncolor{tabgray}}c}{\textbf{21.0}}
  &\multicolumn{1}{>{\columncolor{tabgray}}c}{2411.5}
  &\multicolumn{1}{>{\columncolor{tabgray}}c}{73868.8} \\

  &IHTLS   &-1.7 &29.3 &1.1 &40.3 &428.9 &1184.1 &18573.3 &64988.9 \\
  &H\textsuperscript{2}T &2.1 &35.1 &2.7 &38.8 &503.7 &420.8 &15708.8 &62983.8 \\
   \multirow{-6}{*}{DPM}
  &CMOT    &-1.6 &34.7 &6.1 &36.6 &92.9  &253.1  &24864.9 &58973.8 \\
  &TBD                       &0.5 &38.6 &6.3 &60.7 &474.4 &443.5 &21151.2 &88645.3 \\
  \bottomrule
  \end{tabular}}
\end{table*}

\begin{table*}[t]
  \centering
  \caption{PR-MOTA, PR-MOTP, PR-MT, PR-ML, PR-IDS, PR-FM, PR-FP, and PR-FN scores of the MOT systems constructed by five object detection algorithms and seven object tracking algorithms on the {\bf {\em medium} subset} of the \ds benchmark dataset. The evaluation results of the winners in the UA-DETRAC Challenge 2017 and 2018 are also reported. Bold faces correspond to the best performance of the MOT systems on that metric. The pink, cyan, and gray rows denote the trackers ranked in the first, second, and third places based on the PR-MOTA score with the corresponding detector.}
  \label{tab:tracking-medium-results}
  \setlength{\tabcolsep}{9.5pt}
  \renewcommand{\arraystretch}{0.5}
  \footnotesize{
  \begin{tabular}{cccccccccc}
  \toprule
  \multicolumn{10}{c}{{\bf {\em medium} subset}} \\
  \midrule
  Detection &Tracking &PR-MOTA &PR-MOTP &PR-MT &PR-ML &PR-IDS &PR-FM &PR-FP &PR-FN  \\
  \midrule
   \multicolumn{10}{c}{UA-DETRAC 2017 Challenge Winner} \\
  \midrule
  EB
  &IOU
  &18.2
  &26.6
  &15.5
  &18.3
  &1201.0
  &1271.9
  &5779.6
  &94920.7 \\
  \midrule
  \multicolumn{10}{c}{UA-DETRAC 2018 Challenge Winner} \\
  \midrule
  EB
  &KIOU
  &\textbf{19.9}
  &26.3
  &\textbf{19.6}
  &\textbf{17.3}
  &229.2
  &356.2
  &7628.6
  &\textbf{88101.0} \\
  \midrule
  \midrule
  &GOG                           &13.7  &17.6  &12.2  &19.9  &1151.7  &1287.6  &4917.1  &91880.0 \\
  &CEM                           &2.9  &16.8  &1.3  &30.9  &262.2  &355.4  &8866.1 &127841.6 \\
  &DCT                           &10.9  &18.1 &10.4  &19.8  &355.1  &307.3  &16017.9  &91936.0 \\
  &IHTLS                        &13.7  &18.2  &12.1 &20.0  &242.8  &1060.0  &5600.0  &92061.5 \\
  &H\textsuperscript{2}T &13.9 &17.9 &10.8 &20.5 &364.3 &467.7 &\textbf{3940.1}   &93228.5 \\

  \multirow{-5}{*}{Faster R-CNN}
  &\multicolumn{1}{>{\columncolor{tabcyan}}c}{CMOT}
  &\multicolumn{1}{>{\columncolor{tabcyan}}c}{14.2}
  &\multicolumn{1}{>{\columncolor{tabcyan}}c}{18.1}
  &\multicolumn{1}{>{\columncolor{tabcyan}}c}{13.5}
  &\multicolumn{1}{>{\columncolor{tabcyan}}c}{19.9}
  &\multicolumn{1}{>{\columncolor{tabcyan}}c}{86.8}
  &\multicolumn{1}{>{\columncolor{tabcyan}}c}{359.8}
  &\multicolumn{1}{>{\columncolor{tabcyan}}c}{7033.9}
  &\multicolumn{1}{>{\columncolor{tabcyan}}c}{89273.1} \\

  &\multicolumn{1}{>{\columncolor{tabpink}}c}{TBD}
  &\multicolumn{1}{>{\columncolor{tabpink}}c}{14.3}
  &\multicolumn{1}{>{\columncolor{tabpink}}c}{18.7}
  &\multicolumn{1}{>{\columncolor{tabpink}}c}{12.9}
  &\multicolumn{1}{>{\columncolor{tabpink}}c}{19.8}
  &\multicolumn{1}{>{\columncolor{tabpink}}c}{697.6}
  &\multicolumn{1}{>{\columncolor{tabpink}}c}{776.5}
  &\multicolumn{1}{>{\columncolor{tabpink}}c}{4542.5}
  &\multicolumn{1}{>{\columncolor{tabpink}}c}{90846.1} \\

  &\multicolumn{1}{>{\columncolor{tabgray}}c}{MHT}
  &\multicolumn{1}{>{\columncolor{tabgray}}c}{14.1}
  &\multicolumn{1}{>{\columncolor{tabgray}}c}{31.8}
  &\multicolumn{1}{>{\columncolor{tabgray}}c}{15.0}
  &\multicolumn{1}{>{\columncolor{tabgray}}c}{19.1}
  &\multicolumn{1}{>{\columncolor{tabgray}}c}{291.3}
  &\multicolumn{1}{>{\columncolor{tabgray}}c}{342.9}
  &\multicolumn{1}{>{\columncolor{tabgray}}c}{8348.6}
  &\multicolumn{1}{>{\columncolor{tabgray}}c}{88117.0} \\
  \midrule
  &\multicolumn{1}{>{\columncolor{tabpink}}c}{GOG}
  &\multicolumn{1}{>{\columncolor{tabpink}}c}{14.1}
  &\multicolumn{1}{>{\columncolor{tabpink}}c}{35.1}
  &\multicolumn{1}{>{\columncolor{tabpink}}c}{11.6}
  &\multicolumn{1}{>{\columncolor{tabpink}}c}{20.4}
  &\multicolumn{1}{>{\columncolor{tabpink}}c}{1879.4}
  &\multicolumn{1}{>{\columncolor{tabpink}}c}{1756.1}
  &\multicolumn{1}{>{\columncolor{tabpink}}c}{14534.6}
  &\multicolumn{1}{>{\columncolor{tabpink}}c}{98318.8} \\

  &CEM     &4.6  &33.8 &2.4  &34.6 &137.2  &182.9  &6371.9  &142184.8 \\
  &DCT     &9.8  &35.4 &5.3  &29.7 &75.2   &70.9   &6781.1  &123203.6 \\
  &IHTLS   &11.5 &34.9 &11.7 &20.5 &527.4  &1946.6 &25120.6 &98255.7 \\
  &H\textsuperscript{2}T &13.6 &34.3 &12.4 &19.6 &458.1 &607.1 &21843.7 &94097.3 \\

  \multirow{-6}{*}{CompACT}
  &\multicolumn{1}{>{\columncolor{tabgray}}c}{CMOT}
  &\multicolumn{1}{>{\columncolor{tabgray}}c}{13.8}
  &\multicolumn{1}{>{\columncolor{tabgray}}c}{34.3}
  &\multicolumn{1}{>{\columncolor{tabgray}}c}{13.5}
  &\multicolumn{1}{>{\columncolor{tabgray}}c}{19.1}
  &\multicolumn{1}{>{\columncolor{tabgray}}c}{161.5}
  &\multicolumn{1}{>{\columncolor{tabgray}}c}{846.8}
  &\multicolumn{1}{>{\columncolor{tabgray}}c}{24875.4}
  &\multicolumn{1}{>{\columncolor{tabgray}}c}{90813.1} \\

  &\multicolumn{1}{>{\columncolor{tabpink}}c}{TBD}
  &\multicolumn{1}{>{\columncolor{tabpink}}c}{14.1}
  &\multicolumn{1}{>{\columncolor{tabpink}}c}{34.9}
  &\multicolumn{1}{>{\columncolor{tabpink}}c}{11.7}
  &\multicolumn{1}{>{\columncolor{tabpink}}c}{20.3}
  &\multicolumn{1}{>{\columncolor{tabpink}}c}{1155.2}
  &\multicolumn{1}{>{\columncolor{tabpink}}c}{1407.2}
  &\multicolumn{1}{>{\columncolor{tabpink}}c}{17533.0}
  &\multicolumn{1}{>{\columncolor{tabpink}}c}{94782.5} \\
  \midrule
  &GOG     &10.1 &35.8 &10.7 &19.8 &4538.5 &4234.0 &27352.7 &102789.4 \\
  &CEM     &2.3  &33.4 &1.7  &33.2 &432.7  &600.9  &17734.4 &144310.0 \\
  &DCT     &10.8 &35.7 &7.4 &22.8 &409.8 &401.1 &16512.6 &115166.2 \\
  &IHTLS   &8.4  &35.8 &9.8  &21.0 &863.3  &3290.8 &33694.8 &106288.7 \\

  &\multicolumn{1}{>{\columncolor{tabcyan}}c}{H\textsuperscript{2}T}
  &\multicolumn{1}{>{\columncolor{tabcyan}}c}{11.6}
  &\multicolumn{1}{>{\columncolor{tabcyan}}c}{32.6}
  &\multicolumn{1}{>{\columncolor{tabcyan}}c}{12.1}
  &\multicolumn{1}{>{\columncolor{tabcyan}}c}{19.5}
  &\multicolumn{1}{>{\columncolor{tabcyan}}c}{800.1}
  &\multicolumn{1}{>{\columncolor{tabcyan}}c}{986.9}
  &\multicolumn{1}{>{\columncolor{tabcyan}}c}{30594.7}
  &\multicolumn{1}{>{\columncolor{tabcyan}}c}{97768.2} \\

  \multirow{-6}{*}{R-CNN}
  &\multicolumn{1}{>{\columncolor{tabcyan}}c}{CMOT}
  &\multicolumn{1}{>{\columncolor{tabcyan}}c}{11.6}
  &\multicolumn{1}{>{\columncolor{tabcyan}}c}{34.5}
  &\multicolumn{1}{>{\columncolor{tabcyan}}c}{13.0}
  &\multicolumn{1}{>{\columncolor{tabcyan}}c}{18.6}
  &\multicolumn{1}{>{\columncolor{tabcyan}}c}{302.8}
  &\multicolumn{1}{>{\columncolor{tabcyan}}c}{1510.4}
  &\multicolumn{1}{>{\columncolor{tabcyan}}c}{34643.8}
  &\multicolumn{1}{>{\columncolor{tabcyan}}c}{94466.6} \\

  &\multicolumn{1}{>{\columncolor{tabpink}}c}{TBD}
  &\multicolumn{1}{>{\columncolor{tabpink}}c}{11.8}
  &\multicolumn{1}{>{\columncolor{tabpink}}c}{\textbf{36.2}}
  &\multicolumn{1}{>{\columncolor{tabpink}}c}{11.1}
  &\multicolumn{1}{>{\columncolor{tabpink}}c}{21.3}
  &\multicolumn{1}{>{\columncolor{tabpink}}c}{2286.7}
  &\multicolumn{1}{>{\columncolor{tabpink}}c}{2529.7}
  &\multicolumn{1}{>{\columncolor{tabpink}}c}{25102.6}
  &\multicolumn{1}{>{\columncolor{tabpink}}c}{104902.7} \\
  \midrule
  &\multicolumn{1}{>{\columncolor{tabpink}}c}{GOG}
  &\multicolumn{1}{>{\columncolor{tabpink}}c}{10.7}
  &\multicolumn{1}{>{\columncolor{tabpink}}c}{35.1}
  &\multicolumn{1}{>{\columncolor{tabpink}}c}{10.0}
  &\multicolumn{1}{>{\columncolor{tabpink}}c}{22.6}
  &\multicolumn{1}{>{\columncolor{tabpink}}c}{2241.0}
  &\multicolumn{1}{>{\columncolor{tabpink}}c}{2244.9}
  &\multicolumn{1}{>{\columncolor{tabpink}}c}{21540.0}
  &\multicolumn{1}{>{\columncolor{tabpink}}c}{106318.5} \\

  &CEM     &4.3  &34.0 &2.6  &35.5 &148.2  &204.7  &7907.1  &144650.8 \\
  &DCT     &7.5  &35.7 &4.1  &33.4 &58.8   &55.3   &6184.6  &135241.2 \\
  &IHTLS   &7.0  &34.9 &9.5  &22.5 &700.4  &2634.0 &35626.0 &106777.5 \\

  &\multicolumn{1}{>{\columncolor{tabgray}}c}{H\textsuperscript{2}T}
  &\multicolumn{1}{>{\columncolor{tabgray}}c}{9.5}
  &\multicolumn{1}{>{\columncolor{tabgray}}c}{34.3}
  &\multicolumn{1}{>{\columncolor{tabgray}}c}{10.9}
  &\multicolumn{1}{>{\columncolor{tabgray}}c}{21.0}
  &\multicolumn{1}{>{\columncolor{tabgray}}c}{618.2}
  &\multicolumn{1}{>{\columncolor{tabgray}}c}{801.8}
  &\multicolumn{1}{>{\columncolor{tabgray}}c}{32650.6}
  &\multicolumn{1}{>{\columncolor{tabgray}}c}{101111.4} \\

  \multirow{-6}{*}{ACF}
  &CMOT &8.7 &34.3 &11.8 &20.8 &255.3 &1292.6 &38341.8 &98554.0 \\

  &\multicolumn{1}{>{\columncolor{tabcyan}}c}{TBD}
  &\multicolumn{1}{>{\columncolor{tabcyan}}c}{9.6}
  &\multicolumn{1}{>{\columncolor{tabcyan}}c}{35.0}
  &\multicolumn{1}{>{\columncolor{tabcyan}}c}{10.5}
  &\multicolumn{1}{>{\columncolor{tabcyan}}c}{22.2}
  &\multicolumn{1}{>{\columncolor{tabcyan}}c}{1571.5}
  &\multicolumn{1}{>{\columncolor{tabcyan}}c}{1825.9}
  &\multicolumn{1}{>{\columncolor{tabcyan}}c}{29180.9}
  &\multicolumn{1}{>{\columncolor{tabcyan}}c}{102010.9} \\
  \midrule
  &\multicolumn{1}{>{\columncolor{tabpink}}c}{GOG}
  &\multicolumn{1}{>{\columncolor{tabpink}}c}{5.4}
  &\multicolumn{1}{>{\columncolor{tabpink}}c}{29.4}
  &\multicolumn{1}{>{\columncolor{tabpink}}c}{3.5}
  &\multicolumn{1}{>{\columncolor{tabpink}}c}{28.7}
  &\multicolumn{1}{>{\columncolor{tabpink}}c}{1061.7}
  &\multicolumn{1}{>{\columncolor{tabpink}}c}{1104.7}
  &\multicolumn{1}{>{\columncolor{tabpink}}c}{21447.7}
  &\multicolumn{1}{>{\columncolor{tabpink}}c}{131043.5} \\

  &\multicolumn{1}{>{\columncolor{tabcyan}}c}{CEM}
  &\multicolumn{1}{>{\columncolor{tabcyan}}c}{3.1}
  &\multicolumn{1}{>{\columncolor{tabcyan}}c}{28.8}
  &\multicolumn{1}{>{\columncolor{tabcyan}}c}{1.0}
  &\multicolumn{1}{>{\columncolor{tabcyan}}c}{38.1}
  &\multicolumn{1}{>{\columncolor{tabcyan}}c}{153.7}
  &\multicolumn{1}{>{\columncolor{tabcyan}}c}{188.1}
  &\multicolumn{1}{>{\columncolor{tabcyan}}c}{8581.5}
  &\multicolumn{1}{>{\columncolor{tabcyan}}c}{152847.9} \\

  &\multicolumn{1}{>{\columncolor{tabgray}}c}{DCT}
  &\multicolumn{1}{>{\columncolor{tabgray}}c}{2.6}
  &\multicolumn{1}{>{\columncolor{tabgray}}c}{30.1}
  &\multicolumn{1}{>{\columncolor{tabgray}}c}{0.5}
  &\multicolumn{1}{>{\columncolor{tabgray}}c}{42.7}
  &\multicolumn{1}{>{\columncolor{tabgray}}c}{\textbf{42.7}}
  &\multicolumn{1}{>{\columncolor{tabgray}}c}{\textbf{41.1}}
  &\multicolumn{1}{>{\columncolor{tabgray}}c}{4838.6}
  &\multicolumn{1}{>{\columncolor{tabgray}}c}{158656.3} \\

  &IHTLS   &-2.8 &27.9 &1.1 &30.8 &842.1  &2227.8 &43458.0 &138436.8 \\
  &H\textsuperscript{2}T &-0.4 &29.1 &1.8 &29.5 &946.9 &841.9 &38930.3 &134364.9 \\
  \multirow{-6}{*}{DPM}
  &CMOT    &-2.6 &29.6 &4.3 &27.7 &249.4  &580.2  &55505.1 &126376.9 \\
  &TBD                       &-0.8 &31.4 &5.0 &28.1 &1085.9 &972.0 &48089.8 &128647.7 \\
  \bottomrule
  \end{tabular}}
\end{table*}


\begin{table*}[t]
  \centering
  \caption{PR-MOTA, PR-MOTP, PR-MT, PR-ML, PR-IDS, PR-FM, PR-FP, and PR-FN scores of the MOT systems constructed by five object detection algorithms and seven object tracking algorithms on the {\bf {\em hard} subset} of the \ds benchmark dataset. The evaluation results of the winners in the UA-DETRAC Challenge 2017 and 2018 are also reported. Bold faces correspond to the best performance of the MOT systems on that metric. The pink, cyan, and gray rows denote the trackers ranked in the first, second, and third places based on the PR-MOTA score with the corresponding detector.}
  \label{tab:tracking-hard-results}
  \setlength{\tabcolsep}{9.5pt}
  \renewcommand{\arraystretch}{0.5}
  \footnotesize{
  \begin{tabular}{cccccccccc}
  \toprule
  \multicolumn{10}{c}{{\bf {\em hard} subset}} \\
  \midrule
  Detection &Tracking &PR-MOTA &PR-MOTP &PR-MT &PR-ML &PR-IDS &PR-FM &PR-FP &PR-FN  \\
  \midrule
   \multicolumn{10}{c}{UA-DETRAC 2017 Challenge Winner} \\
  \midrule
  EB
  &IOU
  &11.9
  &26.2
  &12.8
  &18.7
  &565.3
  &598.4
  &6704.9
  &42709.8 \\
  \midrule

  \multicolumn{10}{c}{UA-DETRAC 2018 Challenge Winner} \\
  \midrule
  EB
  &KIOU
  &\textbf{13.1}
  &25.6
  &\textbf{16.2}
  &\textbf{17.4}
  &145.3
  &217.7
  &8443.2
  &39633.6 \\
  \midrule
  \midrule

  &GOG                           &8.2  &30.1  &8.1  &21.1 &507.1 &586.1  &5357.1 &41423.9 \\
  &CEM                           &2.3  &30.7 &1.4  &29.8  &124.7 &169.5  &4508.6 &51601.1 \\
  &DCT                            &6.0  &17.2  &11.1  &20.2 &147.9  &124.7 &10285.5  &40274.1 \\
  &IHTLS                         &8.5  &16.8  &9.6  &20.4  &126.9  &485.4 &5664.8 &41150.7 \\
  &H\textsuperscript{2}T &8.7 &17.1 &9.3 &20.7 &203.4 &230.3 &4536.6 &41893.7 \\

  \multirow{-5}{*}{Faster R-CNN}
  &\multicolumn{1}{>{\columncolor{tabcyan}}c}{CMOT}
  &\multicolumn{1}{>{\columncolor{tabcyan}}c}{9.0}
  &\multicolumn{1}{>{\columncolor{tabcyan}}c}{16.8}
  &\multicolumn{1}{>{\columncolor{tabcyan}}c}{11.2}
  &\multicolumn{1}{>{\columncolor{tabcyan}}c}{20.6}
  &\multicolumn{1}{>{\columncolor{tabcyan}}c}{43.1}
  &\multicolumn{1}{>{\columncolor{tabcyan}}c}{169.2}
  &\multicolumn{1}{>{\columncolor{tabcyan}}c}{6287.1}
  &\multicolumn{1}{>{\columncolor{tabcyan}}c}{39792.5} \\

  &\multicolumn{1}{>{\columncolor{tabpink}}c}{TBD}
  &\multicolumn{1}{>{\columncolor{tabpink}}c}{9.5}
  &\multicolumn{1}{>{\columncolor{tabpink}}c}{17.3}
  &\multicolumn{1}{>{\columncolor{tabpink}}c}{11.1}
  &\multicolumn{1}{>{\columncolor{tabpink}}c}{20.2}
  &\multicolumn{1}{>{\columncolor{tabpink}}c}{299.7}
  &\multicolumn{1}{>{\columncolor{tabpink}}c}{340.7}
  &\multicolumn{1}{>{\columncolor{tabpink}}c}{4645.5}
  &\multicolumn{1}{>{\columncolor{tabpink}}c}{40434.6} \\

  &\multicolumn{1}{>{\columncolor{tabpink}}c}{MHT}
  &\multicolumn{1}{>{\columncolor{tabpink}}c}{9.5}
  &\multicolumn{1}{>{\columncolor{tabpink}}c}{29.0}
  &\multicolumn{1}{>{\columncolor{tabpink}}c}{13.4}
  &\multicolumn{1}{>{\columncolor{tabpink}}c}{18.3}
  &\multicolumn{1}{>{\columncolor{tabpink}}c}{147.9}
  &\multicolumn{1}{>{\columncolor{tabpink}}c}{169.9}
  &\multicolumn{1}{>{\columncolor{tabpink}}c}{7466.0}
  &\multicolumn{1}{>{\columncolor{tabpink}}c}{\textbf{37809.8}} \\
  \midrule
  &\multicolumn{1}{>{\columncolor{tabpink}}c}{GOG}
  &\multicolumn{1}{>{\columncolor{tabpink}}c}{5.5}
  &\multicolumn{1}{>{\columncolor{tabpink}}c}{32.8}
  &\multicolumn{1}{>{\columncolor{tabpink}}c}{7.6}
  &\multicolumn{1}{>{\columncolor{tabpink}}c}{24.8}
  &\multicolumn{1}{>{\columncolor{tabpink}}c}{651.7}
  &\multicolumn{1}{>{\columncolor{tabpink}}c}{669.5}
  &\multicolumn{1}{>{\columncolor{tabpink}}c}{10047.6}
  &\multicolumn{1}{>{\columncolor{tabpink}}c}{49483.2} \\

  &\multicolumn{1}{>{\columncolor{tabgray}}c}{CEM}
  &\multicolumn{1}{>{\columncolor{tabgray}}c}{4.2}
  &\multicolumn{1}{>{\columncolor{tabgray}}c}{32.3}
  &\multicolumn{1}{>{\columncolor{tabgray}}c}{3.1}
  &\multicolumn{1}{>{\columncolor{tabgray}}c}{34.9}
  &\multicolumn{1}{>{\columncolor{tabgray}}c}{59.2}
  &\multicolumn{1}{>{\columncolor{tabgray}}c}{81.6}
  &\multicolumn{1}{>{\columncolor{tabgray}}c}{2522.9}
  &\multicolumn{1}{>{\columncolor{tabgray}}c}{59497.1} \\

  &\multicolumn{1}{>{\columncolor{tabcyan}}c}{DCT}
  &\multicolumn{1}{>{\columncolor{tabcyan}}c}{5.4}
  &\multicolumn{1}{>{\columncolor{tabcyan}}c}{34.0}
  &\multicolumn{1}{>{\columncolor{tabcyan}}c}{3.6}
  &\multicolumn{1}{>{\columncolor{tabcyan}}c}{35.4}
  &\multicolumn{1}{>{\columncolor{tabcyan}}c}{21.9}
  &\multicolumn{1}{>{\columncolor{tabcyan}}c}{20.3}
  &\multicolumn{1}{>{\columncolor{tabcyan}}c}{1966.0}
  &\multicolumn{1}{>{\columncolor{tabcyan}}c}{58331.5} \\

  &IHTLS   &1.5 &31.9 &7.3 &24.6 &228.6 &830.3 &16364.7 &49486.5 \\
  &H\textsuperscript{2}T &1.1 &31.6 &8.0 &24.3 &217.9 &296.5 &18241.6 &48204.0 \\
  \multirow{-6}{*}{CompACT}
  &CMOT    &1.7 &31.5 &9.3 &23.0 &76.0  &380.2 &19313.0 &46389.6 \\
  &TBD                       &3.9 &33.1 &8.9 &23.7 &457.5 &569.3 &14124.4 &48020.5 \\
  \midrule
  &GOG &-0.3 &34.9 &6.4 &25.0 &1583.4 &1555.5 &18856.2 &52261.9 \\

  &\multicolumn{1}{>{\columncolor{tabcyan}}c}{CEM}
  &\multicolumn{1}{>{\columncolor{tabcyan}}c}{0.4}
  &\multicolumn{1}{>{\columncolor{tabcyan}}c}{33.0}
  &\multicolumn{1}{>{\columncolor{tabcyan}}c}{1.9}
  &\multicolumn{1}{>{\columncolor{tabcyan}}c}{34.7}
  &\multicolumn{1}{>{\columncolor{tabcyan}}c}{159.4}
  &\multicolumn{1}{>{\columncolor{tabcyan}}c}{234.9}
  &\multicolumn{1}{>{\columncolor{tabcyan}}c}{9269.2}
  &\multicolumn{1}{>{\columncolor{tabcyan}}c}{62186.8} \\

  &\multicolumn{1}{>{\columncolor{tabpink}}c}{DCT}
  &\multicolumn{1}{>{\columncolor{tabpink}}c}{4.7}
  &\multicolumn{1}{>{\columncolor{tabpink}}c}{\textbf{35.7}}
  &\multicolumn{1}{>{\columncolor{tabpink}}c}{6.3}
  &\multicolumn{1}{>{\columncolor{tabpink}}c}{28.3}
  &\multicolumn{1}{>{\columncolor{tabpink}}c}{165.2}
  &\multicolumn{1}{>{\columncolor{tabpink}}c}{164.4}
  &\multicolumn{1}{>{\columncolor{tabpink}}c}{10498.5}
  &\multicolumn{1}{>{\columncolor{tabpink}}c}{54459.2} \\

  &IHTLS   &-1.7 &32.5 &5.4 &26.5 &326.2  &1219.4 &20631.5 &53732.0 \\
  &H\textsuperscript{2}T &-0.8 &31.3 &7.0 &24.3 &358.7 &468.0 &22443.2 &50609.6 \\
  \multirow{-6}{*}{R-CNN}
  &CMOT    &-0.5 &31.3 &8.1 &23.4 &112.3  &550.7  &24039.8 &48712.5 \\

  &\multicolumn{1}{>{\columncolor{tabgray}}c}{TBD}
  &\multicolumn{1}{>{\columncolor{tabgray}}c}{0.8}
  &\multicolumn{1}{>{\columncolor{tabgray}}c}{35.1}
  &\multicolumn{1}{>{\columncolor{tabgray}}c}{7.5}
  &\multicolumn{1}{>{\columncolor{tabgray}}c}{24.9}
  &\multicolumn{1}{>{\columncolor{tabgray}}c}{835.3}
  &\multicolumn{1}{>{\columncolor{tabgray}}c}{945.7}
  &\multicolumn{1}{>{\columncolor{tabgray}}c}{18547.4}
  &\multicolumn{1}{>{\columncolor{tabgray}}c}{51516.6} \\
  \midrule
  &\multicolumn{1}{>{\columncolor{tabgray}}c}{GOG}
  &\multicolumn{1}{>{\columncolor{tabgray}}c}{1.7}
  &\multicolumn{1}{>{\columncolor{tabgray}}c}{32.8}
  &\multicolumn{1}{>{\columncolor{tabgray}}c}{5.4}
  &\multicolumn{1}{>{\columncolor{tabgray}}c}{29.0}
  &\multicolumn{1}{>{\columncolor{tabgray}}c}{670.0}
  &\multicolumn{1}{>{\columncolor{tabgray}}c}{722.5}
  &\multicolumn{1}{>{\columncolor{tabgray}}c}{12661.8}
  &\multicolumn{1}{>{\columncolor{tabgray}}c}{54198.7} \\

  &\multicolumn{1}{>{\columncolor{tabcyan}}c}{CEM}
  &\multicolumn{1}{>{\columncolor{tabcyan}}c}{2.2}
  &\multicolumn{1}{>{\columncolor{tabcyan}}c}{31.7}
  &\multicolumn{1}{>{\columncolor{tabcyan}}c}{2.0}
  &\multicolumn{1}{>{\columncolor{tabcyan}}c}{38.5}
  &\multicolumn{1}{>{\columncolor{tabcyan}}c}{46.5}
  &\multicolumn{1}{>{\columncolor{tabcyan}}c}{69.5}
  &\multicolumn{1}{>{\columncolor{tabcyan}}c}{3506.8}
  &\multicolumn{1}{>{\columncolor{tabcyan}}c}{63274.2} \\

  &\multicolumn{1}{>{\columncolor{tabpink}}c}{DCT}
  &\multicolumn{1}{>{\columncolor{tabpink}}c}{2.4}
  &\multicolumn{1}{>{\columncolor{tabpink}}c}{34.2}
  &\multicolumn{1}{>{\columncolor{tabpink}}c}{1.6}
  &\multicolumn{1}{>{\columncolor{tabpink}}c}{40.2}
  &\multicolumn{1}{>{\columncolor{tabpink}}c}{13.7}
  &\multicolumn{1}{>{\columncolor{tabpink}}c}{13.3}
  &\multicolumn{1}{>{\columncolor{tabpink}}c}{2606.0}
  &\multicolumn{1}{>{\columncolor{tabpink}}c}{63907.1} \\

  &IHTLS   &-2.9 &31.9 &4.8 &29.0 &231.7 &864.1 &19763.8 &54480.2 \\
  &H\textsuperscript{2}T &-3.0 &32.0 &6.1 &27.8 &243.8 &326.3 &21673.6 &52671.8 \\
  \multirow{-6}{*}{ACF}
  &CMOT    &-3.0 &31.6 &7.0 &26.8 &81.5  &417.2 &23304.4 &51213.2 \\
  &TBD                       &-1.1 &33.1 &6.9 &28.0 &506.7 &611.4 &18449.9 &52744.6 \\
  \midrule
  &\multicolumn{1}{>{\columncolor{tabcyan}}c}{GOG}
  &\multicolumn{1}{>{\columncolor{tabcyan}}c}{3.0}
  &\multicolumn{1}{>{\columncolor{tabcyan}}c}{30.3}
  &\multicolumn{1}{>{\columncolor{tabcyan}}c}{4.9}
  &\multicolumn{1}{>{\columncolor{tabcyan}}c}{31.6}
  &\multicolumn{1}{>{\columncolor{tabcyan}}c}{298.6}
  &\multicolumn{1}{>{\columncolor{tabcyan}}c}{355.7}
  &\multicolumn{1}{>{\columncolor{tabcyan}}c}{9361.4}
  &\multicolumn{1}{>{\columncolor{tabcyan}}c}{55825.4} \\

  &\multicolumn{1}{>{\columncolor{tabpink}}c}{CEM}
  &\multicolumn{1}{>{\columncolor{tabpink}}c}{3.3}
  &\multicolumn{1}{>{\columncolor{tabpink}}c}{30.2}
  &\multicolumn{1}{>{\columncolor{tabpink}}c}{2.3}
  &\multicolumn{1}{>{\columncolor{tabpink}}c}{38.5}
  &\multicolumn{1}{>{\columncolor{tabpink}}c}{51.9}
  &\multicolumn{1}{>{\columncolor{tabpink}}c}{64.7}
  &\multicolumn{1}{>{\columncolor{tabpink}}c}{2053.4}
  &\multicolumn{1}{>{\columncolor{tabpink}}c}{62950.7} \\

  &\multicolumn{1}{>{\columncolor{tabgray}}c}{DCT}
  &\multicolumn{1}{>{\columncolor{tabgray}}c}{2.3}
  &\multicolumn{1}{>{\columncolor{tabgray}}c}{30.2}
  &\multicolumn{1}{>{\columncolor{tabgray}}c}{0.6}
  &\multicolumn{1}{>{\columncolor{tabgray}}c}{42.5}
  &\multicolumn{1}{>{\columncolor{tabgray}}c}{\textbf{10.4}}
  &\multicolumn{1}{>{\columncolor{tabgray}}c}{\textbf{9.9}}
  &\multicolumn{1}{>{\columncolor{tabgray}}c}{\textbf{679.0}}
  &\multicolumn{1}{>{\columncolor{tabgray}}c}{65790.5} \\

  &IHTLS   &-4.1 &30.2 &1.0 &33.3 &317.1 &784.4 &17543.5 &58228.9 \\
  &H\textsuperscript{2}T &-2.9 &29.6 &2.6 &31.7 &304.4 &278.7 &17837.4 &56158.0 \\
  \multirow{-6}{*}{DPM}
  &CMOT    &-5.9 &29.5 &5.8 &29.8 &113.9 &230.4 &24917.4 &53682.9 \\
  &TBD                       &-3.5 &32.3 &5.6 &30.4 &451.4 &402.7 &20267.4 &54478.8 \\
  \bottomrule
  \end{tabular}}
\end{table*}

\begin{table*}[t]
  \centering
  \caption{PR-MOTA, PR-MOTP, PR-MT, PR-ML, PR-IDS, PR-FM, PR-FP, and PR-FN scores of the MOT systems constructed by five object detection algorithms and seven object tracking algorithms on the {\bf {\em cloudy} subset} of the \ds benchmark dataset. The evaluation results of the winners in the UA-DETRAC Challenge 2017 and 2018 are also reported. Bold faces correspond to the best performance of the MOT systems on that metric. The pink, cyan, and gray rows denote the trackers ranked in the first, second, and third places based on the PR-MOTA score with the corresponding detector.}
  \setlength{\tabcolsep}{9.5pt}
  \renewcommand{\arraystretch}{0.5}
  \footnotesize{
  \begin{tabular}{cccccccccc}
  \toprule
  \multicolumn{10}{c}{{\bf {\em cloudy} subset}} \\
  \midrule
  Detection &Tracking &PR-MOTA &PR-MOTP &PR-MT &PR-ML &PR-IDS &PR-FM &PR-FP &PR-FN  \\
  \midrule
   \multicolumn{10}{c}{UA-DETRAC 2017 Challenge Winner} \\
  \midrule
  EB
  &IOU
  &16.0
  &22.0
  &13.1
  &17.9
  &530.7
  &551.9
  &1637.4
  &47778.5\\
  \midrule

  \multicolumn{10}{c}{UA-DETRAC 2018 Challenge Winner} \\
  \midrule
  EB
  &KIOU
  &17.4
  &21.8
  &16.5
  &17.5
  &83.5
  &123.7
  &2248.3
  &\textbf{44824.6} \\
  \midrule
  \midrule
  &GOG &14.7 &18.8 &11.1 &19.9 &672.5 &753.6 &1807.0 &51264.1 \\
  &CEM                            &2.7  &17.2  &1.7  &31.4  &153.1  &214.8  &4603.2  &72521.6 \\
  &DCT                            &12.1 &19.1  &10.5  &20.0  &212.1 &174.0  &7412.2  &51210.0 \\
  &IHTLS &14.7 &19.3 &11.1 &20.0 &125.5 &605.1 &2101.2 &51490.4 \\
  &H\textsuperscript{2}T &14.7 &19.0 &9.8 &20.5 &121.7 &177.0 &\textbf{1292.7} &52195.0 \\

  \multirow{-5}{*}{Faster R-CNN}
  &\multicolumn{1}{>{\columncolor{tabpink}}c}{CMOT}
  &\multicolumn{1}{>{\columncolor{tabpink}}c}{15.3}
  &\multicolumn{1}{>{\columncolor{tabpink}}c}{19.2}
  &\multicolumn{1}{>{\columncolor{tabpink}}c}{13.0}
  &\multicolumn{1}{>{\columncolor{tabpink}}c}{19.7}
  &\multicolumn{1}{>{\columncolor{tabpink}}c}{44.6}
  &\multicolumn{1}{>{\columncolor{tabpink}}c}{189.6}
  &\multicolumn{1}{>{\columncolor{tabpink}}c}{2938.5}
  &\multicolumn{1}{>{\columncolor{tabpink}}c}{49427.3} \\

  &\multicolumn{1}{>{\columncolor{tabcyan}}c}{TBD}
  &\multicolumn{1}{>{\columncolor{tabcyan}}c}{15.0}
  &\multicolumn{1}{>{\columncolor{tabcyan}}c}{19.8}
  &\multicolumn{1}{>{\columncolor{tabcyan}}c}{11.7}
  &\multicolumn{1}{>{\columncolor{tabcyan}}c}{19.9}
  &\multicolumn{1}{>{\columncolor{tabcyan}}c}{423.7}
  &\multicolumn{1}{>{\columncolor{tabcyan}}c}{469.0}
  &\multicolumn{1}{>{\columncolor{tabcyan}}c}{1817.3}
  &\multicolumn{1}{>{\columncolor{tabcyan}}c}{50939.6} \\

  &\multicolumn{1}{>{\columncolor{tabgray}}c}{MHT}
  &\multicolumn{1}{>{\columncolor{tabgray}}c}{14.9}
  &\multicolumn{1}{>{\columncolor{tabgray}}c}{33.3}
  &\multicolumn{1}{>{\columncolor{tabgray}}c}{14.2}
  &\multicolumn{1}{>{\columncolor{tabgray}}c}{18.6}
  &\multicolumn{1}{>{\columncolor{tabgray}}c}{168.9}
  &\multicolumn{1}{>{\columncolor{tabgray}}c}{195.4}
  &\multicolumn{1}{>{\columncolor{tabgray}}c}{4365.1}
  &\multicolumn{1}{>{\columncolor{tabgray}}c}{48682.4} \\
  \midrule
  &\multicolumn{1}{>{\columncolor{tabpink}}c}{GOG}	
  &\multicolumn{1}{>{\columncolor{tabpink}}c}{\textbf{19.4}}
  &\multicolumn{1}{>{\columncolor{tabpink}}c}{41.6}
  &\multicolumn{1}{>{\columncolor{tabpink}}c}{20.3}
  &\multicolumn{1}{>{\columncolor{tabpink}}c}{16.5}
  &\multicolumn{1}{>{\columncolor{tabpink}}c}{1124.9}
  &\multicolumn{1}{>{\columncolor{tabpink}}c}{1074.2}
  &\multicolumn{1}{>{\columncolor{tabpink}}c}{11435.8}
  &\multicolumn{1}{>{\columncolor{tabpink}}c}{54070.5} \\

  &CEM	&6.0	&38.8	&3.8	&34.6	&113.1	&150.8	&5544.4	 &87167.1\\
  &DCT	&16.6	&41.3	&11.9	&21.9	&73.2	&70.5	&5874.5	 &66122.3\\
  &IHTLS	&15.9	&41.3	&19.9	&16.7	&321.3	&1064.0	&18422.9	 &54696.4\\

  &\multicolumn{1}{>{\columncolor{tabgray}}c}{H\textsuperscript{2}T}	
  &\multicolumn{1}{>{\columncolor{tabgray}}c}{17.8}
  &\multicolumn{1}{>{\columncolor{tabgray}}c}{39.6}
  &\multicolumn{1}{>{\columncolor{tabgray}}c}{21.4}
  &\multicolumn{1}{>{\columncolor{tabgray}}c}{16.3}
  &\multicolumn{1}{>{\columncolor{tabgray}}c}{214.3}
  &\multicolumn{1}{>{\columncolor{tabgray}}c}{322.6}
  &\multicolumn{1}{>{\columncolor{tabgray}}c}{16976.1}
  &\multicolumn{1}{>{\columncolor{tabgray}}c}{52638.8} \\

  \multirow{-6}{*}{CompACT}
  &\multicolumn{1}{>{\columncolor{tabgray}}c}{CMOT}
  &\multicolumn{1}{>{\columncolor{tabgray}}c}{17.8}
  &\multicolumn{1}{>{\columncolor{tabgray}}c}{40.6}
  &\multicolumn{1}{>{\columncolor{tabgray}}c}{22.9}
  &\multicolumn{1}{>{\columncolor{tabgray}}c}{15.4}
  &\multicolumn{1}{>{\columncolor{tabgray}}c}{90.6}
  &\multicolumn{1}{>{\columncolor{tabgray}}c}{445.5}
  &\multicolumn{1}{>{\columncolor{tabgray}}c}{19320.9}
  &\multicolumn{1}{>{\columncolor{tabgray}}c}{50260.6} \\

  &\multicolumn{1}{>{\columncolor{tabcyan}}c}{TBD}
  &\multicolumn{1}{>{\columncolor{tabcyan}}c}{18.7}
  &\multicolumn{1}{>{\columncolor{tabcyan}}c}{41.7}
  &\multicolumn{1}{>{\columncolor{tabcyan}}c}{21.6}
  &\multicolumn{1}{>{\columncolor{tabcyan}}c}{16.3}
  &\multicolumn{1}{>{\columncolor{tabcyan}}c}{593.9}
  &\multicolumn{1}{>{\columncolor{tabcyan}}c}{682.9}
  &\multicolumn{1}{>{\columncolor{tabcyan}}c}{14864.3}
  &\multicolumn{1}{>{\columncolor{tabcyan}}c}{52584.0} \\
  \midrule
  &GOG &15.5 &41.4 &20.9 &16.7 &2301.5 &2129.8 &18389.2 &55330.5 \\
  &CEM	&1.9	&37.3	&2.5	&35.7	&282.9	&400.8	&12314.4	 &90119.0\\
  &DCT	&15.1	&40.5	&13.7	&20.1	&239.8	&232.5	&11866.7	 &64729.8\\
  &IHTLS	&13.3	&38.5	&19.1	&17.7	&553.7	&1795.3	&21932.2	 &57962.8\\

  &\multicolumn{1}{>{\columncolor{tabpink}}c}{H\textsuperscript{2}T}	
  &\multicolumn{1}{>{\columncolor{tabpink}}c}{17.7}
  &\multicolumn{1}{>{\columncolor{tabpink}}c}{37.2}
  &\multicolumn{1}{>{\columncolor{tabpink}}c}{22.2}
  &\multicolumn{1}{>{\columncolor{tabpink}}c}{\textbf{15.7}}
  &\multicolumn{1}{>{\columncolor{tabpink}}c}{386.7}	
  &\multicolumn{1}{>{\columncolor{tabpink}}c}{520.3}
  &\multicolumn{1}{>{\columncolor{tabpink}}c}{18618.0}
  &\multicolumn{1}{>{\columncolor{tabpink}}c}{52843.4} \\

  \multirow{-6}{*}{R-CNN}
  &\multicolumn{1}{>{\columncolor{tabgray}}c}{CMOT}
  &\multicolumn{1}{>{\columncolor{tabgray}}c}{16.3}
  &\multicolumn{1}{>{\columncolor{tabgray}}c}{40.5}	
  &\multicolumn{1}{>{\columncolor{tabgray}}c}{23.4}	
  &\multicolumn{1}{>{\columncolor{tabgray}}c}{\textbf{15.7}}	
  &\multicolumn{1}{>{\columncolor{tabgray}}c}{178.6}
  &\multicolumn{1}{>{\columncolor{tabgray}}c}{744.5}
  &\multicolumn{1}{>{\columncolor{tabgray}}c}{23241.4}
  &\multicolumn{1}{>{\columncolor{tabgray}}c}{51077.1} \\

  &\multicolumn{1}{>{\columncolor{tabcyan}}c}{TBD}
  &\multicolumn{1}{>{\columncolor{tabcyan}}c}{17.0}
  &\multicolumn{1}{>{\columncolor{tabcyan}}c}{\textbf{42.7}}
  &\multicolumn{1}{>{\columncolor{tabcyan}}c}{\textbf{21.9}}
  &\multicolumn{1}{>{\columncolor{tabcyan}}c}{20.4}
  &\multicolumn{1}{>{\columncolor{tabcyan}}c}{1174.2}
  &\multicolumn{1}{>{\columncolor{tabcyan}}c}{1177.9}
  &\multicolumn{1}{>{\columncolor{tabcyan}}c}{17672.9}
  &\multicolumn{1}{>{\columncolor{tabcyan}}c}{61218.5} \\

  \midrule
  &\multicolumn{1}{>{\columncolor{tabpink}}c}{GOG}
  &\multicolumn{1}{>{\columncolor{tabpink}}c}{16.6}
  &\multicolumn{1}{>{\columncolor{tabpink}}c}{42.4}
  &\multicolumn{1}{>{\columncolor{tabpink}}c}{19.2}
  &\multicolumn{1}{>{\columncolor{tabpink}}c}{17.8}
  &\multicolumn{1}{>{\columncolor{tabpink}}c}{1419.5}
  &\multicolumn{1}{>{\columncolor{tabpink}}c}{1427.4}
  &\multicolumn{1}{>{\columncolor{tabpink}}c}{15848.4}
  &\multicolumn{1}{>{\columncolor{tabpink}}c}{58228.8} \\

  &CEM	&5.7	&39.6	&3.8	&36.8	&121.3	&166.5	&6274.1	 &90616.5\\
  &DCT	&13.8	&42.3	&8.6	&27.2	&65.0	&60.2	&5345.3	 &75672.1\\
  &IHTLS	&12.0	&42.2	&18.1	&18.2	&459.2	&1526.3	&24547.2	 &59523.1\\

  &\multicolumn{1}{>{\columncolor{tabgray}}c}{H\textsuperscript{2}T}	
  &\multicolumn{1}{>{\columncolor{tabgray}}c}{14.4}
  &\multicolumn{1}{>{\columncolor{tabgray}}c}{40.5}	
  &\multicolumn{1}{>{\columncolor{tabgray}}c}{20.1}
  &\multicolumn{1}{>{\columncolor{tabgray}}c}{17.7}
  &\multicolumn{1}{>{\columncolor{tabgray}}c}{297.7}
  &\multicolumn{1}{>{\columncolor{tabgray}}c}{429.7}
  &\multicolumn{1}{>{\columncolor{tabgray}}c}{22640.9}
  &\multicolumn{1}{>{\columncolor{tabgray}}c}{56851.3} \\

  \multirow{-6}{*}{ACF}
  &CMOT &13.9 &41.4 &21.6 &16.5 &144.1 &643.7 &26604.7 &54021.6 \\

  &\multicolumn{1}{>{\columncolor{tabcyan}}c}{TBD}
  &\multicolumn{1}{>{\columncolor{tabcyan}}c}{15.0}
  &\multicolumn{1}{>{\columncolor{tabcyan}}c}{42.6}
  &\multicolumn{1}{>{\columncolor{tabcyan}}c}{20.8}
  &\multicolumn{1}{>{\columncolor{tabcyan}}c}{17.5}
  &\multicolumn{1}{>{\columncolor{tabcyan}}c}{841.0}
  &\multicolumn{1}{>{\columncolor{tabcyan}}c}{911.9}
  &\multicolumn{1}{>{\columncolor{tabcyan}}c}{21684.6}
  &\multicolumn{1}{>{\columncolor{tabcyan}}c}{56224.8} \\

  \midrule
  &\multicolumn{1}{>{\columncolor{tabpink}}c}{GOG}
  &\multicolumn{1}{>{\columncolor{tabpink}}c}{5.6}
  &\multicolumn{1}{>{\columncolor{tabpink}}c}{28.2}
  &\multicolumn{1}{>{\columncolor{tabpink}}c}{3.0}
  &\multicolumn{1}{>{\columncolor{tabpink}}c}{25.2}
  &\multicolumn{1}{>{\columncolor{tabpink}}c}{585.3}	
  &\multicolumn{1}{>{\columncolor{tabpink}}c}{614.1}	
  &\multicolumn{1}{>{\columncolor{tabpink}}c}{11827.8}
  &\multicolumn{1}{>{\columncolor{tabpink}}c}{70211.2} \\

  &\multicolumn{1}{>{\columncolor{tabgray}}c}{CEM}
  &\multicolumn{1}{>{\columncolor{tabgray}}c}{2.6}
  &\multicolumn{1}{>{\columncolor{tabgray}}c}{27.2}
  &\multicolumn{1}{>{\columncolor{tabgray}}c}{1.0}
  &\multicolumn{1}{>{\columncolor{tabgray}}c}{35.8}
  &\multicolumn{1}{>{\columncolor{tabgray}}c}{103.1}
  &\multicolumn{1}{>{\columncolor{tabgray}}c}{124.3}
  &\multicolumn{1}{>{\columncolor{tabgray}}c}{5646.3}
  &\multicolumn{1}{>{\columncolor{tabgray}}c}{82784.7} \\

  &\multicolumn{1}{>{\columncolor{tabcyan}}c}{DCT}
  &\multicolumn{1}{>{\columncolor{tabcyan}}c}{3.6}
  &\multicolumn{1}{>{\columncolor{tabcyan}}c}{28.6}
  &\multicolumn{1}{>{\columncolor{tabcyan}}c}{0.8}
  &\multicolumn{1}{>{\columncolor{tabcyan}}c}{37.1}
  &\multicolumn{1}{>{\columncolor{tabcyan}}c}{\textbf{35.3}}
  &\multicolumn{1}{>{\columncolor{tabcyan}}c}{\textbf{35.1}}
  &\multicolumn{1}{>{\columncolor{tabcyan}}c}{4236.5}
  &\multicolumn{1}{>{\columncolor{tabcyan}}c}{82335.9} \\

  &IHTLS	&-2.7	&26.9	&0.8	&27.9	&471.4	&1237.6	&23659.1	 &74763.0\\
  &H\textsuperscript{2}T	&1.4	&28.6	&1.7	&25.6	&465.1	&428.0	 &18898.6	&71484.7\\
  \multirow{-6}{*}{DPM}
  &CMOT	&-1.9	&28.6	&3.6	&24.2	&115.9	&318.8	&29176.7	 &68000.8\\
  &TBD                       &-3.5 &32.3 &5.6 &30.4 &451.4 &402.7 &20267.4 &54478.8 \\
  \bottomrule

  \end{tabular}}
  \label{tab:tracking-cloudy-results}
\end{table*}

\subsection{Multi-Object Tracking}
The tracking results of the MOT systems constructed by six object detection and ten object tracking methods on different subsets of the \ds benchmark are presented in the following tables: {\em overall} (Table \ref{tab:mot-results}); {\em easy} (Table \ref{tab:tracking-easy-results}), {\em medium} (Table \ref{tab:tracking-medium-results}), and {\em hard} (Table \ref{tab:tracking-hard-results}); {\em cloudy} (Table \ref{tab:tracking-cloudy-results}), {\em rainy} (Table \ref{tab:tracking-rainy-results}), {\em sunny} (Table \ref{tab:tracking-sunny-results}), and {\em night} (Table \ref{tab:tracking-night-results}). In addition, we present the performance trends of the MOT systems constructed by the evaluated detection and tracking algorithms, on the {\em overall} dataset and three subsets (\ie, {\em easy}, {\em medium}, and {\em hard}) in Figure \ref{fig:performance-trends}. In Table \ref{tab:comparison}, we compare the PR-MOTA score and the corresponding best MOTA score among all the detection thresholds for every detector. Note that the results of the best performing trackers are reported.

As shown in Table \ref{tab:mot-results}, existing MOT systems do not perform well, \eg, the top PR-MOTA score (defined in Section \ref{sec:protocol-MOT}) is only $21.1\%$, \ie, the EB$+$KIOU method. These experimental results demonstrate that more efforts are needed to improve the current tracking methods to handle challenging scenarios in the \ds dataset. Besides, the Faster R-CNN$+$CMOT ($14.2\%$ PR-MOTA score), Faster R-CNN$+$TBD ($14.4\%$ PR-MOTA score), Faster R-CNN$+$MHT ($14.5\%$ PR-MOTA score), and CompACT $+$GOG ($14.2\%$ PR-MOTA score) methods perform equally well while the DPM$+$CMOT scheme performs worst among all evaluated systems with the lowest PR-MOTA score of $-3.4\%$.

Figure \ref{fig:performance-trends} shows that a complete MOT system in general achieves better performance with better detections. The average PR-MOTA scores of all object tracking methods with the DPM, ACF, R-CNN, CompACT, and Faster R-CNN detectors are $0.41\%$, $7.84\%$, $9.49\%$, $11.40\%$, and $11.98\%$ respectively. For example, the difference of PR-MOTA scores of the DPM$+$H\textsuperscript{2}T and Faster R-CNN$+$H\textsuperscript{2}T methods ($-0.7\%$ and $13.8\%$) is significant. On the other hand, the CEM method performs relatively stably with different detectors than other trackers, \eg, the difference of the PR-MOTA scores of the ACF$+$CEM and CompACT$+$CEM (\ie, $0.6\%$) methods is much smaller than the difference of the CompACT$+$CMOT and ACF$+$CMOT (\ie, $4.8\%$) methods, and the difference of the CompACT$+$GOG and ACF$+$GOG (\ie, $3.4\%$) methods. The H\textsuperscript{2}T, DCT and IHTLS tracking algorithms use local to global optimization strategies to associate input detections, and do not resolve false positives well. When effective detections are used, the effective appearance (\eg, CMOT) or motion models (\eg, H\textsuperscript{2}T and IHTLS), and trajectory refining mechanism (\eg, DCT) adopted in these methods help tracking the objects accurately. The TBD algorithm uses a generative model to learn the high-level semantics in terms of traffic patterns. Thus, when combined with an effective detector, it can learn an accurate generative model to generate accurate object trajectories for better MOT performance in traffic scenes and achieve the best results (\ie, $14.4\%$ PR-MOTA score with the Faster R-CNN scheme). Different from these methods, the CEM scheme uses a global energy minimization strategy to reduce false positives, which helps achieve better performance than other trackers with less accurate input detections (\eg, DPM). Since the CEM scheme does not exploit target appearance information, the performance gain is not significant when more accurate detectors are used. These results show the importance of exploiting the strength of detectors and trackers in developing robust MOT systems.

However, there also exist some counter-examples. For example, the CompACT$+$GOG method performs better than the Faster R-CNN$+$GOG method with PR-MOTA of $14.2\%$ and $13.7\%$, respectively (Faster R-CNN better than CompACT); the ACF$+$CEM method performs better than the R-CNN$+$CEM method with PR-MOTA of $4.5\%$ and $2.7\%$, respectively (R-CNN performs better than ACF); and the R-CNN$+$DCT method performs better than the CompACT$+$DCT method, with PR-MOTA of $11.7\%$ and $10.8\%$, respectively (CompACT performs better than R-CNN). As shown in Figure \ref{fig:performance-trends}, the performance curves of different detectors over seven trackers on the overall dataset and three subsets (\ie, {\em easy}, {\em medium}, and {\em hard}) are intertwined with each other. Specifically, we note that different trackers perform best with different detectors, \eg, GOG achieves the best performance with CompACT on the overall set, while DCT achieves the best performance with R-CNN on the overall set. These results suggest that it is important to choose effective detector for each object tracking algorithm when constructing an MOT system. On the other hand, it is necessary to develop different types of detectors for evaluating object tracking methods comprehensively and fairly (rather than one specific detector).

\begin{table*}[t]
  \centering
  \caption{PR-MOTA, PR-MOTP, PR-MT, PR-ML, PR-IDS, PR-FM, PR-FP, and PR-FN scores of the MOT systems constructed by five object detection algorithms and seven object tracking algorithms on the {\bf {\em sunny} subset} of the \ds benchmark dataset. The evaluation results of the winners in the UA-DETRAC Challenge 2017 and 2018 are also reported. Bold faces correspond to the best performance of the MOT systems on that metric. The pink, cyan, and gray rows denote the trackers ranked in the first, second, and third places based on the PR-MOTA score with the corresponding detector.}
  \setlength{\tabcolsep}{9.5pt}
  \renewcommand{\arraystretch}{0.5}
  \footnotesize{
  \begin{tabular}{cccccccccc}
  \toprule
  \multicolumn{10}{c}{{\bf {\em sunny} subset}} \\
  \midrule
  Detection &Tracking &PR-MOTA &PR-MOTP &PR-MT &PR-ML &PR-IDS &PR-FM &PR-FP &PR-FN  \\
  \midrule
   \multicolumn{10}{c}{UA-DETRAC 2017 Challenge Winner} \\
  \midrule
  EB
  &IOU
  &35.6
  &42.8
  &32.7
  &19.3
  &401.0
  &426.9
  &2240.4
  &21065.2 \\
  \midrule

  \multicolumn{10}{c}{UA-DETRAC 2018 Challenge Winner} \\
  \midrule
  EB
  &KIOU
  &\textbf{38.3}
  &42.5
  &\textbf{38.8}
  &19.4
  &59.0
  &97.4
  &2769.1
  &18899.4 \\
  \midrule
  \midrule
  &GOG                           &20.9  &22.3   &19.9  &20.9  &352.7  &396.0  &1496.0  &20087.9 \\
  &CEM                           &5.8  &21.7 &3.8  &33.0  &90.4  &107.2  &2833.2  &30197.5 \\
  &DCT                           &16.6  &23.4  &17.8  &20.8   &80.6  &69.5  &4464.9  &20600.8 \\
  &IHTLS                         &20.6  &24.0  &20.1  &20.9   &74.1  &324.7  &1898.0  &20162.8 \\

  &\multicolumn{1}{>{\columncolor{tabgray}}c}{H\textsuperscript{2}T}
  &\multicolumn{1}{>{\columncolor{tabgray}}c}{21.4}
  &\multicolumn{1}{>{\columncolor{tabgray}}c}{24.1}
  &\multicolumn{1}{>{\columncolor{tabgray}}c}{17.5}
  &\multicolumn{1}{>{\columncolor{tabgray}}c}{20.6}
  &\multicolumn{1}{>{\columncolor{tabgray}}c}{119.4}
  &\multicolumn{1}{>{\columncolor{tabgray}}c}{145.2}
  &\multicolumn{1}{>{\columncolor{tabgray}}c}{992.8}
  &\multicolumn{1}{>{\columncolor{tabgray}}c}{20467.6} \\

  \multirow{-5}{*}{Faster R-CNN}
  &\multicolumn{1}{>{\columncolor{tabgray}}c}{CMOT}
  &\multicolumn{1}{>{\columncolor{tabgray}}c}{21.4}
  &\multicolumn{1}{>{\columncolor{tabgray}}c}{24.5}
  &\multicolumn{1}{>{\columncolor{tabgray}}c}{21.5}
  &\multicolumn{1}{>{\columncolor{tabgray}}c}{21.0}
  &\multicolumn{1}{>{\columncolor{tabgray}}c}{21.5}
  &\multicolumn{1}{>{\columncolor{tabgray}}c}{90.8}
  &\multicolumn{1}{>{\columncolor{tabgray}}c}{1934.8}
  &\multicolumn{1}{>{\columncolor{tabgray}}c}{19569.8} \\

  &\multicolumn{1}{>{\columncolor{tabcyan}}c}{TBD}
  &\multicolumn{1}{>{\columncolor{tabcyan}}c}{22.0}
  &\multicolumn{1}{>{\columncolor{tabcyan}}c}{24.9}
  &\multicolumn{1}{>{\columncolor{tabcyan}}c}{21.1}
  &\multicolumn{1}{>{\columncolor{tabcyan}}c}{20.8}
  &\multicolumn{1}{>{\columncolor{tabcyan}}c}{208.0}
  &\multicolumn{1}{>{\columncolor{tabcyan}}c}{232.2}
  &\multicolumn{1}{>{\columncolor{tabcyan}}c}{1142.1}
  &\multicolumn{1}{>{\columncolor{tabcyan}}c}{19763.0} \\

  &\multicolumn{1}{>{\columncolor{tabpink}}c}{MHT}
  &\multicolumn{1}{>{\columncolor{tabpink}}c}{23.0}
  &\multicolumn{1}{>{\columncolor{tabpink}}c}{25.0}
  &\multicolumn{1}{>{\columncolor{tabpink}}c}{24.0}
  &\multicolumn{1}{>{\columncolor{tabpink}}c}{20.6}
  &\multicolumn{1}{>{\columncolor{tabpink}}c}{28.8}
  &\multicolumn{1}{>{\columncolor{tabpink}}c}{40.8}
  &\multicolumn{1}{>{\columncolor{tabpink}}c}{1706.2}
  &\multicolumn{1}{>{\columncolor{tabpink}}c}{\textbf{18650.1}} \\
  \midrule
  &\multicolumn{1}{>{\columncolor{tabpink}}c}{GOG}
  &\multicolumn{1}{>{\columncolor{tabpink}}c}{22.5}
  &\multicolumn{1}{>{\columncolor{tabpink}}c}{44.6}
  &\multicolumn{1}{>{\columncolor{tabpink}}c}{21.5}
  &\multicolumn{1}{>{\columncolor{tabpink}}c}{16.3}
  &\multicolumn{1}{>{\columncolor{tabpink}}c}{386.1}
  &\multicolumn{1}{>{\columncolor{tabpink}}c}{361.6}	
  &\multicolumn{1}{>{\columncolor{tabpink}}c}{3561.2}
  &\multicolumn{1}{>{\columncolor{tabpink}}c}{20674.2} \\

  &CEM	&7.1	&41.5	&4.6	&37.2	&36.0	&38.6	&1725.5	 &34204.4\\
  &DCT	&16.3	&43.0	&10.2	&26.1	&23.1	&21.2	&1958.8	 &27218.9\\
  &IHTLS &18.9 &44.4 &21.1 &16.8 &112.9 &402.3 &6181.7 &20950.1 \\
  &H\textsuperscript{2}T &18.9 &43.8 &22.0 &16.9 &132.4 &138.2 &6800.5 &20338.8 \\

  \multirow{-6}{*}{CompACT}
  &\multicolumn{1}{>{\columncolor{tabgray}}c}{CMOT}
  &\multicolumn{1}{>{\columncolor{tabgray}}c}{20.2}
  &\multicolumn{1}{>{\columncolor{tabgray}}c}{43.8}
  &\multicolumn{1}{>{\columncolor{tabgray}}c}{23.5}
  &\multicolumn{1}{>{\columncolor{tabgray}}c}{\textbf{15.4}}
  &\multicolumn{1}{>{\columncolor{tabgray}}c}{27.5}
  &\multicolumn{1}{>{\columncolor{tabgray}}c}{133.8}
  &\multicolumn{1}{>{\columncolor{tabgray}}c}{6877.3}
  &\multicolumn{1}{>{\columncolor{tabgray}}c}{19377.2} \\

  &\multicolumn{1}{>{\columncolor{tabcyan}}c}{TBD}
  &\multicolumn{1}{>{\columncolor{tabcyan}}c}{21.5}
  &\multicolumn{1}{>{\columncolor{tabcyan}}c}{45.5}
  &\multicolumn{1}{>{\columncolor{tabcyan}}c}{22.8}
  &\multicolumn{1}{>{\columncolor{tabcyan}}c}{15.9}
  &\multicolumn{1}{>{\columncolor{tabcyan}}c}{252.3}
  &\multicolumn{1}{>{\columncolor{tabcyan}}c}{321.7}
  &\multicolumn{1}{>{\columncolor{tabcyan}}c}{4905.6}
  &\multicolumn{1}{>{\columncolor{tabcyan}}c}{20150.4} \\
  \midrule

  &GOG &20.1 &47.6 &23.1 &19.6 &940.3 &898.2 &6706.5 &22922.1 \\
  &CEM	&7.8	&43.7	&4.8	&35.5	&113.9	&142.5	&4226.1	 &35310.9\\

  &\multicolumn{1}{>{\columncolor{tabcyan}}c}{DCT}
  &\multicolumn{1}{>{\columncolor{tabcyan}}c}{21.1}
  &\multicolumn{1}{>{\columncolor{tabcyan}}c}{45.9}
  &\multicolumn{1}{>{\columncolor{tabcyan}}c}{19.1}
  &\multicolumn{1}{>{\columncolor{tabcyan}}c}{20.3}
  &\multicolumn{1}{>{\columncolor{tabcyan}}c}{111.1}
  &\multicolumn{1}{>{\columncolor{tabcyan}}c}{113.4}
  &\multicolumn{1}{>{\columncolor{tabcyan}}c}{5372.9}
  &\multicolumn{1}{>{\columncolor{tabcyan}}c}{24327.4} \\

  &IHTLS	&18.1	&44.2	&19.3	&21.0	&203.6	&779.4	&7675.2	 &24179.7\\
  &H\textsuperscript{2}T	&18.7	&42.7	&23.5	&21.6	&209.4	&209.8	 &8622.5	&22738.9\\

  \multirow{-6}{*}{R-CNN}
  &\multicolumn{1}{>{\columncolor{tabgray}}c}{CMOT}
  &\multicolumn{1}{>{\columncolor{tabgray}}c}{21.0}
  &\multicolumn{1}{>{\columncolor{tabgray}}c}{46.8}
  &\multicolumn{1}{>{\columncolor{tabgray}}c}{25.5}
  &\multicolumn{1}{>{\columncolor{tabgray}}c}{18.7}
  &\multicolumn{1}{>{\columncolor{tabgray}}c}{44.5}
  &\multicolumn{1}{>{\columncolor{tabgray}}c}{251.6}
  &\multicolumn{1}{>{\columncolor{tabgray}}c}{8495.2}
  &\multicolumn{1}{>{\columncolor{tabgray}}c}{21375.7} \\

  &\multicolumn{1}{>{\columncolor{tabpink}}c}{TBD}
  &\multicolumn{1}{>{\columncolor{tabpink}}c}{21.3}
  &\multicolumn{1}{>{\columncolor{tabpink}}c}{\textbf{49.8}}
  &\multicolumn{1}{>{\columncolor{tabpink}}c}{22.9}
  &\multicolumn{1}{>{\columncolor{tabpink}}c}{27.5}
  &\multicolumn{1}{>{\columncolor{tabpink}}c}{612.1}
  &\multicolumn{1}{>{\columncolor{tabpink}}c}{633.8}
  &\multicolumn{1}{>{\columncolor{tabpink}}c}{5526.5}
  &\multicolumn{1}{>{\columncolor{tabpink}}c}{28281.8} \\
  \midrule

  &\multicolumn{1}{>{\columncolor{tabpink}}c}{GOG}	
  &\multicolumn{1}{>{\columncolor{tabpink}}c}{20.5}
  &\multicolumn{1}{>{\columncolor{tabpink}}c}{45.8}
  &\multicolumn{1}{>{\columncolor{tabpink}}c}{20.4}
  &\multicolumn{1}{>{\columncolor{tabpink}}c}{17.3}
  &\multicolumn{1}{>{\columncolor{tabpink}}c}{526.9}
  &\multicolumn{1}{>{\columncolor{tabpink}}c}{513.6}
  &\multicolumn{1}{>{\columncolor{tabpink}}c}{5292.7}
  &\multicolumn{1}{>{\columncolor{tabpink}}c}{22036.3} \\

  &CEM	&9.5	&42.9	&5.7	&36.2	&45.7	&52.1	&2195.5	 &33771.0\\
  &DCT	&14.6	&44.1	&8.7	&31.1	&20.8	&20.5	&2153.3	 &30084.2\\
  &IHTLS	&16.0	&44.8	&18.9	&17.8	&168.0	&615.9	&8579.1	 &22476.0\\
  &H\textsuperscript{2}T &16.2 &44.3 &21.0 &18.1 &184.0 &183.2 &9383.8 &21531.4 \\

  \multirow{-6}{*}{ACF}
  &\multicolumn{1}{>{\columncolor{tabgray}}c}{CMOT}
  &\multicolumn{1}{>{\columncolor{tabgray}}c}{17.2}
  &\multicolumn{1}{>{\columncolor{tabgray}}c}{44.9}
  &\multicolumn{1}{>{\columncolor{tabgray}}c}{23.5}
  &\multicolumn{1}{>{\columncolor{tabgray}}c}{16.2}
  &\multicolumn{1}{>{\columncolor{tabgray}}c}{37.4}
  &\multicolumn{1}{>{\columncolor{tabgray}}c}{221.0}
  &\multicolumn{1}{>{\columncolor{tabgray}}c}{9834.3}
  &\multicolumn{1}{>{\columncolor{tabgray}}c}{20491.2} \\

  &\multicolumn{1}{>{\columncolor{tabcyan}}c}{TBD}
  &\multicolumn{1}{>{\columncolor{tabcyan}}c}{19.1}
  &\multicolumn{1}{>{\columncolor{tabcyan}}c}{46.8}
  &\multicolumn{1}{>{\columncolor{tabcyan}}c}{23.0}
  &\multicolumn{1}{>{\columncolor{tabcyan}}c}{16.7}
  &\multicolumn{1}{>{\columncolor{tabcyan}}c}{354.9}
  &\multicolumn{1}{>{\columncolor{tabcyan}}c}{409.7}
  &\multicolumn{1}{>{\columncolor{tabcyan}}c}{7388.1}
  &\multicolumn{1}{>{\columncolor{tabcyan}}c}{21178.9} \\
  \midrule

  &\multicolumn{1}{>{\columncolor{tabpink}}c}{GOG}
  &\multicolumn{1}{>{\columncolor{tabpink}}c}{7.2}
  &\multicolumn{1}{>{\columncolor{tabpink}}c}{29.4}
  &\multicolumn{1}{>{\columncolor{tabpink}}c}{6.3}
  &\multicolumn{1}{>{\columncolor{tabpink}}c}{16.8}
  &\multicolumn{1}{>{\columncolor{tabpink}}c}{316.2}
  &\multicolumn{1}{>{\columncolor{tabpink}}c}{334.0}
  &\multicolumn{1}{>{\columncolor{tabpink}}c}{6017.9}
  &\multicolumn{1}{>{\columncolor{tabpink}}c}{22758.3} \\

  &\multicolumn{1}{>{\columncolor{tabcyan}}c}{CEM}	
  &\multicolumn{1}{>{\columncolor{tabcyan}}c}{3.3}
  &\multicolumn{1}{>{\columncolor{tabcyan}}c}{28.0}
  &\multicolumn{1}{>{\columncolor{tabcyan}}c}{1.6}
  &\multicolumn{1}{>{\columncolor{tabcyan}}c}{32.6}
  &\multicolumn{1}{>{\columncolor{tabcyan}}c}{26.2}
  &\multicolumn{1}{>{\columncolor{tabcyan}}c}{27.5}
  &\multicolumn{1}{>{\columncolor{tabcyan}}c}{2060.1}
  &\multicolumn{1}{>{\columncolor{tabcyan}}c}{29890.9} \\

  &\multicolumn{1}{>{\columncolor{tabgray}}c}{DCT}
  &\multicolumn{1}{>{\columncolor{tabgray}}c}{2.2}
  &\multicolumn{1}{>{\columncolor{tabgray}}c}{31.9}
  &\multicolumn{1}{>{\columncolor{tabgray}}c}{0.3}
  &\multicolumn{1}{>{\columncolor{tabgray}}c}{40.2}
  &\multicolumn{1}{>{\columncolor{tabgray}}c}{\textbf{8.5}}	
  &\multicolumn{1}{>{\columncolor{tabgray}}c}{\textbf{7.5}}
  &\multicolumn{1}{>{\columncolor{tabgray}}c}{\textbf{838.3}}
  &\multicolumn{1}{>{\columncolor{tabgray}}c}{31962.1} \\

  &IHTLS	&-3.1	&29.4	&1.2	&19.6	&242.8	&666.5	&11080.1	 &25429.3\\
  &H\textsuperscript{2}T	&-1.6	&29.4	&3.2	&19.0	&364.3	&301.3	 &10876.6	&24378.6\\
  \multirow{-6}{*}{DPM}
  &CMOT	&-5.1	&29.7	&7.5	&15.8	&62.1	&171.8	&16312.8	 &21810.8\\
  &TBD                       &-2.3 &34.3 &8.7 &22.5 &345.1 &290.7 &13655.1 &26388.1 \\
  \bottomrule
  \end{tabular}}
  \label{tab:tracking-sunny-results}
\end{table*}

On the \ds-test set (see Table \ref{tab:mot-results}), the EB$+$KIOU method achieves the best result, \ie, $21.1\%$ PR-MOTA score. Followed by the EB$+$KIOU method, the EB$+$IOU and Faster R-CNN$+$MHT methods achieve higher PR-MOTA scores (\ie, $19.4\%$ and $14.5\%$ respectively) than other combinations of MOT systems, while the CEM scheme performs relative worse with five different detection algorithms, \ie, Faster R-CNN, CompACT, R-CNN, ACF, and DPM.

On the {\em easy} subset (see Table \ref{tab:tracking-easy-results}), both the IOU and KIOU methods perform well with approximate $35.0\%$ PR-MOTA score using the EB object detector. 
The GOG algorithm combined with the CompACT scheme performs inferior with $25.0\%$ PR-MOTA score, while the H\textsuperscript{2}T method achieves comparable PR-MOTA score of $24.4\%$ with the R-CNN detection algorithm. It is worth mentioning that these approaches outperform the Faster R-CNN$+$MHT method with PR-MOTA of $23.0\%$, which also indicates the importance of selecting combinations of detectors and trackers. On the {\em hard} subset (see Table \ref{tab:tracking-hard-results}), none of the evaluated methods perform well, \eg, with best PR-MOTA score of $13.1\%$ (EB$+$KIOU). The PR-MOTA scores of twelve MOT systems, \eg, R-CNN$+$GOG, ACF$+$CMOT, DPM$+$TBD, are all less than $0\%$, which demonstrates the difficulty of the proposed \ds dataset and the badly need of developing more robust methods for real-world applications.

On the {\em cloudy} subset (see Table \ref{tab:tracking-cloudy-results}) and {\em rainy} subset (see Table \ref{tab:tracking-rainy-results}), the compACT$+$GOG method achieves higher PR-MOTA score than the tracking methods with the input detections generated by the EB detector. It shows that the necessity to use different detectors on different scenes for better accuracy.

\begin{table*}[t]
  \centering
  \caption{PR-MOTA, PR-MOTP, PR-MT, PR-ML, PR-IDS, PR-FM, PR-FP, and PR-FN scores of the MOT systems constructed by five object detection algorithms and seven object tracking algorithms on the {\bf {\em night} subset} of the \ds benchmark dataset. The evaluation results of the winners in the UA-DETRAC Challenge 2017 and 2018 are also reported. Bold faces correspond to the best performance of the MOT systems on that metric. The pink, cyan, and gray rows denote the trackers ranked in the first, second, and third places based on the PR-MOTA score with the corresponding detector.}
  \setlength{\tabcolsep}{9.5pt}
  \renewcommand{\arraystretch}{0.5}
  \footnotesize{
  \begin{tabular}{cccccccccc}
  \toprule
  \multicolumn{10}{c}{{\bf {\em night} subset}} \\
  \midrule
  Detection &Tracking &PR-MOTA &PR-MOTP &PR-MT &PR-ML &PR-IDS &PR-FM &PR-FP &PR-FN  \\
  \midrule
   \multicolumn{10}{c}{UA-DETRAC 2017 Challenge Winner} \\
  \midrule
  EB
  &IOU
  &26.8
  &35.5
  &23.3
  &18.9
  &772.4
  &829.5
  &4139.3
  &44702.2 \\
  \midrule
  \multicolumn{10}{c}{UA-DETRAC 2018 Challenge Winner} \\
  \midrule
  EB
  &KIOU
  &\textbf{28.5}
  &35.0
  &\textbf{28.7}
  &\textbf{18.2}
  &181.7
  &276.8
  &5926.0
  &40838.6 \\
  \midrule
  \midrule
  &GOG                            &17.2  &\textbf{36.5}  &14.8  &21.1  &670.4 &702.4  &3028.5 &40709.1 \\
  &CEM                            &5.3  &\textbf{36.5}  &2.1  &33.0  &155.7 &206.9  &4291.1 &58646.4 \\
  &DCT                            &14.8  &20.3  &14.7  &20.4 &173.2 &154.7 &7978.2  &39980.2 \\
  &IHTLS                         &17.6  &20.2  &15.9  &20.5  &151.5  &583.5  &3209.4  &40374.0 \\
  &H\textsuperscript{2}T &17.7 &20.1 &14.3 &21.1 &290.0 &319.2 &2078.0 &41237.6 \\

  \multirow{-5}{*}{Faster R-CNN}
  &\multicolumn{1}{>{\columncolor{tabgray}}c}{CMOT}
  &\multicolumn{1}{>{\columncolor{tabgray}}c}{18.3}
  &\multicolumn{1}{>{\columncolor{tabgray}}c}{20.2}
  &\multicolumn{1}{>{\columncolor{tabgray}}c}{17.4}
  &\multicolumn{1}{>{\columncolor{tabgray}}c}{20.8}
  &\multicolumn{1}{>{\columncolor{tabgray}}c}{45.3}
  &\multicolumn{1}{>{\columncolor{tabgray}}c}{191.7}
  &\multicolumn{1}{>{\columncolor{tabgray}}c}{3591.8}
  &\multicolumn{1}{>{\columncolor{tabgray}}c}{38945.3} \\

  &\multicolumn{1}{>{\columncolor{tabcyan}}c}{TBD}
  &\multicolumn{1}{>{\columncolor{tabcyan}}c}{18.7}
  &\multicolumn{1}{>{\columncolor{tabcyan}}c}{20.7}
  &\multicolumn{1}{>{\columncolor{tabcyan}}c}{17.4}
  &\multicolumn{1}{>{\columncolor{tabcyan}}c}{20.3}
  &\multicolumn{1}{>{\columncolor{tabcyan}}c}{362.0}
  &\multicolumn{1}{>{\columncolor{tabcyan}}c}{403.4}
  &\multicolumn{1}{>{\columncolor{tabcyan}}c}{2187.2}
  &\multicolumn{1}{>{\columncolor{tabcyan}}c}{39499.6} \\

  &\multicolumn{1}{>{\columncolor{tabpink}}c}{MHT}
  &\multicolumn{1}{>{\columncolor{tabpink}}c}{19.5}
  &\multicolumn{1}{>{\columncolor{tabpink}}c}{20.7}
  &\multicolumn{1}{>{\columncolor{tabpink}}c}{19.9}
  &\multicolumn{1}{>{\columncolor{tabpink}}c}{20.2}
  &\multicolumn{1}{>{\columncolor{tabpink}}c}{90.3}
  &\multicolumn{1}{>{\columncolor{tabpink}}c}{119.9}
  &\multicolumn{1}{>{\columncolor{tabpink}}c}{3052.0}
  &\multicolumn{1}{>{\columncolor{tabpink}}c}{\textbf{37570.5}} \\
  \midrule
  &\multicolumn{1}{>{\columncolor{tabpink}}c}{GOG}
  &\multicolumn{1}{>{\columncolor{tabpink}}c}{10.9}
  &\multicolumn{1}{>{\columncolor{tabpink}}c}{33.2}
  &\multicolumn{1}{>{\columncolor{tabpink}}c}{6.2}
  &\multicolumn{1}{>{\columncolor{tabpink}}c}{22.6}
  &\multicolumn{1}{>{\columncolor{tabpink}}c}{831.3}
  &\multicolumn{1}{>{\columncolor{tabpink}}c}{805.1}
  &\multicolumn{1}{>{\columncolor{tabpink}}c}{4692.6}
  &\multicolumn{1}{>{\columncolor{tabpink}}c}{44419.8} \\

  &CEM	&5.5	&32.6	&1.6	&33.9	&38.0	&47.9	&\textbf{1206.9}	 &57282.7\\
  &DCT	&6.9	&34.1	&1.7	&34.3	&12.4	&9.0	 &2298.5	&53994.2\\
  &IHTLS	&8.6	&32.8	&6.6	&22.2	&190.9	&1002.9	&9683.2	 &43678.6\\
  &H\textsuperscript{2}T &9.6 &30.9 &6.9 &21.5 &250.0 &316.2 &9995.4 &41819.3 \\

  \multirow{-6}{*}{CompACT}
  &\multicolumn{1}{>{\columncolor{tabpink}}c}{CMOT}
  &\multicolumn{1}{>{\columncolor{tabpink}}c}{10.9}
  &\multicolumn{1}{>{\columncolor{tabpink}}c}{32.0}
  &\multicolumn{1}{>{\columncolor{tabpink}}c}{8.3}
  &\multicolumn{1}{>{\columncolor{tabpink}}c}{20.4}
  &\multicolumn{1}{>{\columncolor{tabpink}}c}{52.2}
  &\multicolumn{1}{>{\columncolor{tabpink}}c}{453.1}
  &\multicolumn{1}{>{\columncolor{tabpink}}c}{9888.6}
  &\multicolumn{1}{>{\columncolor{tabpink}}c}{40081.7} \\

  &\multicolumn{1}{>{\columncolor{tabpink}}c}{TBD}
  &\multicolumn{1}{>{\columncolor{tabpink}}c}{10.9}
  &\multicolumn{1}{>{\columncolor{tabpink}}c}{33.1}
  &\multicolumn{1}{>{\columncolor{tabpink}}c}{7.6}
  &\multicolumn{1}{>{\columncolor{tabpink}}c}{21.4}
  &\multicolumn{1}{>{\columncolor{tabpink}}c}{573.2}
  &\multicolumn{1}{>{\columncolor{tabpink}}c}{774.5}
  &\multicolumn{1}{>{\columncolor{tabpink}}c}{7188.8}
  &\multicolumn{1}{>{\columncolor{tabpink}}c}{42231.5} \\
  \midrule
  &GOG	&7.2	&33.0	&5.2	&22.1	&1961.4	&1874.6	&8673.6	 &46152.3\\
  &CEM	&4.6	&31.3	&1.6	&29.2	&178.1	&239.6	&6785.0	 &53971.9\\

  &\multicolumn{1}{>{\columncolor{tabcyan}}c}{DCT}
  &\multicolumn{1}{>{\columncolor{tabcyan}}c}{9.3}
  &\multicolumn{1}{>{\columncolor{tabcyan}}c}{33.3}
  &\multicolumn{1}{>{\columncolor{tabcyan}}c}{4.4}
  &\multicolumn{1}{>{\columncolor{tabcyan}}c}{23.1}
  &\multicolumn{1}{>{\columncolor{tabcyan}}c}{187.3}
  &\multicolumn{1}{>{\columncolor{tabcyan}}c}{182.7}
  &\multicolumn{1}{>{\columncolor{tabcyan}}c}{6785.0}
  &\multicolumn{1}{>{\columncolor{tabcyan}}c}{46554.3} \\

  &IHTLS	&6.0	&32.8	&5.2	&23.5	&258.6	&1391.7	&11238.1 &47199.5\\

  &H\textsuperscript{2}T &7.7 &31.6 &6.1 &21.9 &371.6 &469.0 &11767.8 &43851.8 \\

  \multirow{-6}{*}{R-CNN}
  &\multicolumn{1}{>{\columncolor{tabpink}}c}{CMOT}
  &\multicolumn{1}{>{\columncolor{tabpink}}c}{9.5}
  &\multicolumn{1}{>{\columncolor{tabpink}}c}{31.3}
  &\multicolumn{1}{>{\columncolor{tabpink}}c}{6.9}
  &\multicolumn{1}{>{\columncolor{tabpink}}c}{20.9}
  &\multicolumn{1}{>{\columncolor{tabpink}}c}{73.2}
  &\multicolumn{1}{>{\columncolor{tabpink}}c}{644.3}
  &\multicolumn{1}{>{\columncolor{tabpink}}c}{11033.5}
  &\multicolumn{1}{>{\columncolor{tabpink}}c}{42052.3} \\

  &\multicolumn{1}{>{\columncolor{tabgray}}c}{TBD}
  &\multicolumn{1}{>{\columncolor{tabgray}}c}{9.0}
  &\multicolumn{1}{>{\columncolor{tabgray}}c}{32.9}
  &\multicolumn{1}{>{\columncolor{tabgray}}c}{6.3}
  &\multicolumn{1}{>{\columncolor{tabgray}}c}{22.2}
  &\multicolumn{1}{>{\columncolor{tabgray}}c}{942.3}
  &\multicolumn{1}{>{\columncolor{tabgray}}c}{1154.0}
  &\multicolumn{1}{>{\columncolor{tabgray}}c}{8023.6}
  &\multicolumn{1}{>{\columncolor{tabgray}}c}{45063.3} \\
  \midrule
  &\multicolumn{1}{>{\columncolor{tabpink}}c}{GOG}	
  &\multicolumn{1}{>{\columncolor{tabpink}}c}{6.8}
  &\multicolumn{1}{>{\columncolor{tabpink}}c}{32.1}
  &\multicolumn{1}{>{\columncolor{tabpink}}c}{3.8}
  &\multicolumn{1}{>{\columncolor{tabpink}}c}{25.5}
  &\multicolumn{1}{>{\columncolor{tabpink}}c}{889.1}
  &\multicolumn{1}{>{\columncolor{tabpink}}c}{885.8}
  &\multicolumn{1}{>{\columncolor{tabpink}}c}{5845.5}
  &\multicolumn{1}{>{\columncolor{tabpink}}c}{48213.3} \\

  &CEM	&3.4	&31.0	&0.9	&35.3	&26.8	&40.5	&1949.4	 &58206.8\\
  &DCT	&3.4	&32.5	&1.0	&36.4	&\textbf{6.5}	&\textbf{5.1}	 &2515.3	&57697.3\\
  &IHTLS	&4.2	&31.4	&4.3	&24.9	&196.2	&1193.3	&11318.5	 &47441.2\\
  &H\textsuperscript{2}T &4.8 &30.3 &4.9 &23.6 &272.5 &368.3 &12896.2 &44873.1 \\

  \multirow{-6}{*}{ACF}
  &\multicolumn{1}{>{\columncolor{tabcyan}}c}{CMOT}
  &\multicolumn{1}{>{\columncolor{tabcyan}}c}{6.0}
  &\multicolumn{1}{>{\columncolor{tabcyan}}c}{31.0}
  &\multicolumn{1}{>{\columncolor{tabcyan}}c}{5.3}
  &\multicolumn{1}{>{\columncolor{tabcyan}}c}{23.0}
  &\multicolumn{1}{>{\columncolor{tabcyan}}c}{65.6}
  &\multicolumn{1}{>{\columncolor{tabcyan}}c}{608.2}
  &\multicolumn{1}{>{\columncolor{tabcyan}}c}{12288.1}
  &\multicolumn{1}{>{\columncolor{tabcyan}}c}{43784.3} \\

  &\multicolumn{1}{>{\columncolor{tabgray}}c}{TBD}
  &\multicolumn{1}{>{\columncolor{tabgray}}c}{6.3}
  &\multicolumn{1}{>{\columncolor{tabgray}}c}{31.7}
  &\multicolumn{1}{>{\columncolor{tabgray}}c}{5.3}
  &\multicolumn{1}{>{\columncolor{tabgray}}c}{24.2}
  &\multicolumn{1}{>{\columncolor{tabgray}}c}{687.2}
  &\multicolumn{1}{>{\columncolor{tabgray}}c}{915.3}
  &\multicolumn{1}{>{\columncolor{tabgray}}c}{9189.5}
  &\multicolumn{1}{>{\columncolor{tabgray}}c}{45802.3} \\
  \midrule
  &\multicolumn{1}{>{\columncolor{tabpink}}c}{GOG}
  &\multicolumn{1}{>{\columncolor{tabpink}}c}{8.0}
  &\multicolumn{1}{>{\columncolor{tabpink}}c}{31.6}
  &\multicolumn{1}{>{\columncolor{tabpink}}c}{4.6}
  &\multicolumn{1}{>{\columncolor{tabpink}}c}{28.3}
  &\multicolumn{1}{>{\columncolor{tabpink}}c}{515.6}
  &\multicolumn{1}{>{\columncolor{tabpink}}c}{602.2}
  &\multicolumn{1}{>{\columncolor{tabpink}}c}{9933.9}
  &\multicolumn{1}{>{\columncolor{tabpink}}c}{54420.5} \\

  &\multicolumn{1}{>{\columncolor{tabcyan}}c}{CEM}
  &\multicolumn{1}{>{\columncolor{tabcyan}}c}{5.2}
  &\multicolumn{1}{>{\columncolor{tabcyan}}c}{31.8}
  &\multicolumn{1}{>{\columncolor{tabcyan}}c}{1.3}
  &\multicolumn{1}{>{\columncolor{tabcyan}}c}{38.0}
  &\multicolumn{1}{>{\columncolor{tabcyan}}c}{60.6}
  &\multicolumn{1}{>{\columncolor{tabcyan}}c}{73.4}
  &\multicolumn{1}{>{\columncolor{tabcyan}}c}{3141.3}
  &\multicolumn{1}{>{\columncolor{tabcyan}}c}{66044.1} \\

  &\multicolumn{1}{>{\columncolor{tabgray}}c}{DCT}
  &\multicolumn{1}{>{\columncolor{tabgray}}c}{3.5}
  &\multicolumn{1}{>{\columncolor{tabgray}}c}{34.2}
  &\multicolumn{1}{>{\columncolor{tabgray}}c}{0.5}
  &\multicolumn{1}{>{\columncolor{tabgray}}c}{45.3}
  &\multicolumn{1}{>{\columncolor{tabgray}}c}{16.2}
  &\multicolumn{1}{>{\columncolor{tabgray}}c}{15.3}
  &\multicolumn{1}{>{\columncolor{tabgray}}c}{1945.7}
  &\multicolumn{1}{>{\columncolor{tabgray}}c}{69830.9} \\

  &IHTLS	&-2.4	&32.4	&1.4	&29.6	&443.5	&1303.5	&22572.0	 &58075.6\\
  &H\textsuperscript{2}T	&-1.6	&31.2	&2.2	&28.4	&567.8	&487.8	 &23406.8	&55813.0\\
  \multirow{-6}{*}{DPM}
  &CMOT	&-2.9	&31.3	&5.8	&26.8	&136.6	&321.5	&30127.8	 &51544.9\\
  &TBD                       &-0.2 &34.2 &6.5 &27.3 &613.9 &549.4 &24528.6 &52590.9 \\
  \bottomrule
  \end{tabular}}
  \label{tab:tracking-night-results}
\end{table*}

\begin{table*}[t]
  \centering
  \caption{PR-MOTA, PR-MOTP, PR-MT, PR-ML, PR-IDS, PR-FM, PR-FP, and PR-FN scores of the MOT systems constructed by five object detection algorithms and seven object tracking algorithms on the {\bf {\em rainy} subset} of the \ds benchmark dataset. The evaluation results of the winners in the UA-DETRAC Challenge 2017 and 2018 are also reported. Bold faces correspond to the best performance of the MOT systems on that metric. The pink, cyan, and gray rows denote the trackers ranked in the first, second, and third places based on the PR-MOTA score with the corresponding detector.}
  \setlength{\tabcolsep}{9.5pt}
  \renewcommand{\arraystretch}{0.5}
  \footnotesize{
  \begin{tabular}{cccccccccc}
  \toprule
  \multicolumn{10}{c}{{\bf {\em rainy} subset}} \\
  \midrule
  Detection &Tracking &PR-MOTA &PR-MOTP &PR-MT &PR-ML &PR-IDS &PR-FM &PR-FP &PR-FN  \\
  \midrule
   \multicolumn{10}{c}{UA-DETRAC 2017 Challenge Winner} \\
  \midrule
  EB
  &IOU
  &8.8
  &18.3
  &7.2
  &17.6
  &512.5
  &533.0
  &4381.4
  &48462.9 \\
  \midrule
  \multicolumn{10}{c}{UA-DETRAC 2018 Challenge Winner} \\
  \midrule
  EB
  &KIOU
  &10.0
  &18.0
  &10.3
  &\textbf{16.4}
  &103.8
  &151.6
  &5438.7
  &\textbf{45341.3} \\
  \midrule

  \midrule
  &GOG                             &8.4  &29.3   &7.5   &20.5   &550.0 &639.8  &5191.1 &53530.4 \\
  &CEM                             &1.5  &29.2 &0.9 &29.5 &114.9  &154.6 &5333.6  &68080.7 \\
  &DCT                             &5.8   &16.1  &8.3   &20.3   &165.7  &136.2  &11567.3  &52910.9 \\

  &\multicolumn{1}{>{\columncolor{tabgray}}c}{IHTLS}
  &\multicolumn{1}{>{\columncolor{tabgray}}c}{8.5}
  &\multicolumn{1}{>{\columncolor{tabgray}}c}{16.0}
  &\multicolumn{1}{>{\columncolor{tabgray}}c}{7.6}
  &\multicolumn{1}{>{\columncolor{tabgray}}c}{20.6}
  &\multicolumn{1}{>{\columncolor{tabgray}}c}{112.6}
  &\multicolumn{1}{>{\columncolor{tabgray}}c}{518.6}
  &\multicolumn{1}{>{\columncolor{tabgray}}c}{5477.6}
  &\multicolumn{1}{>{\columncolor{tabgray}}c}{53501.5} \\

  &\multicolumn{1}{>{\columncolor{tabgray}}c}{H\textsuperscript{2}T}
  &\multicolumn{1}{>{\columncolor{tabgray}}c}{8.5}
  &\multicolumn{1}{>{\columncolor{tabgray}}c}{16.1}
  &\multicolumn{1}{>{\columncolor{tabgray}}c}{7.2}
  &\multicolumn{1}{>{\columncolor{tabgray}}c}{21.3}
  &\multicolumn{1}{>{\columncolor{tabgray}}c}{172.3}
  &\multicolumn{1}{>{\columncolor{tabgray}}c}{216.2}
  &\multicolumn{1}{>{\columncolor{tabgray}}c}{4676.7}
  &\multicolumn{1}{>{\columncolor{tabgray}}c}{54339.2} \\

  \multirow{-5}{*}{Faster R-CNN}
  &\multicolumn{1}{>{\columncolor{tabcyan}}c}{CMOT}
  &\multicolumn{1}{>{\columncolor{tabcyan}}c}{8.6}
  &\multicolumn{1}{>{\columncolor{tabcyan}}c}{15.9}
  &\multicolumn{1}{>{\columncolor{tabcyan}}c}{9.0}
  &\multicolumn{1}{>{\columncolor{tabcyan}}c}{20.4}
  &\multicolumn{1}{>{\columncolor{tabcyan}}c}{46.4}
  &\multicolumn{1}{>{\columncolor{tabcyan}}c}{178.2}
  &\multicolumn{1}{>{\columncolor{tabcyan}}c}{6743.1}
  &\multicolumn{1}{>{\columncolor{tabcyan}}c}{52141.1} \\

  &\multicolumn{1}{>{\columncolor{tabpink}}c}{TBD}
  &\multicolumn{1}{>{\columncolor{tabpink}}c}{8.9}
  &\multicolumn{1}{>{\columncolor{tabpink}}c}{16.4}
  &\multicolumn{1}{>{\columncolor{tabpink}}c}{8.3}
  &\multicolumn{1}{>{\columncolor{tabpink}}c}{20.3}
  &\multicolumn{1}{>{\columncolor{tabpink}}c}{351.8}
  &\multicolumn{1}{>{\columncolor{tabpink}}c}{391.7}
  &\multicolumn{1}{>{\columncolor{tabpink}}c}{4959.0}
  &\multicolumn{1}{>{\columncolor{tabpink}}c}{52948.0} \\
  &MHT &8.4  &27.7 &10.8 &18.9 &186.6 &204.3 &8223.9 &50866.5\\

  \midrule
  &\multicolumn{1}{>{\columncolor{tabpink}}c}{GOG}
  &\multicolumn{1}{>{\columncolor{tabpink}}c}{\textbf{10.4}}
  &\multicolumn{1}{>{\columncolor{tabpink}}c}{34.1}
  &\multicolumn{1}{>{\columncolor{tabpink}}c}{10.9}
  &\multicolumn{1}{>{\columncolor{tabpink}}c}{20.2}
  &\multicolumn{1}{>{\columncolor{tabpink}}c}{1183.2}
  &\multicolumn{1}{>{\columncolor{tabpink}}c}{1112.5}
  &\multicolumn{1}{>{\columncolor{tabpink}}c}{11993.6}
  &\multicolumn{1}{>{\columncolor{tabpink}}c}{61194.0} \\

  &CEM	&4.5	&33.3	&3.2	&35.2	&93.0	&126.1	&3929.3	 &82490.1\\
  &DCT	&7.7	&35.2	&4.6	&32.3	&43.9	&41.1	&3138.7	 &76723.7\\
  &IHTLS	&7.0	&33.8	&10.6	&20.4	&347.3	&1232.1	&19588.9	 &61250.8\\

  &\multicolumn{1}{>{\columncolor{tabgray}}c}{H\textsuperscript{2}T}	
  &\multicolumn{1}{>{\columncolor{tabgray}}c}{8.9}
  &\multicolumn{1}{>{\columncolor{tabgray}}c}{33.5}
  &\multicolumn{1}{>{\columncolor{tabgray}}c}{12.0}
  &\multicolumn{1}{>{\columncolor{tabgray}}c}{19.2}
  &\multicolumn{1}{>{\columncolor{tabgray}}c}{297.5}
  &\multicolumn{1}{>{\columncolor{tabgray}}c}{379.7}
  &\multicolumn{1}{>{\columncolor{tabgray}}c}{18311.1}
  &\multicolumn{1}{>{\columncolor{tabgray}}c}{58742.4} \\

  \multirow{-6}{*}{CompACT}
  &CMOT &8.5 &33.1 &\textbf{13.0} &18.8 &118.7 &530.9 &21476.6 &56650.2 \\

  &\multicolumn{1}{>{\columncolor{tabcyan}}c}{TBD}
  &\multicolumn{1}{>{\columncolor{tabcyan}}c}{9.8}
  &\multicolumn{1}{>{\columncolor{tabcyan}}c}{34.9}
  &\multicolumn{1}{>{\columncolor{tabcyan}}c}{12.5}
  &\multicolumn{1}{>{\columncolor{tabcyan}}c}{19.6}
  &\multicolumn{1}{>{\columncolor{tabcyan}}c}{731.5}
  &\multicolumn{1}{>{\columncolor{tabcyan}}c}{840.0}
  &\multicolumn{1}{>{\columncolor{tabcyan}}c}{15887.3}
  &\multicolumn{1}{>{\columncolor{tabcyan}}c}{58987.5} \\

  \midrule
  &GOG &3.8 &36.3 &9.8 &21.8 &2529.9 &2397.6 &24592.1 &66215.2 \\
  &CEM	&0.3	&33.9	&1.4	&37.7	&194.9	&283.1	&10826.2	 &89416.9\\

  &\multicolumn{1}{>{\columncolor{tabpink}}c}{DCT}
  &\multicolumn{1}{>{\columncolor{tabpink}}c}{7.1}
  &\multicolumn{1}{>{\columncolor{tabpink}}c}{36.7}
  &\multicolumn{1}{>{\columncolor{tabpink}}c}{7.1}
  &\multicolumn{1}{>{\columncolor{tabpink}}c}{26.9}
  &\multicolumn{1}{>{\columncolor{tabpink}}c}{218.1}
  &\multicolumn{1}{>{\columncolor{tabpink}}c}{212.5}
  &\multicolumn{1}{>{\columncolor{tabpink}}c}{12174.4}
  &\multicolumn{1}{>{\columncolor{tabpink}}c}{74121.5} \\

  &IHTLS	&2.3	&34.7	&8.0	&23.2	&531.1	&1910.8	&27517.5	 &68421.8\\

  &\multicolumn{1}{>{\columncolor{tabgray}}c}{H\textsuperscript{2}T}	
  &\multicolumn{1}{>{\columncolor{tabgray}}c}{5.1}
  &\multicolumn{1}{>{\columncolor{tabgray}}c}{34.4}
  &\multicolumn{1}{>{\columncolor{tabgray}}c}{11.2}
  &\multicolumn{1}{>{\columncolor{tabgray}}c}{20.5}
  &\multicolumn{1}{>{\columncolor{tabgray}}c}{489.3}
  &\multicolumn{1}{>{\columncolor{tabgray}}c}{574.7}
  &\multicolumn{1}{>{\columncolor{tabgray}}c}{26781.2}
  &\multicolumn{1}{>{\columncolor{tabgray}}c}{63481.3} \\

  \multirow{-6}{*}{R-CNN}
  &CMOT &3.8 &35.1 &12.1 &20.5 &211.3 &836.9 &31568.7 &61486.0 \\

  &\multicolumn{1}{>{\columncolor{tabcyan}}c}{TBD}
  &\multicolumn{1}{>{\columncolor{tabcyan}}c}{5.4}
  &\multicolumn{1}{>{\columncolor{tabcyan}}c}{36.5}
  &\multicolumn{1}{>{\columncolor{tabcyan}}c}{10.8}
  &\multicolumn{1}{>{\columncolor{tabcyan}}c}{21.8}
  &\multicolumn{1}{>{\columncolor{tabcyan}}c}{1329.3}
  &\multicolumn{1}{>{\columncolor{tabcyan}}c}{1386.8}
  &\multicolumn{1}{>{\columncolor{tabcyan}}c}{22967.6}
  &\multicolumn{1}{>{\columncolor{tabcyan}}c}{64593.4} \\

  \midrule
  &\multicolumn{1}{>{\columncolor{tabpink}}c}{GOG}
  &\multicolumn{1}{>{\columncolor{tabpink}}c}{6.4}
  &\multicolumn{1}{>{\columncolor{tabpink}}c}{35.4}
  &\multicolumn{1}{>{\columncolor{tabpink}}c}{8.6}
  &\multicolumn{1}{>{\columncolor{tabpink}}c}{23.9}
  &\multicolumn{1}{>{\columncolor{tabpink}}c}{1306.9}
  &\multicolumn{1}{>{\columncolor{tabpink}}c}{1341.9}
  &\multicolumn{1}{>{\columncolor{tabpink}}c}{17473.7}
  &\multicolumn{1}{>{\columncolor{tabpink}}c}{68161.9} \\

  &CEM &3.6 &34.9 &2.7 &38.5 &82.7 &119.0 &4900.1 &87777.9 \\

  &\multicolumn{1}{>{\columncolor{tabcyan}}c}{DCT}
  &\multicolumn{1}{>{\columncolor{tabcyan}}c}{4.7}
  &\multicolumn{1}{>{\columncolor{tabcyan}}c}{\textbf{36.8}}
  &\multicolumn{1}{>{\columncolor{tabcyan}}c}{2.4}
  &\multicolumn{1}{>{\columncolor{tabcyan}}c}{39.9}
  &\multicolumn{1}{>{\columncolor{tabcyan}}c}{28.2}
  &\multicolumn{1}{>{\columncolor{tabcyan}}c}{26.5}
  &\multicolumn{1}{>{\columncolor{tabcyan}}c}{3234.7}
  &\multicolumn{1}{>{\columncolor{tabcyan}}c}{87241.7} \\

  &IHTLS	&1.5	&35.0	&7.8	&24.1	&432.4	&1515.6	&27693.8	 &68772.8\\
  &H\textsuperscript{2}T	&3.5	&34.6	&9.9	&22.0	&403.0	&498.0	 &26793.2	&65664.9\\
  \multirow{-6}{*}{ACF}
  &CMOT	&2.1	&34.4	&10.9	&21.8	&169.5	&722.9	&31718.4	 &63816.6\\

  &\multicolumn{1}{>{\columncolor{tabgray}}c}{TBD}
  &\multicolumn{1}{>{\columncolor{tabgray}}c}{4.0}
  &\multicolumn{1}{>{\columncolor{tabgray}}c}{36.2}
  &\multicolumn{1}{>{\columncolor{tabgray}}c}{10.7}
  &\multicolumn{1}{>{\columncolor{tabgray}}c}{23.0}
  &\multicolumn{1}{>{\columncolor{tabgray}}c}{928.8}
  &\multicolumn{1}{>{\columncolor{tabgray}}c}{1013.1}
  &\multicolumn{1}{>{\columncolor{tabgray}}c}{25173.5}
  &\multicolumn{1}{>{\columncolor{tabgray}}c}{65689.2} \\

  \midrule
  &\multicolumn{1}{>{\columncolor{tabpink}}c}{GOG}	
  &\multicolumn{1}{>{\columncolor{tabpink}}c}{5.1}
  &\multicolumn{1}{>{\columncolor{tabpink}}c}{30.5}
  &\multicolumn{1}{>{\columncolor{tabpink}}c}{5.1}
  &\multicolumn{1}{>{\columncolor{tabpink}}c}{27.5}
  &\multicolumn{1}{>{\columncolor{tabpink}}c}{596.5}
  &\multicolumn{1}{>{\columncolor{tabpink}}c}{596.9}
  &\multicolumn{1}{>{\columncolor{tabpink}}c}{11988}
  &\multicolumn{1}{>{\columncolor{tabpink}}c}{71805.7} \\

  &\multicolumn{1}{>{\columncolor{tabcyan}}c}{CEM}
  &\multicolumn{1}{>{\columncolor{tabcyan}}c}{4.0}
  &\multicolumn{1}{>{\columncolor{tabcyan}}c}{30.3}
  &\multicolumn{1}{>{\columncolor{tabcyan}}c}{2.2}
  &\multicolumn{1}{>{\columncolor{tabcyan}}c}{36.2}
  &\multicolumn{1}{>{\columncolor{tabcyan}}c}{83.0}
  &\multicolumn{1}{>{\columncolor{tabcyan}}c}{101.7}
  &\multicolumn{1}{>{\columncolor{tabcyan}}c}{3201.4}
  &\multicolumn{1}{>{\columncolor{tabcyan}}c}{83325.6} \\

  &\multicolumn{1}{>{\columncolor{tabgray}}c}{DCT}
  &\multicolumn{1}{>{\columncolor{tabgray}}c}{2.6}
  &\multicolumn{1}{>{\columncolor{tabgray}}c}{30.2}
  &\multicolumn{1}{>{\columncolor{tabgray}}c}{0.6}
  &\multicolumn{1}{>{\columncolor{tabgray}}c}{41.8}
  &\multicolumn{1}{>{\columncolor{tabgray}}c}{\textbf{18.3}}
  &\multicolumn{1}{>{\columncolor{tabgray}}c}{\textbf{16.7}}
  &\multicolumn{1}{>{\columncolor{tabgray}}c}{\textbf{1007.0}}
  &\multicolumn{1}{>{\columncolor{tabgray}}c}{88439.9} \\

  &IHTLS	&-2.2	&30.5	&1.5	&29.5	&471.6	&1193.3	&23492.5	 &75360.0\\
  &H\textsuperscript{2}T	&-0.3	&30.9	&2.8	&27.7	&471.3	&409.8	 &21537.0	&73429.6\\
  \multirow{-6}{*}{DPM}
  &CMOT	&-2.7	&29.9	&6.3	&26.1	&162.5	&296.9	&31485.3	 &68800.3\\
  &TBD                       &-1.0 &32.3 &6.0 &26.6 &648.1 &541.1 &26449.2 &69884.8 \\
  \bottomrule
  \end{tabular}}
  \label{tab:tracking-rainy-results}
\end{table*}

In terms of other metrics, the GOG method achieves good PR-MOTA scores combined with the CompACT scheme in the \ds benchmark but with the higher PR-IDS and PR-FM scores. For a given detection method, the MOT systems using the GOG method achieve almost more than twice larger PR-IDS scores comparison to other methods, \eg, Faster R-CNN$+$GOG ($2213.4$ PR-IDS) {\em vs.} Faster R-CNN$+$H\textsuperscript{2}T ($686.6$ PR-IDS), and CompACT$+$GOG ($3334.6$ PR-IDS) {\em vs.} CompACT$+$CMOT ($285.3$ PR-IDS). The highest PR-IDS ($7834.5$) and PR-FM ($7401.0$) scores are all generated by the R-CNN$+$GOG approach, which are also almost twice larger than other tracking methods. The GOG scheme uses a greedy algorithm to solve the optimization problem on a flow network, which may generate more false trajectories of objects, indicated by higher PR-IDS and PR-FM scores than other trackers. Thus, it is less effective for surveillance scenarios when accuracy of trajectories (\ie, lower PR-IDS and PR-FM scores) is of great importance.

For surveillance scenarios, our results suggest that the EB$+$KIOU and EB$+$KIOU methods are more effective than other alternatives with higher PR-MOTA scores and lower PR-IDS and PR-FM scores (see Table \ref{tab:mot-results}). While for the traffic safety monitoring scenarios, where false negatives (indicated by PR-FN) are more of the concern than identity switches and trajectory fragmentations, the Faster R-CNN$+$MHT or EB$+$KIOU tracking systems seem more suitable and lead to reliable performance.

\section{Run-time Performance}
We report the run-time of the evaluated object detection algorithms in Table \ref{tab:detection-speed}. Since object detection algorithms are developed on various platforms (\eg, the R-CNN \citep{DBLP:conf/cvpr/GirshickDDM14} and Faster R-CNN \citep{DBLP:journals/pami/RenHG017} methods requires a GPU for both training and testing), it is difficult to compare the running time efficiency fairly.

For the object tracking algorithms, given the input detection generated by different detection algorithms (\eg, DPM \citep{DBLP:journals/pami/FelzenszwalbGMR10}, ACF \citep{DBLP:journals/pami/DollarABP14}, R-CNN \citep{DBLP:conf/cvpr/GirshickDDM14}, CompACT \citep{DBLP:conf/iccv/CaiMN15}, and Faster R-CNN \citep{DBLP:journals/pami/RenHG017}) with the largest F-score, the average run-time on $40$ sequences in the \ds-test set are presented in Table \ref{tab:MOT-speed}. The run-time is measured on a computer with a 2.9 GHz Intel i7 processor and 16 GB memory.

The detection approaches based on deep learning are more attractive than other methods (\eg, DPM and ACF) in terms of accuracy when computing resources are not constrained. For the surveillance applications, taking both accuracy and speed into account (see Table \ref{tab:mot-results}, \ref{tab:detection-speed}, and \ref{tab:MOT-speed}), the EB$+$IOU and EB$+$KIOU systems lead to the most accurate results with relative high efficiency based on our evaluation. However, for applications with constrained computing resources, the ACF$+$CMOT and ACF$+$H\textsuperscript{2}T methods achieve higher PR-MOTA score and lower PR-IDS and PR-FM scores with relative high efficiency.

\begin{table*}[t]
\caption{Average frame-per-second of the object detection algorithms on the \ds-test set.}
\centering
\setlength{\tabcolsep}{3.0pt}
\scriptsize{
  \begin{tabular}{c|c|c|c|c|c|c|c}
  \hline
  Detectors &DPM &ACF &R-CNN &CompACT &Faster R-CNN &GP-FRCNN &EB\\
  \hline
  \multirow{3}{*}{Platform} &CPU: 4$\times$Intel Core &CPU: 2$\times$Intel Xeon  &CPU: 2$\times$Intel Xeon  &CPU: 2$\times$Intel Xeon &CPU: &CPU: 12$\times$Intel Xeon &CPU: 4$\times$Intel Core\\
  &i7-6600U (2.60GHz) &E5-2470v2 (2.40GHz) &E5-2470v2 (2.40GHz) &E5-2470v2 (2.40GHz) &E5-2695v3 (2.30GHz) &E5-2690v3 (2.60GHz) &i7-4770 (3.40GHz)\\
  & & &GPU: Tesla K40 &GPU: Tesla K40 &GPU: TitanX &GPU: Tesla K40 &GPU: TitanX\\
  \hline
  Codes &Matlab,C++ &Matlab &Matlab,C++ &Matlab,C++ &C++ &Python, C++ &C++\\
  \hline
  Speed  &0.17 &0.67 &0.10 &0.22 &11.11 &4.0 &11.00\\
  \hline
  \end{tabular}}
  \label{tab:detection-speed}
\end{table*}

\begin{table*}[t]
\caption{Average frame-per-second of the object tracking algorithms on the \ds-test set with the largest F-score detection results generated by five different detection algorithms, \ie, DPM \citep{DBLP:journals/pami/FelzenszwalbGMR10}, ACF \citep{DBLP:journals/pami/DollarABP14}, R-CNN \citep{DBLP:conf/cvpr/GirshickDDM14}, CompACT \citep{DBLP:conf/iccv/CaiMN15}, and Faster R-CNN \citep{DBLP:journals/pami/RenHG017}.}
\centering
  \setlength{\tabcolsep}{18.0pt}
  \scriptsize{
  \begin{tabular}{c|c|c|c|c|c|c|c}
  \hline
  Trackers &Codes &DPM &ACF &R-CNN &CompACT &Faster R-CNN &Average \\
  \hline

  CEM &Matlab &4.49 &3.74 &5.40 &4.62 &5.71 &4.79 \\
  GOG &Matlab &476.52 &319.29 &352.80 &389.51 &484.95 &404.61 \\
  DCT &Matlab,C++ &2.85 &1.29 &0.71 &2.19 &1.32 &1.67 \\
  IHTLS &Matlab &7.94 &5.09 &11.96 &19.79 &38.92 &16.74 \\
  H\textsuperscript{2}T &C++ &1.77 &1.08 &2.78 &3.02 &6.29 &2.99 \\
  CMOT &Matlab &4.48 &3.12 &3.59 &3.79 &3.50 &3.70 \\
  TBD     &Matlab  &0.60 &3.17 &3.17 &4.88 &8.99 &4.16 \\
  \hline
  \end{tabular}}
  \label{tab:MOT-speed}
\end{table*}

\section{Conclusions and Future Research Directions}
\label{sec:discussion-future-work}
In this work, we present a large scale multi-object tracking benchmark (\ds) consisting of $100$ videos with rich annotations. We carry out extensive experiments to evaluate the performance of twelve object detection and ten object tracking methods. We show it is necessary to understand the effect of detection accuracy on the complete MOT system performance. Using the proposed \ds metrics, we analyze the quantitative results and conclude with a discussion of the state of the art in both object detection and tracking approaches. Based on the analysis and discussion, there are several research directions worth exploring:

{\flushleft {\bf Protocol}.}
The proposed evaluation protocol can only be used to evaluate tracking methods taking detection boxes as inputs. For tracking methods that operate on likelihood maps, we can use thresholds to determine bounding boxes for each frame, such that the proposed evaluation protocol can also be used. In addition, to reduce the computational complexity in evaluation, it is necessary to design a new protocol without tuning the precision and recall rates of input detections for MOT.

{\flushleft {\bf Metrics}.}
The current evaluation metrics are designed for general tracking applications. However, in surveillance applications, we need reliably trajectories of objects. Thus, more emphasis should be made on the identity switches in evaluation. While for autonomous driving, we should avoid false negatives at all cost, but can live with identity switches. Thus, it is interesting to design some comprehensive metrics to replace the current ones, \eg, PR-MOTA and PR-MOTP, to adapt to the requirements of different application scenarios.

{\flushleft {\bf Joint detection and tracking}.}
Performance of object detection significantly affects the tracking performance, and the temporal coherency in tracking can help detection vice versa. It is of great interest to combine object detection and tracking in a unified framework to boost each other for better performance. We expect additional gains in performance of object detection and tracking from continued research on combining them.

{\flushleft {\bf Real-time issue}.}
The deep learning approaches surpass other methods by a large margin in terms of performance, especially in object detection field. However, the requirements of computational resources are very harsh to some extent. Some recent approaches \citep{DBLP:conf/cvpr/ZhangZMHS15,DBLP:conf/eccv/RastegariORF16,DBLP:conf/nips/WenWWCL16,DBLP:journals/corr/HowardZCKWWAA17} focus on pruning, compressing, or low-bit representing a ``basic'' network to adapt to embedded platforms, which aim to achieve high efficiency while maintaining comparable accuracy. We expect a stream of works coming out, focusing more on real-time constraints for object detection and MOT.

{\flushleft {\bf Data}.}
We expect to extend the current \ds dataset to include more sequences and richer annotations for evaluation on pedestrian detection and tracking approaches. We will also conduct studies to examine the effect of quantity and type of training data versus performance for both object detection and tracking fields.

\section*{Acknowledgement}
This work was supported in part by US Natural Science Foundation under Grant IIS1816227, and in part by National Nature Science Foundation of China under Grant 61472388 and Grant 61771341. 

\bibliographystyle{model2-names}
\bibliography{refs}

\section*{Appendix}
We present the proof of the range of the PR-MOTA score $\Omega^\ast$. The PR-MOTA score $\Omega^\ast$ is defined as the line integral along the PR curve, \ie,
\begin{equation}
\Omega^\ast = \frac{1}{2}\int_{{\cal C}} \Psi(p,r) \mathrm{d}{\bf s}
\end{equation}
where ${\cal C}$ is the PR curve, and $\Psi(p,r)$ is the MOTA value corresponding to the precision $p$ and recall $r$ on the PR curve. Since any MOTA score $\Psi(p,r) \in (-\infty, 100]$, we have the lower bound of $\Omega^\ast$: $-\infty$. We determine the upper bound of $\Omega^\ast$ as follows. Let ${\it C}_{0},\cdots, {\it C}_{n}$ be the dividing points on the PR curve, where the $i$-th point is ${\it C}_{i}=(p_i, r_i)$ corresponding to the precision $p_i$ and recall $r_i$ on the PR curve, and ${\it C}_{0}$ and ${\it C}_{n}$ are two end points on the PR curve. We denote the length of the $i$-th arc determined by ${\it C}_{i-1} {\it C}_{i}$ as $\Delta s_{i}$. Let $\lambda = \max_{1\leq i \leq n}\Delta s_i$. Thus, we have $\Delta s_{i} = \sqrt{{\Delta p_i}^2 + {\Delta r_i}^2}$, if $\lambda \to 0$. Then, the PR-MOTA score $\Omega^\ast$ can be rewritten as
\begin{equation}
\begin{aligned}
\Omega^\ast = \frac{1}{2}\int_{{\cal C}}\Psi(p,r) \mathrm{d}{\bf s} = \frac{1}{2}\lim_{\lambda \to 0} \sum_{i=1}^{n} \Psi(p_i, r_i)\Delta s_i
\end{aligned}
\end{equation}
$\forall i$, we have $\Psi(p_i,r_i) \leq 100$. Thus, we can get
\begin{equation}
\begin{aligned}
\Omega^\ast &\leq \frac{1}{2} \cdot 100 \cdot \lim_{\lambda \to 0} \sum_{i=1}^{n} \Delta s_i \nonumber \\
& = 50 \cdot \lim_{\lambda \to 0} \sum_{i=1}^{n} \sqrt{{\Delta p_i}^2 + {\Delta r_i}^2} \nonumber \\
& \leq 50 \cdot \lim_{\lambda \to 0} \sum_{i=1}^{n} \big( |\Delta p_i| + |\Delta r_i| \big) \nonumber \\
& = 50 \cdot \lim_{\lambda \to 0} \big( \sum_{i=1}^{n} |\Delta p_i| + \sum_{i=1}^{n} |\Delta r_i| \big)
\end{aligned}
\end{equation}
Since the precision $p$ and recall $r$ on the PR curve are in the interval $[0,1]$, \ie, $p \in [0,1]$ and $r \in [0,1]$,
we have $\sum_{i=1}^{n}|\Delta p_i| \leq 1$ and $\sum_{i=1}^{n} |\Delta r_i| \leq 1$. Thus, we obtain $\Omega^\ast \in (-\infty, 100]$. Note that the equal sign is achieved under two constraints: 1) when precision $p \neq 0$, recall $r=0$, and recall $r \neq 0$, precision $p=0$; and 2) for any precision $p$ and recall $r$ values of the input detection, the object tracking method can always achieve the MOTA score of $100$. The two constraints are the idea cases that usually do not hold in real-world applications.

\end{document}